\documentclass{article}

\usepackage[preprint]{neurips_2026}
\usepackage[utf8]{inputenc}
\usepackage[T1]{fontenc}
\usepackage{xcolor}
\usepackage{hyperref}
\definecolor{darkcite}{HTML}{0B3D91}
\definecolor{darkref}{HTML}{8B1A1A}
\hypersetup{
  colorlinks=true,
  citecolor=darkcite,
  linkcolor=darkref,
  urlcolor=darkcite,
}
\usepackage{url}
\usepackage{booktabs}
\usepackage{amsmath}
\usepackage{amsfonts}
\usepackage{microtype}
\usepackage{graphicx}
\usepackage{multirow}
\usepackage{enumitem}
\usepackage{float}
\usepackage{wrapfig}
\usepackage[font=small]{subcaption}

\newcommand{\sref}[1]{\S\ref{#1}}
\newcommand{\TabPFN}{TabPFNv2}
\newcommand{\TabICL}{TabICLv2}

\newcommand{\TabICLvOne}{TabICLv1}
\newcommand{\Mitra}{Mitra}

\title{A Mechanistic Study of Tabular Foundation Models}

\author{%
  Marin Bilo\v{s}, James T.\ Wilson, Anderson Schneider, Yuriy Nevmyvaka\\
  Machine Learning Research, Morgan Stanley \\
  \texttt{\{firstName.lastName\}@morganstanley.com}
}

\begin{document}

\maketitle

\begin{abstract}
\looseness=-1
Tabular foundation models with different architectures
converge in accuracy across a range of classification and
regression tasks. This raises questions a leaderboard
cannot answer: (i) whether the models execute the same in-context
algorithm, (ii) where row,
column, and class-permutation invariances originate, and (iii) how
robust they are under
perturbations engineered against the inferred mechanism. We characterize
all three. The model families realize qualitatively distinct
similarity-based readouts: from an attention-weighted vote
over context labels to a
class-conditional mean readout, each confirmed by
causal intervention.
We find that the representation collapse highlighted in
prior work is not a practical concern for them.
Each model's
permutation invariances trace to specific positional
parameters whose removal preserves accuracy and makes approximate
invariance exact. Perturbations engineered against each readout reproduce
predicted failure modes; hub and rank attacks
isolate them from refit baselines. Together these
results give a mechanistic account of contemporary tabular
foundation models and identify which inductive biases govern
both their accuracy and characteristic failures.
\end{abstract}

\section{Introduction}\label{sec:intro}

% Figure 1: integrative summary, placed at top of page 2.
\begin{figure}[t]
  \centering
  \captionsetup[subfigure]{font=footnotesize,skip=2pt}
  %--- top row: three readouts -------------------------------------
  \begin{subfigure}[t]{0.32\linewidth}
    \centering
    \includegraphics[height=1.45in]{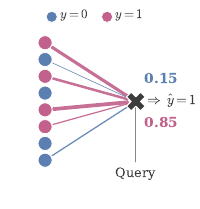}
    \subcaption{Row-attention vote: shared readout of \TabPFN{} L9
      and \Mitra{} L9}
    \label{fig:overview-a}
  \end{subfigure}\hfill
  \begin{subfigure}[t]{0.32\linewidth}
    \centering
    \includegraphics[height=1.45in]{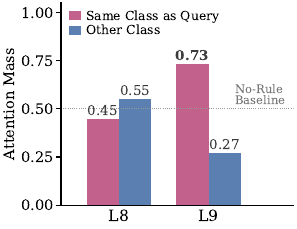}
    \subcaption{\TabPFN{} mean attention mass on context rows at L8 and L9}
    \label{fig:overview-b}
  \end{subfigure}\hfill
  \begin{subfigure}[t]{0.32\linewidth}
    \centering
    \includegraphics[height=1.45in]{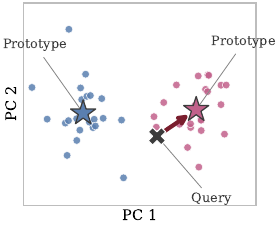}
    \subcaption{\TabICL{} predicts by nearest class prototype on
      $\mathrm{L}11$ row tokens}
    \label{fig:overview-c}
  \end{subfigure}

  \vspace{0.45em}

  %--- bottom row: causal evidence ---------------------------------
  \begin{subfigure}[t]{0.32\linewidth}
    \centering
    \includegraphics[height=1.55in]{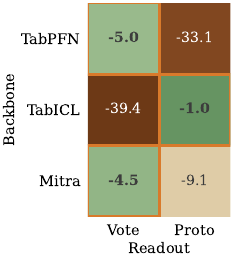}
    \subcaption{Cross-transplant accuracy drop (pp)
      vs.\ backbone native accuracy}
    \label{fig:overview-d}
  \end{subfigure}\hfill
  \begin{subfigure}[t]{0.32\linewidth}
    \centering
    \includegraphics[height=1.55in]{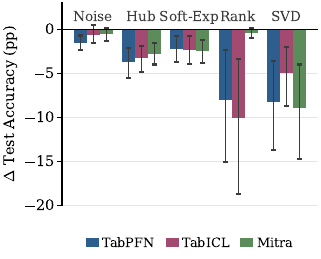}
    \subcaption{Five mechanism-grounded perturbations engineered
      against the readouts}
    \label{fig:overview-e}
  \end{subfigure}\hfill
  \begin{subfigure}[t]{0.32\linewidth}
    \centering
    \includegraphics[height=1.55in]{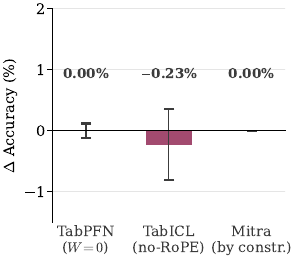}
    \subcaption{Relative \% change in accuracy after
      enforcing exact column invariance}
    \label{fig:overview-f}
  \end{subfigure}

  \caption{\textbf{Subset of results.}
  \textit{Illustration of readout mechanisms:}
  (a) attention-weighted vote for \TabPFN{} and \Mitra{}, both on layer 9;
  (b) \TabPFN{} L9 puts most of its attention mass on same-class context rows;
  (c) nearest class prototype for \TabICL{}, at L11.
  \textit{Causal evidence:}
  (d) each backbone's native readout stays within a few pp of
  end-to-end accuracy, while cross-backbone readouts collapse;
  (e) five mechanism-grounded perturbations engineered against these
  models cut test accuracy;
  (f) enforcing exact column-permutation invariance does not change the
  accuracy on all three backbones.}
  \label{fig:overview}
\end{figure}

Tabular foundation models such as
\TabPFN{} \citep{hollmann2025tabpfn},
\TabICL{} \citep{qu2025tabicl}, and
\Mitra{} \citep{zhang2025mitra} match or surpass specialized
models on tabular classification and regression
without task-specific training.
Given a labeled table at inference time,
each model predicts labels for held-out query rows
entirely through in-context learning \citep{dong2024survey}.
All three are transformer-based~\citep{vaswani2017attention}
and reach comparable benchmark accuracy
despite differing substantially in architecture.
This convergence leaves open mechanistic questions.

First, since the architectures differ in non-trivial ways,
we ask whether the learned in-context algorithm is shared.
We find that each model forms its prediction
at a qualitatively different depth and through
a mechanistically distinct readout---the
way each model turns its internal representations into a
prediction (\sref{sec:repr}).
These recipes are implemented in different parts of each
network and are only partly interchangeable across backbones:
transplanting one's readout onto another collapses
accuracy (\sref{sec:readout}).

A second question concerns a property any tabular model
should have: reordering rows or columns or renaming the classes
should not change the prediction. No model enforces this by
construction, yet we isolate the architectural components
that break each symmetry and show that the pretrained models have
largely learned to ignore them. Simple surgical edits
restore exact invariance at no cost to accuracy, and
large-scale pretraining experiments further contextualize
how invariance interacts with accuracy (\sref{sec:invariance}).
We also revisit the representation collapse concern from \citet{qu2025tabicl}
and find it does not materially affect the released models.
We prove a fundamental expressiveness barrier for
column-invariant architectures on collapse-prone data,
then validate this with minimal prototype models and
stress tests on the released models. We find that
each model's architectural mitigations sidestep the failure mode
in practice (\sref{sec:repr-collapse}).

Finally, we turn the mechanistic findings into targeted attacks.
We design data perturbations that should be invisible to a
sensible learner but strike the specific machinery each model
relies on, exposing distinct, readout-specific weaknesses
that confirm the mechanistic picture (\sref{sec:attacks}).

Together, these results show that models which score alike
on benchmarks disagree on how they form predictions, which
symmetries they respect, and where they break under pressure.
The audit surfaces many additional findings along the way;
among them, that some models have a large number of layers
that carry little load
and can be removed without hurting accuracy (\sref{sec:repr}).
Many of our results apply directly to existing checkpoints,
from restoring exact invariance to diagnosing fragilities,
while the patterns that emerge across the three families
offer concrete guidance for designing future tabular
foundation models with stronger invariances and fewer
redundant components.

\noindent\textbf{Related work.}\enspace
The three architectures we audit, per-cell attention (\TabPFN{}),
row-token attention (\TabICL{}), and a factorized label-slot
design (\Mitra{}), span the dedicated-architecture tabular
foundation model class and set the current SOTA
\citep{landsgesell2026scoringbench}; LLM-based tabular models
\citep{sui2024table} inherit a textual prior rather than a tabular
one, and same-architecture retrains on real data
\citep{ma2024tabdpt} reuse a backbone we already audit. Prior
empirical work on these models is limited:
\citep{mcelfresh2023neural} catalogues their permutation invariances
without locating the responsible parameters, \citep{gupta2026tabpfn}
traces task-relevant logic into the middle layers, and
\citet{hu2025tabpfn} maps signal-vs-noise separation in late
\TabPFN{} layers on synthetic tasks. We work on more
families, datasets, and interventions than prior work, pinning
down what current models actually do and flagging design choices
for future work.

In-context learning has been cast as gradient descent
\citep{vonoswald2023transformers,akyurek2023what}, Bayesian
inference under a prior \citep{muller2022transformers}, or
function-class learning \citep{garg2022can}; our recipes
(\sref{sec:readout}) make precise which view each backbone
instantiates. The expressiveness limit we use in
\sref{sec:repr-collapse} builds on the permutation-invariance
bottleneck of set and context-based architectures
\citep{zaheer2017deep,garnelo2018neural}, applied here to the
column-permutation case.

\section{Where representations form}
\label{sec:repr}

The goal of this section is to locate, layer by layer, where each
model builds representations that carry enough class information
for a simple classifier to read off the answer.

All three models process a table formed by labeled context rows and
unlabeled query rows.
\TabPFN{}~\citep{hollmann2025tabpfn} treats each cell of this table as
a separate token: its 12-block stack alternates row-wise attention
(mixing information across rows within each column) with column-wise
attention (mixing across columns within row).
\Mitra{}~\citep{zhang2025mitra} resembles \TabPFN{} in its per-cell
layout but differs in two important ways:
labels live in a separate per-row slot rather than being
mixed into feature tokens, and there is no positional encoding on either
axis.
\TabICL{}~\citep{qu2025tabicl} takes a different route: a
column-embedding network first compresses each row's features into a
single vector, producing one token per row. A 12-block ICL
transformer then mixes these row tokens, with context labels injected
before the first ICL block.
Full architectural details are in \sref{app:protocol-arch}.

We denote the $k$th layer of any model with $\mathrm{L}k$ and inspect
the row representations at that layer by taking activations for
both the context and query rows.
This allows us to train a simple classifier on context rows
using the known context labels, and evaluate the fit using query rows.
This linear probe tells us at which depth the model has
organized its representations to be linearly separable.
We complement this probe with three other
measurements: a silhouette score on frozen activations, dimensionality
summaries of the activation matrix, and per-block knockout
experiments.
The probe protocols and dataset lists are in \sref{app:protocol}.
A compact summary of all findings appears in
Table~\ref{tab:claim-scope}.

\subsection{Layerwise probes}\label{sec:probes}

The three models reach linearly readable class structure at very
different depths. \TabPFN{} keeps mean probe accuracy near chance
through most of the layers, then jumps sharply between
$\mathrm{L}8$ and $\mathrm{L}9$, from $0.47$ to $0.78$ on the
$49$-dataset benchmark, closing most of the gap to its original accuracy
(\sref{app:probe-convergence}; Figure~\ref{fig:layer-traj}).
The regression version of the model behaves
identically, with the peak shifted one layer earlier
(\sref{app:tabpfn-reg}).
\Mitra{} follows a similar late-jump
profile, peaking at $\mathrm{L}9$ (\sref{app:mitra-repr}).

\TabICL{} inverts this picture entirely. Its column-embedding
network plus the first ICL block already produce per-row vectors
that a $k$NN classifier reads at $0.824$ accuracy (within
$4$\,pp of end-to-end: $0.864$) so most class structure is
laid down before the deep ICL trunk runs
(\sref{app:tabicl-probes}). This means the $\sim$26M-parameter
ICL transformer, roughly $95\%$ of \TabICL{}'s parameters, adds
only marginal accuracy on top of what the column embedder provides,
suggesting that a lighter cross-row architecture could recover
most of the performance at a fraction of the cost.

\TabPFN{} and \Mitra{} build class-readable representations late
and at a single layer; \TabICL{} is readable throughout
and has an inefficient design.
This contrast motivates the readout analysis of \sref{sec:readout}.

\subsection{How the geometry evolves through the stack}
\label{sec:rank-profile}\label{sec:repr-shapes}

The models reshape their internal geometry in opposite
directions, as measured by effective rank, participation ratio, and the
twoNN intrinsic dim.\ \citep{facco2017estimating} on per-layer
activations (\sref{app:rank-profile}).

\TabPFN{} compresses early: effective rank drops from ${\sim}40$ at
blocks~$0$--$1$ to ${\sim}10$ at block~$2$.
At the final layer, rank correlates with the number of classes
($r{=}0.64$ at $\mathrm{L}11$, $r{=}0.46$ at the readout layer
$\mathrm{L}9$; \sref{app:E13}), meaning the model
uses more representational directions when more classes must be
distinguished. Because \TabPFN{}'s readout operates on attention
patterns rather than feature-space geometry (\sref{sec:readout}),
low rank need not be a bottleneck for the native prediction.

\TabICL{} runs the other way: effective rank expands monotonically
through two thirds of the stack.
This is consistent with our proposed readout rule (\sref{sec:tabicl-prototype});
it benefits from well-separated class means,
though geometry alone does not determine when class
structure becomes readable (\sref{app:tabicl-probes}).

\subsection{One dominant early block per model}
\label{sec:critical-block}\label{sec:block0-reconciliation}

Knocking out individual blocks reveals that in each model, one early
block matters far more than any other: the first transformer block
in \TabPFN{} and \Mitra{} (block~$0$), and the final block of the
column-embedding network in \TabICL{} (ColEmb-2)
(\sref{app:block-ko}). Each plays the same role: it sets up a
coordinate frame that the downstream layers rely on.

\noindent\textbf{\TabPFN{}.}\enspace
Skipping block~$0$'s MLP in the residual stream cuts
the frozen-probe accuracy at $\mathrm{L}9$ by $37.6$\,pp, yet a
probe retrained from scratch on the post-knockout activations still
recovers labels above chance: class information survives but in a
frame the rest of the network cannot read. Removing block~$9$
instead costs little with a retrained probe but cuts the frozen
probe by $32.9$\,pp (\sref{app:frozen-probe}). Block~$0$ sets up
the coordinate system; block~$9$ organizes it into the form the
readout expects.

\noindent\textbf{\Mitra{}.}\enspace
Replacing the first interleaved row/feature block with the identity
costs $15.8$\,pp on v1 and $41.1$\,pp on v1.1, the largest
single-block drop in both versions of the model.
The next-worst block costs
$\le 2$\,pp on v1 and $19.6$\,pp on v1.1
(\sref{app:mitra-block-ko}). The gap between versions suggests
that different training procedures make one lean more heavily on
its first block.

\noindent\textbf{\TabICL{}.}\enspace
ColEmb-2 is a SetTransformer that compresses per-feature
into per-row vectors. Zeroing it cuts final accuracy by $28$\,pp,
yet a probe retrained on the post-knockout ICL-input
representations still recovers labels at $0.752$ (vs.\ $0.812$
baseline). The same probe frozen from the intact model
collapses to $0.555$ (\sref{app:tabicl-colemb-frame}). ColEmb-2
plays the same coordinate-setting role as block~$0$ does for
\TabPFN{}. There is no analogous late organizing block:
knocking out any ICL block costs $<3$\,pp.

Across all three measurements, the same two-way split emerges.
\TabPFN{} and \Mitra{} concentrate their critical computation in
one early and one late block, compressing geometry in between;
\TabICL{} spreads the work across a large, mostly interchangeable
stack whose individual blocks contribute little. This split
sets up the readout analysis of \sref{sec:readout}.

\section{Readout mechanisms}
\label{sec:readout}

\looseness=-1
In \sref{sec:repr} we located \emph{where} class-readable representations
form in each stack; this section establishes \emph{how} each model
turns those representations into predictions.
Pinning down the readout tells us what each model actually
computes, and predicts where it will break.
For all models we state a simple,
falsifiable surrogate rule, test it causally with interventions, and transplant it
onto the other backbones. The classification numbers below aggregate the full
$49$-dataset benchmark; details are in \sref{app:tabpfn-cls},
\sref{app:tabicl-cls}, and \sref{app:mitra-readout}.

\looseness=-1
\noindent\textbf{Faithful surrogate.}\enspace
A candidate rule is \emph{faithful} when (i) its predicted
probabilities track the model's predicted probabilities at
Pearson $r{\geq}0.85$ on the
$49$-dataset mean, (ii) its accuracy lies within $\le3$\,pp,
(iii) argmax-agreement reaches Cohen's $\kappa{\geq}0.8$ on
a held-out test split, and
(iv) it survives an intervention
that preserves only the components named by the rule.
Per model, the joint
criterion is met on $40/49$ datasets for \TabPFN{} and on $43/49$
for \TabICL{}.
These cut-offs are conventional thresholds, but to demonstrate
the robustness of our claims,
we jointly vary $r$ and $\Delta_{\mathrm{acc}}$
within $\pm 0.05$, which shifts dataset counts by a few datasets,
preserving the original conclusions (\sref{app:rule-coverage}).
We find the following rules: an attention-weighted vote at
$\mathrm{L}9$ for \TabPFN{} and \Mitra{}, and a class-mean prototype on the final
representation for \TabICL{}.
The linear-prototype rule fails on the \TabICL{} regression model.

\subsection{TabPFN: an attention-weighted vote at a late layer}
\label{sec:tabpfn-mech}\label{sec:tabpfn-vote}

At layer~$9$ of \TabPFN{}, row-attention sharpens abruptly:
$\mathrm{L}9$ is the sharpest layer on all $49$ datasets, and the
attention distribution at that layer concentrates on a handful of
training rows (mean maximum weight $0.925$; entropy roughly half its
$\mathrm{L}0$ value; \sref{app:row-attn}).
Figure~\ref{fig:layer-traj} (middle, \sref{app:tabpfn-cls-where}) shows
that $\mathrm{L}9$ is the first layer whose attention pattern
strongly correlates with \TabPFN{}'s predictions.

\noindent\textbf{The rule.}\enspace
Read off the attention weights at $\mathrm{L}9$ and take a weighted
vote over context labels. The vote tracks the model's predicted
probabilities at mean Pearson $r{=}0.89$ (\sref{app:vote-full}).
The same picture holds on regression
at the same late readout layer (\sref{app:tabpfn-reg}).
The fidelity decreases monotonically
with class count ($\rho{=}{-}0.60$),
from binary mean $0.93$ to ten-class mean $0.80$; in that regime,
the late MLPs at $\mathrm{L}9$-$\mathrm{L}11$ carry the residual signal.
The rule is therefore strongest on binary and few-class problems.

\looseness=-1
\noindent\textbf{Causal test.}\enspace
Replacing the $\mathrm{L}9$ attention pattern with a uniform one collapses
accuracy from $0.87$ to $0.49$ (the majority-class baseline;
\sref{app:causal-attn}). The vote thus relies on the same block
that \sref{sec:critical-block} identified as the readout-building
block. The vote signal spreads across
all six attention heads at $\mathrm{L}9$ rather than concentrating in one
(\sref{app:E12}). A $k$NN classifier on \TabPFN{}'s final-layer
features agrees with its predictions on only about half of test
points, well below input-space $k$NN (\sref{app:knn-vs-tabpfn}).
The $\mathrm{L}9$ attention pattern itself is best predicted by representation
distance at L8 ($R^2 \approx 0.48$, \sref{app:E1}).
The rule therefore acts as a \emph{learned
similarity} over the context set, not a metric $k$NN on the final
features. Uniform attention intervention establishes necessity:
MLPs at $\mathrm{L}9$-$\mathrm{L}11$ may still contribute
on multi-class problems.

\subsection{\Mitra{}: an attention-weighted vote at a late layer}\label{sec:mitra-readout}

\Mitra{}'s probe peaks at the same late layer as \TabPFN{}'s vote
(we report $\mathrm{L}9$ for v1 here, $\mathrm{L}12$ for
v1.1 in \sref{app:mitra-readout}). We find the
attention-weighted vote is the closest similarity-based surrogate to
native, though it does not fully meet the faithful-surrogate bar.

\noindent\textbf{The rule.}\enspace
At $\mathrm{L}9$, read off the row-attention weights from each
query to context rows, and take a weighted vote over context labels.
Mean Pearson $r$ between vote and native probabilities is $0.89$;
and the accuracy gap is $4.5$\,pp
(\sref{app:mitra-readout}).
This gap exceeds the $\le3$\,pp faithful-surrogate threshold.
The residual error likely
reflects \Mitra{}'s column-attention pathway, which mixes $y$-slot
and feature-slot information across observations before the
row-attention vote and is not captured by the vote rule alone
(\sref{app:mitra-readout}).
Cosine $k$NN at the same layer
lies $7.0$\,pp behind the original and within
$\sim 1.5$\,pp of the $\mathrm{L}9$ linear probe. The prototype rule
trails by a further $9.0$\,pp. Despite the gap, the vote
dominates all other surrogates and the causal ablation in
\sref{app:mitra-collapse} confirms that row attention is the
load-bearing path, placing \Mitra{} in the same
``retrieval over context rows"
family as \TabPFN{}.

\subsection{TabICL: a nearest-prototype readout in the final representation}
\label{sec:tabicl-mech}\label{sec:tabicl-prototype}

\TabICL{}'s final-block representation is class-clustered enough
that a parameter-free class mean readout reproduces its predictions,
while the attention vote from \TabPFN{} fails on \TabICL{}.

\noindent\textbf{The rule.}\enspace
Form one \emph{prototype} per class by averaging the
final-block representations of context examples sharing that label,
then assign each query to the nearest prototype (Euclidean or
cosine). This rule recovers nearly all of \TabICL{}'s accuracy
(\sref{app:tabicl-prototype}); a small-$k$ $k$NN on the same
representation matches predictions on most queries
(\sref{app:knn-tabicl}). The ordering replicates across all three
released versions (Table~\ref{tab:tabicl-vx-readout}).
A linear head also reaches $0.859$, but the prototype is
parameter-free and reveals the geometric structure the model
builds, not merely that the representation is linearly separable.

\noindent\textbf{Built early, refined distributedly.}\enspace
The class-clustered geometry is laid down before the ICL stack:
within-class proximity sits significantly above chance already at
the column-embedding output, and a
linear probe here reaches $0.812$ versus $0.864$
native (\sref{app:tabicl-colemb-frame}).
No \emph{individual} ICL
block is necessary (\sref{app:tabicl-uniform-single}), but forcing
uniform attention across all twelve drops accuracy by $29$\,pp
(\sref{app:tabicl-uniform-all}), worse than the probe.
For regression, the readout is non-linear: no simple surrogate
reproduces the model's predictions, though the final representation
retains useful local structure
(\sref{app:K2}, \sref{app:tabicl-reg-frozen-postko}).

Where \TabPFN{} retrieves specific rows through sharp attention,
\TabICL{} uses no sharp attention at any layer and instead pools
entire classes into prototypes: same accuracy with a different algorithm.

\subsection{Mechanism transplantation: each readout needs its own backbone}
\label{sec:transplant}

If the three models implement distinct readouts, then a
readout fit to one backbone should fail on the others. We pass the
final-block representation of one backbone through each candidate
readout and score on the full benchmark
(\sref{app:K4}).

Swapping readouts across backbones is catastrophic
(Table~\ref{tab:transplant-summary}). The prototype rule
on \TabPFN{} drops by $33$\,pp;
the vote on \TabICL{} drops by $40$\,pp.
On \Mitra{}, which shares \TabPFN{}'s vote family, every
alternative readout (prototype, $k$NN, linear head) falls $7$ to
$9$\,pp behind native.
A freshly fit logistic head on frozen representations
closes the gap on \TabICL{} ($0.859$ vs.\ $0.864$) but
not on \TabPFN{} ($0.582$ on the shared $42$-dataset subset, well
below the $0.78$ probe at L9; \sref{app:adapt-mlphead}); a nonlinear MLP head
gains only ${\sim}1.6$\,pp more.
\TabICL{} admits a class-clustered linear readout family;
\TabPFN{} does not.

\begin{table}[t]
  \centering\small
  \setlength{\tabcolsep}{4pt}
  \caption{Readout-rule transplantation on the full $49$-dataset
    classification grid (over $5$ seeds). Each
    readout column is applied to the row backbone's frozen
    representations at the readout's preferred layer. Column headers
    list the model each rule is reported as the surrogate for.
    Brackets contain the percentage-point drop vs.\ the row's
    native accuracy. Bold marks each backbone's reported surrogate.
    Per-dataset numbers, bootstrap CIs, preferred layers, and
    the MLP variant are in
    \sref{app:K4},~\ref{app:E8},~\ref{app:adapt-mlphead},~\ref{app:mitra-readout}.}
  \label{tab:transplant-summary}
  \begin{tabular}{lccccc}
    Backbone & Native & Vote (\TabPFN{}/\Mitra{}) & $k$NN5 & Prototype (\TabICL{}) & Linear head \\
    \midrule
    \TabPFN{} & $0.854$
              & $\mathbf{0.804}\ (-5.0)$
              & $0.547\ (-30.7)$
              & $0.523\ (-33.1)$
              & $0.617\ (-23.7)$ \\
    \TabICL{} & $0.864$
              & $0.470\ (-39.5)$
              & $0.856\ (-0.8)$
              & $\mathbf{0.854}\ (-1.1)$
              & $0.859\ (-0.5)$ \\
    \Mitra{}  & $0.860$
              & $\mathbf{0.815}\ (-4.5)$
              & $0.789\ (-7.0)$
              & $0.769\ (-9.0)$
              & $0.776\ (-8.3)$ \\
  \end{tabular}
\end{table}

\noindent\textbf{Why surrogate rules matter.}\enspace
The readout rules are not claims about internal computation;
they are simple, falsifiable descriptions that reproduce each
model's input-output behavior. Their value is threefold.
First, they predict failure modes: because \TabPFN{} acts as if it
performs a weighted vote over neighbors, hub poisoning (flipping
labels of the most-attended points) directly corrupts the vote;
because \TabICL{} classifies by nearest prototype, rank warp
destroys the absolute distances its prototypes are calibrated to
(\sref{sec:attacks}).
Second, the transplant confirms that readout and backbone are
jointly designed: each model builds geometry matched to its own
rule, and swapping rules across backbones costs $33$ to $40$\,pp.
Third, the rules localize the critical computation: \TabPFN{}'s
prediction hinges on L9 attention, \TabICL{}'s on the column
embedder, giving concrete targets for pruning, debugging, and
future architecture design.

\section{Models want to be permutation invariant}
\label{sec:invariance}

A model is \emph{permutation-invariant} along an axis if its
predictions do not depend on the order in which inputs are presented
along that axis: shuffling rows of the context set, reordering the
columns of every row consistently, or relabelling classes each leave
predictions unchanged.
Invariance is the dual of the collapse
question of \sref{sec:repr-collapse}: collapse asks when distinct
inputs become identically encoded; invariance asks when
permutation-related inputs receive identical predictions.
We ask whether each model acquires invariance natively and,
where it does not, whether the missing symmetry can be installed
without retraining and without sacrificing accuracy.
We find that most models can achieve all forms of invariance,
leaving the benchmark accuracy intact.
These edits are free because the positional devices are barely used
on real data, but they exist for a reason:
\sref{sec:repr-collapse} stress-tests what happens when the
defenses they provide are truly needed.

\noindent\textbf{Architectural starting point.}\enspace
The three architectures begin from different positions on the
column-permutation axis. \TabPFN{} applies feature grouping and
a learned positional embedding to every group.
\TabICLvOne{} and \TabICL{} apply RoPE along the feature
dimension; v2 additionally enforces circular feature grouping.
\Mitra{}, in contrast, has \emph{no} positional encoding
on either axis and uses a single shared cell embedding
applied identically to every column,
making it exactly column-invariant (\sref{app:mitra-inv}).
All models are row-invariant, though \TabPFN{} has a row signature
that can be disabled at no cost to performance.
No classification model in the three families is
class-invariant by construction.

\subsection{Making pretrained models invariant}
\label{sec:inv-mechanism}

\noindent\textbf{\TabPFN{}: zeroing the per-feature positional
weight matrix.}\enspace
Setting the per-feature positional weight matrix $W$ to zero while
keeping the bias grants \emph{exact} invariance to within-pair swaps
and pair permutations and leaves benchmark accuracy unchanged on the
benchmark suite.
Interestingly, replacing the bias with only its sign also suffices
(\sref{app:inv-pfn}).

\noindent\textbf{\TabICL{}: removing RoPE.}\enspace
Removing RoPE in v2 grants \emph{exact} invariance to circular column
shifts and leaves benchmark accuracy essentially unchanged on every
dataset (\sref{app:inv-icl}). The same ablation in v1 is
catastrophic ($\sim$$15$\,pp drop), because v1 lacks v2's circular
grouping fallback (\sref{sec:repr-collapse-doubleablation}).
In \sref{sec:inv-pretraining} we pretrain an
invariant \TabICLvOne{} model from scratch.

\noindent\textbf{\Mitra{}: column invariance is exact by
construction.}\enspace
With a single shared cell embedding and no positional encoding on
either axis, column attention sees feature tokens as a set, so
feature reordering is a no-op up to numerics. Across all datasets, both
released checkpoints produce mean $|\Delta p|\!\le\!0.3\%$ on row
and column permutations and predicted-label agreement $\ge 99.5\%$
(\sref{app:mitra-inv}). The same protocol on \TabPFN{} and \TabICL{}
v1/v2 \emph{before} the edits above produces order-of-magnitude
larger column drift; our edits then close that gap.

In summary, we verify that:
\TabPFN{} with $W{=}0$
is group-permutation invariant, \TabICL{} without RoPE is
circular-shift invariant, and \TabICLvOne{} without RoPE is
arbitrary-column invariant (\sref{app:inv-synthetic}).

\noindent\textbf{Class invariance via One-vs-All.}\enspace
All classifiers depend on class order through
in-context target embeddings and a fixed output head. Wrapping any
of them in a One-vs-All classifier grants \emph{full} class-order
agreement on every dataset and leaves benchmark accuracy unchanged
(\sref{app:inv-ova}).
This works because the models were trained to solve many different tasks,
including binary classification.
Our solution does not change models' weights but introduces
a simple data processing step.
It requires calling the model $C$ times for $C$ distinct
classes, which can be trivially parallelized.

\noindent\textbf{The cost of not being invariant.}\enspace
Our edits eliminate any variability in model predictions
at zero accuracy and runtime
cost; One-vs-All adds linear-in-$C$ inference
(\sref{app:wallclock}).
Without these edits, permuting columns in a dataset causes up to
$8$\,pp accuracy difference (Figure~\ref{fig:col-perm-spread}; \sref{app:inv-bestworst}).
This is one of the reasons why released models rely on ensembling.
While it is a valid strategy to obtain approximate invariance,
we argue that it is better to include invariance from the start,
as \Mitra{} does.
The ensembling is then done over different model checkpoints
instead of different column orderings.

\subsection{Pretraining an invariant \TabICLvOne{} from scratch}
\label{sec:inv-pretraining}

\begin{wraptable}{r}{0.44\linewidth}
\centering\small
\vspace*{-0.45cm}
\caption{\TabICLvOne{} retrained from scratch. Win rate is the
per-dataset fraction at which each configuration is best of four.}
\label{tab:tabicl_pretrain_results_main}
\begin{tabular}{cccc}
Class-inv. & RoPE & Acc.\ & Win \\
\midrule
No  & Yes & $0.829$ & $0.327$ \\
No  & No  & $0.821$ & $0.249$ \\
Yes & Yes & $0.826$ & $0.282$ \\
Yes & No  & $0.803$ & $0.141$ \\
\end{tabular}
\end{wraptable}

Can an invariant model be \emph{trained} from scratch at no cost?
We conduct a large pretraining experiment of the \TabICLvOne{} model
under the stage-1 protocol of
\citet{qu2025tabicl} ($100$k iterations, batch size $512$) using
four combinations of two switches: (i) RoPE on or off along the
feature axis; (ii) the class-conditional output head replaced by a
class-invariant head (one-hot label columns, shared per-class
decoder, see \sref{app:inv-pretraining}).

\looseness=-1
Each lever alone is essentially free (RoPE removal $-0.8$\,pp,
class-invariant head $-0.3$\,pp); combined they cost $-2.6$\,pp.
The fully invariant model trains stably and stays close to the
non-invariant baseline, confirming that invariance is compatible
with ICL pretraining.
In \sref{sec:repr-collapse}, we identify which architectural components
are missing in \TabICLvOne{}'s design, which prevent it from
achieving better performance when invariant.
\Mitra{} demonstrates that the remaining
gap can be closed entirely: column invariance is a property of
its architecture, and the trained model is competitive on
benchmark accuracy and exhibits no representation collapse
(\sref{sec:repr-collapse}).
Together, the two results suggest
the path forward is architectural invariance rather than
positional hacks.

\section{On representation collapse}
\label{sec:repr-collapse}\label{sec:repr-collapse-def}

\looseness=-1
\emph{Representation collapse}~\citep{qu2025tabicl} is a per-sample
within-row identifiability failure: when several columns share the same
marginal and the row aggregator is fully column-permutation invariant,
two rows that are permutations of each other along those columns
receive identical representations even when their labels differ. The
canonical illustration is balance-scale ($4$ features, $5$ equally
frequent values; Fig.~4 in \citealp{qu2025tabicl}).
In \sref{sec:invariance}, we showed that removing one positional device is
free; this section asks what these devices actually defend against
and identifies which architectural component shields each model
from collapse.
As we have seen in \sref{sec:invariance},
\TabICLvOne{} adds RoPE along columns,
\TabICL{} additionally enforces
repeated circular feature grouping with a target-aware embedding;
\TabPFN{} uses random per-feature embedding together with
feature grouping;
and \Mitra{} does not use either positional encoding or feature grouping.

\subsection{Hand-crafted models pin down what each device does}
\label{sec:repr-collapse-handcrafted}

To explain \emph{why} different approaches succeed,
we strip the architectures down to single attention
layers we analyze by hand. All results are computed in
closed form and reproduced numerically.

\noindent\textbf{Setup.}\enspace
A row $x \in \{0,1\}^m$ is presented as $m$ cell tokens, one per
column. We study three label rules with shared Bernoulli column
marginals (so column-marginal information alone is useless):
(A) $y = x_1$ (one column is the label);
(B) $y = x_1 \oplus x_2$ (label depends jointly on two columns);
(C) $y = \mathrm{majority}(x)$ (a multiset-decidable control).
Each model is one attention layer plus a hard-argmax readout, and we
report accuracy on the full $2^m$-row enumeration with $m{=}3$. Any
model whose row representation is invariant to all column permutations
is constant on the orbits of $S_m$ and so cannot exceed the
orbit-counting bound: for the binary targets of Tasks~A and~B at
$m{=}3$ this bound equals $0.75$ (Lemma~1 for Task~A in its general
orbit-majority form; Proposition~2 for Task~B for orbit-wise
majority of $x_1\!\oplus\!x_2$; \sref{app:repr-collapse-multiset}).
The bound applies to the toy models
M0-M3 below under their assumed query-side invariance;
the mapping to the released backbones is by analogy.

\looseness=-1
\noindent\textbf{The models.}\enspace
\textbf{M0} is a naive shared-cell-embedding mean-pool: every cell
gets the same embedding, the row token averages them, and the readout
sees a column-permutation-invariant statistic. M0 hits the $0.75$
multiset bound on Task~A (\sref{app:repr-collapse-handcrafted}). The
other three models break the symmetry differently. \textbf{M1
(content route, \TabPFN{}-style)} attaches an explicit column
identifier to each cell embedding, so cell-level cross-attention
conditions on \emph{which column} as well as \emph{which value}.
\textbf{M2 (structural route, \TabICL{}-style)} uses no positional
encoding but masks attention so each head sees only one column; on
this task family M1 and M2 produce identical predictions.
\textbf{M3 (pair-grouped)} uses tokens that are super-cells over
unordered feature pairs $(j,k)$ carrying a $4$-way value and a pair
identifier: the construction we exhibit that also solves
Task~B (XOR). The released models implement variants of these
routes.
Accuracies for all four models on Tasks~A-C appear in
Figure~\ref{fig:handcrafted} (\sref{app:repr-collapse-handcrafted}).

\noindent\textbf{Mapping to the released models.}\enspace
\TabPFN{}'s feature-grouping (groups of two) is the M3 analogue and
its positional embeddings are the M1 analogue.
\TabICLvOne{} pairs the M2 analogue (column-stream attention) with
RoPE as a within-row symmetry-breaker. \TabICL{} pairs the M2
analogue with the M3 analogue (circular grouping) and adds
target-aware embeddings. Each v2 architecture therefore carries
two within-row routes, while \TabICLvOne{} carries only
one. This matches the benchmark pattern of \sref{sec:inv-mechanism}:
removing any single v2 mitigation is free, while removing RoPE in
v1 is catastrophic.

\subsection{Direct stress test on collapse-prone data}
\label{sec:repr-collapse-doubleablation}

The released v2 checkpoints carry redundant within-row defenses;
collapse only appears when we strip them. We probe \emph{which}
mitigations are load-bearing on
the balance-scale dataset ($n{=}625$, $d{=}4$) and a $d{=}12$ balance-like
synthetic stress dataset with identical marginals
(\sref{app:K8}).

\noindent\textbf{One vs.\ two within-row devices removed (v2 backbones).}\enspace
Removing any \emph{single} within-row defense on a v2 backbone is
approximately free for $W{=}0$ on \TabPFN{} and no-RoPE on \TabICL{}
(both within $|\Delta|\!\le\!0.001$); the M3 pair-channel ablation
($\textnormal{2nd-slot}{=}0$) costs up to ${\sim}14$\,pp on the most
stressed slice (Tables~\ref{tab:K8-collapse-fsweep} and
\ref{tab:K8b-v2-combined}).
Removing \emph{both} exposes the
latent collapse: \TabICL{} with RoPE and circular grouping removed
drops to the majority baseline on the $d{=}12$ stress ($-50$\,pp),
and \TabPFN{} with both within-row routes off ($W{=}0$ and pair
second-slot$=0$) drops by $7$--$12$\,pp (Table~\ref{tab:K8b-v2-combined}).
Each v2 architecture therefore carries a \emph{redundant} within-row
stack, not a single defense.

\looseness=-1
\noindent\textbf{\TabICLvOne{} versus stripped v2.}\enspace
\TabICLvOne{} no-RoPE drops $44$\,pp on balance-scale, larger than
stripped v2. Under the harder $d{=}12$ stress,
stripped \TabICL{} also collapses by $50$\,pp. The buffer that
delays v2's collapse is the target-aware embedding, which v1 lacks
and which survives both v2 ablations. Both ICL backbones
collapse to the majority baseline
once their within-row defenses are gone.

\noindent\textbf{What makes \Mitra{} special.}\enspace
\Mitra{} has no within-row symmetry-breaker by design and
still matches \TabPFN{} with all its defenses in place
($0.96 / 0.97$ vs.\ $0.96 / 0.97$ on balance-scale and balance-like
$d{=}12$). The relevant counterfactual is v2 \emph{stripped} of its
within-row defenses: \TabPFN{} with both routes off only drops to
$0.89 / 0.86$, well above majority and in the same regime as
\Mitra{}. Compared to \TabICL{},
\Mitra{} and \TabPFN{} are more robust to combined
within-row stripping.
We posit that introducing the context label earlier in the stack
and the lack of row compression are the main drivers of this phenomenon.
The attacks of \sref{sec:attacks}
follow from the resulting structural difference.

\looseness=-1
\noindent\textbf{A recipe for the next generation of models.}\enspace
When designing the next tabular foundation model, we suggest
adopting all three of the following:
(i) make the model jointly column-invariant by construction;
(ii) inject the label early, as a slot visible to column attention
from the first layer;
(iii) read out from the in-context labels rather than from a fixed
class-conditional head, for example via one-hot labels.
Built-in invariance is the more principled design:
it makes the predictor a deterministic function of the
context, removes the need for permutation ensembling
(\sref{sec:invariance}), and closes the symmetry-attack surface that
the released models exhibit (\sref{sec:attacks}).
The three are complementary, not interchangeable.
\Mitra{} adopts all three and reaches benchmark accuracy without any
within-row positional device. Adopting only (i) and (iii) leaves an
unrecovered gap: \TabICLvOne{} retrained from scratch in this
configuration loses $2.6$\,pp relative to its non-invariant baseline
because it lacks (ii) (\sref{sec:inv-pretraining}).

\section{Mechanism-grounded adversarial tests}
\label{sec:attacks}

If the readouts of \sref{sec:readout} describe what each model
computes, they predict which context-set perturbations should hurt
which model. We design eight perturbations
(\sref{app:attack-defs}) grouped by what they target;
the readouts' geometry: \emph{hub poison} flips the labels of
context points that sit closest on average to all others, and
\emph{rank warp} replaces values by their per-column rank,
preserving order but destroying absolute scale; context
corruptions that affect any flexible classifier: \emph{noise
padding}, \emph{centroid} and \emph{boundary} label injection, and
\emph{SVD burial} (joint reweighting of high and low-variance
directions); smooth monotone warps: \emph{cube} and
\emph{soft-exp}, a Bayesian average over context-conditional
predictors should largely absorb them; and a null-space PGD diagnostic
(\sref{app:attack-nullspace}). All three
foundation models are run on the same grid; Ridge, XGBoost, MLP
baselines fit fresh on the same poisoned context
(\sref{app:attacks}).

\begin{table}[t]
  \centering\small
  \setlength{\tabcolsep}{4pt}
  \caption{
    Mechanism-grounded attacks on a common $24$-classification
    grid ($\times\,5$ seeds): within-model $\Delta$\,pp with
    bootstrap CIs. Clean accuracies are $0.882$ (\TabPFN{}),
    $0.888$ (\TabICL{}), and $0.881$ (\Mitra{} v1.1; v1 within
    $1$\,pp). $^{*}$ marks Holm-corrected signif.\
    ($\alpha{=}0.05$). See \sref{app:attacks},~\ref{app:mitra-attacks}
    for more results.}
  \label{tab:attacks-drops}
  \begin{tabular}{lccc}
    Attack            & \TabPFN{} $\Delta$\,pp [95\% CI] & \TabICL{} $\Delta$\,pp [95\% CI] & \Mitra{} $\Delta$\,pp [95\% CI] \\
    \midrule
    Noise pad         & $-1.5^{*}\,[-2.4,-0.7]$    & $-0.6\;\;[-1.5,+0.5]$  & $-0.5\;\;[-1.3,+0.2]$ \\
    Hub poison        & $-3.7^{*}\,[-5.5,-2.1]$    & $-3.3^{*}\,[-4.8,-1.9]$ & $-2.8^{*}\,[-4.0,-1.6]$ \\
    Centroid inj.     & $-0.2\;\;[-0.9,+0.6]$      & $-0.3\;\;[-1.2,+0.5]$  & $-0.3\;\;[-0.9,+0.4]$ \\
    Boundary poison   & $-1.7^{*}\,[-2.5,-1.0]$    & $-1.2\;\;[-2.2,-0.1]$  & $-1.5\;\;[-3.3,+0.4]$ \\
    Cube warp         & $-1.1\;\;[-2.5,+0.1]$      & $-0.3\;\;[-1.4,+1.2]$  & $-0.3\;\;[-0.8,+0.0]$ \\
    Soft-exp warp     & $-2.2^{*}\,[-3.7,-0.8]$    & $-2.3^{*}\,[-3.9,-0.8]$ & $-2.5^{*}\,[-3.8,-1.2]$ \\
    Rank warp         & $-8.0^{*}\,[-15.1,-2.4]$   & $-10.1^{*}\,[-18.7,-3.4]$ & $-0.4\;\;[-1.0,+0.1]$ \\
    SVD burial        & $-8.3^{*}\,[-13.7,-3.6]$   & $-5.0^{*}\,[-8.7,-2.0]$ & $-8.9^{*}\,[-14.7,-4.0]$ \\
  \end{tabular}
\end{table}

The drops in Table~\ref{tab:attacks-drops}, set against the refit
baselines of \sref{app:attacks}, follow this split. The two
geometry attacks hurt \TabPFN{}/\TabICL{} substantially more than
a tuned MLP refit on the same poisoned context, isolating the
readouts as the source; \Mitra{} matches the MLP on hub poisoning
and is immune to rank warp by virtue of its quantile front-end.
The four context corruptions hurt the MLP at least as much as the
foundation models, so the drops there---including the large SVD
burial drops on \TabPFN{} and \Mitra{}---reflect sensitivity any
flexible classifier inherits from a corrupted context. The two
monotone warps hurt the MLP roughly twice as much as the
foundation models, consistent with their pretraining prior
already covering smooth monotone warps that a freshly fit MLP
has to learn from a single context. A pre-specified Wilcoxon test on
(rank warp, hub poison) vs.\ (cube warp, centroid injection)
rejects at $p<10^{-3}$ for \TabPFN{}/\TabICL{}; the $16$-dataset
mixed-type slice and the $49$-dataset \Mitra{} sweep reproduce
the ordering.

\section{Discussion and limitations}\label{sec:limitations}\label{sec:summary}\label{sec:limit-no-pretraining}\label{sec:repro}

The audit covers $49$ classification and $10$ regression
datasets, multiple permutation and perturbation grids and paired
bootstrap intervals throughout: in aggregate, tens of thousands
of evaluation points across the three foundation models.
\TabPFN{} and \Mitra{} read out through an
\textbf{attention-weighted vote at a late layer} and \TabICL{}
through a \textbf{nearest-prototype rule} in its final
representation, and swapping these heads across backbones costs
$33$ to $40$ accuracy points: readout and backbone are
\textbf{jointly designed}. Much of each backbone is
\textbf{not} load-bearing: the v2 models carry a
\textbf{redundant within-row stack}, per-block knockouts cost
little at almost every layer, the \TabPFN{} readout is nailed at
one late block, and on \TabICL{} most of the class signal is
already present at the ICL input. \Mitra{} contains \emph{no}
within-row symmetry-breakers yet matches the others, proving
representation collapse is not fundamental but
\textbf{architectural}. On the permutation axis, \TabPFN{} and
\TabICL{} admit a \textbf{one-line edit to exact column
invariance} and a One-vs-All wrapper grants \textbf{exact class
invariance} on any of the three; a from-scratch invariant
\TabICLvOne{} trains stably but gains no accuracy,
providing further evidence for our prescribed model.
The same story predicts the readout-specific breakages:
\textbf{rank warp} and \textbf{hub poisoning} hurt
\TabPFN{}/\TabICL{} more than a refit MLP does; \Mitra{} is
immune to cube and rank warps but shares the $k$NN-style hub
weakness.

\noindent\textbf{Limitations.}\enspace
Two directions are out of scope. We hold the
pretraining mixture fixed and do not probe how the synthetic data
generator itself shapes the mechanisms we audit, and we do not
cover the very large dataset regime, where in-context
inference becomes infeasible.

\noindent\textbf{Future work.}\enspace
Our findings suggest a hypothesis for a next-generation tabular
foundation model: combine \Mitra{}'s backbone (column invariance),
an early label-as-column slot, and a One-vs-All output head (class
invariance). Each ingredient is accuracy-neutral or better in
isolation. Two further follow-ups remain open: (i) the training data
itself, where varying the synthetic prior might isolate which
behaviors are properties of the architecture versus the data;
(ii) mechanism-grounded adversarial defenses, where hub-aware
context reweighting and robust quantile transforms can be evaluated under the
same paired protocol. Each direction
promises foundation models that are simultaneously more accurate,
more robust, and more interpretable than the current generation.

\bibliographystyle{plainnat}
\bibliography{neurips_2026}

\newpage
\appendix
\section{Experimental protocol}\label{app:protocol}

All experiments use 49 classification datasets (4 synthetic, 3 from
\texttt{sklearn}, and 42 from OpenML) and 10 regression datasets, listed
in Tab.~\ref{tab:datasets}.
We use 5 seeds for invariance experiments, the common 24-dataset attack
grid, and the readout-grid calibration table; 10 seeds for the broader
TabPFN-vs-XGBoost attack sweep; and 20 seeds for vote-decomposition
permutation tests. Exact seed and dataset counts are stated in the
relevant captions or subsections.

\noindent\textbf{Context/query partitions.}\enspace For real and OpenML classification datasets, each (dataset, seed) cell
starts from an 80/20 stratified context/query split seeded by the seed
index. Regression uses the same 80/20 split
without stratification.
Synthetic controls use their generator-defined train/test sizes.
Unless stated otherwise, a given
(dataset, seed) context/query partition is reused across the foundation
models, surrogate readouts, classical baselines, and attacks.

\subsection{Architectures}\label{app:protocol-arch}

This subsection specifies the released classification models audited in the paper, their parameter and shape accounting, the released versions used in each experiment, and the software environment.

\textbf{\TabPFN{}} \citep{hollmann2025tabpfn}.
A per-cell transformer (hidden width $D = 192$, 12 blocks) operating on the
$(n_\text{rows} + 1) \times n_\text{cols}$ table formed by the context set
plus the query row. Each block alternates row-wise attention (mixing across
context rows for a fixed column) with column-wise attention (mixing across
columns for a fixed row), interleaved with MLPs. Inputs are passed through
a fixed three-copy preprocessing pipeline (raw, scaled, quantile-rotated).
Numerical features are encoded via the per-feature positional matrix $W$
and bias $b$; class identity is read off the query cell at the output
head. For a single query and $n$ context rows with $d$ raw features, the
stack therefore carries $3(n{+}1)d$ cell tokens: row attention sees
$n{+}1$ entries in each of $3d$ feature streams, and column attention
sees $3d$ entries in each row.

\textbf{\TabICL{}} \citep{qu2025tabicl}.
A row-then-ICL transformer. In the accounting of
Table~\ref{tab:arch-accounting}, the model splits into three
column-embedding blocks, three pre-label row-interaction blocks
(``Row~3''), and then the 12-block row-level ICL transformer
(``ICL~12''). The first two stages process the table column by column
and emit a single vector per row, i.e.\ a per-row embedding matrix in
$\mathbb{R}^{(n_\text{context}+n_\text{query}) \times D_\text{row}}$;
the third column-embedding block is a SetTransformer readout that
compresses per-feature into per-sample vectors before the row stage.
The 12 ICL blocks then mix the context rows together with the query row
after labels are injected. RoPE is applied along the feature dimension
and circular feature grouping is applied to inputs. The class
prediction is read off the final ICL block's output for the query row.
In shape terms, the ColEmb stack starts from $d$ feature tokens per row,
compresses them to $(n_\text{context}+n_\text{query})$ row vectors by
ColEmb-2, and from that point onward both the Row~3 and ICL~12 stages
operate only on those row tokens.

\textbf{\Mitra{}} \citep{zhang2025mitra}.
A 12-block transformer trained with mixed synthetic priors and
released as part of AutoGluon. It differs from \TabPFN{} and
\TabICL{} along axes the main analysis identifies as load-bearing:
labels live in a \emph{separate} per-row token slot rather than
being added to feature tokens; the per-feature embedding is
\emph{shared} across columns; and the architecture has \emph{no
positional encoding} on either axis. Each block is an interleaved
row/feature \texttt{Layer}: a row-attention call lets each query
token attend to the context tokens for the same feature, and a
column-attention call mixes across features within a row. Both
released checkpoints (v1 and v1.1) share this 12-block structure;
they differ in their training mixtures. We load \Mitra{} as a single
model with AutoGluon's ensembling/stacking disabled, so the meta-ensemble
defaults that would otherwise wrap it are turned off and we measure
the model in isolation. Per-experiment \Mitra{} details
(invariance, layerwise probes, knockouts, readout fidelity, attacks,
collapse) appear in \sref{app:mitra}.

For \TabPFN{}, $7.08$M of $7.24$M parameters sit in the $12$-block
transformer stack; ``no thinking rows'' means this model does not
prepend any extra synthetic rows before the context/query table. For
\TabICL{}, the split is $0.88$M / $0.40$M / $26.28$M across the
column, row, and ICL stages. For \Mitra{}, $75.66$M of $75.67$M
parameters sit in the $12$-block interleaved row/column stack; the
remainder is the per-feature $1{\to}512$ embedding, the
$10$-class label embedding, and the final $512{\to}10$ head. The
released \TabPFN{} model used here is the $12$-block
PerFeatureTransformer, not the newer $24$-block v2.5/v2.6
architectures, which are released under licenses that are not
open-source compatible with this analysis pipeline and are therefore
out of scope.

\begin{table}[H]
\centering\scriptsize\setlength{\tabcolsep}{4pt}
\caption{Architecture accounting for the three main classification
models, derived from the released model files used throughout this
paper.}
\label{tab:arch-accounting}
\begin{tabular}{p{0.10\textwidth}cp{0.30\textwidth}p{0.10\textwidth}cp{0.20\textwidth}}
Model & Params & Core stack & Heads & FFN hidden & Normalization \\
\midrule
\TabPFN{} & 7.24M & 12 PerFeatureTransformer blocks, $d{=}192$, no thinking rows & 6 feature + 6 item & 768 & 3 bias-free LN / block \\
\TabICL{} & 27.55M & ColEmb 3 + Row 3 + ICL 12, $d_\text{row}{=}128$, $d_\text{ICL}{=}512$ & 8 / 8 / 8 & 256 / 256 / 1024 & pre-norm LN with bias \\
\Mitra{} & 75.67M & 12 interleaved row/column blocks, $d{=}512$, separate per-row label slot, no positional encoding & 4 row + 4 col & 2048 & 4 LN with bias / block \\
\end{tabular}
\end{table}

\begin{table}[H]
\centering\scriptsize\setlength{\tabcolsep}{4pt}
\caption{Shape and accounting proxy for the three audited classification
models on a single-query context set with $n$ context rows and $d$
raw features. The ``attention-work proxy'' counts attention-matrix
elements and ignores heads, MLPs, and constant factors; it is a
scaling summary, not an exact FLOPs count.}
\label{tab:arch-scaling-proxy}
\begin{tabular}{p{0.1\linewidth} p{0.27\linewidth} p{0.35\linewidth} p{0.17\linewidth}}
Model & Serialized state & Attention-work proxy & Measured cost \\
\midrule
\TabPFN{} &
$3(n{+}1)d$ cell tokens throughout the 12-block stack &
$3d(n{+}1)^2$ from row attention + $(n{+}1)(3d)^2$ from column attention,
per block &
$\sim 1.0$\,s, $\sim 3$\,GiB \\
\TabICL{} &
$d$ feature tokens / row until ColEmb-2, then $n{+}1$ row tokens
through Row~3 + ICL~12 &
$3(n{+}1)d^2$ in the ColEmb stage + $15(n{+}1)^2$ across Row~3 and ICL~12 &
$\sim 1.5$\,s, $\sim 4$\,GiB \\
\Mitra{} &
$(n{+}1)(d{+}1)$ tokens per layer ($d$ feature cells + $1$ label slot per row),
through the 12-block stack &
$d(n{+}1)^2$ from row attention + $(n{+}1)(d{+}1)^2$ from column attention,
per block &
--- \\
\end{tabular}
\end{table}

\noindent\textbf{Software environment.}\enspace All experiments run under Python and PyTorch, using the official
\TabPFN{} and \TabICL{} Python packages together with
\texttt{sklearn} for the classical baselines and probes.

\noindent\textbf{Categorical features and missingness.}\enspace Categorical columns from OpenML datasets (e.g.\ car,
tic-tac-toe, cmc, credit-approval, credit-g, cylinder-bands,
eucalyptus) are ordinal-encoded once with a fixed seed and
then passed identically to \TabPFN{}, \TabICL{}, Ridge, and XGBoost,
so that any model differential is not driven by encoding choice.
This equalises the inputs across baselines; it is not a benchmark of
native-categorical CatBoost or native-categorical XGBoost, which use
their own typed pathways.
Missing values are median-imputed for numerical columns and
mode-imputed for categorical columns at fetch time.
The post-encoding feature count $d$ in Tab.~\ref{tab:datasets}
reflects this encoding.

\subsection{Probes and separability measurements}\label{app:protocol-probes}\label{app:probes}

This subsection specifies the linear probe, its random-init control, the silhouette score, and the three geometry summaries used to characterize per-layer representations.

\textbf{Linear probe.} Given a frozen activation tensor $h \in \mathbb{R}^{n
\times D}$ at some layer, we fit a logistic-regression classifier
$\hat{y} = \arg\max_c W_c h$ on a labeled subset of the context rows and
report its accuracy on a held-out subset of the same dataset. Unless stated
otherwise, the probe is trained per (dataset, layer, seed) using
multiclass logistic regression with $\ell_2$ regularization $C{=}1$ and an
L-BFGS solver, on an 80/20 train/eval split of
the context rows under a per-seed shuffle. The maximum number of
solver iterations is raised to $2\,000$ to ensure convergence on the
higher-dimensional probe inputs (semeion $d{=}256$ and mfeat-pixel $d{=}240$).
We always report probe accuracy alongside two reference points: (i) the
per-dataset chance baseline (the majority-class frequency, mean
$\approx 0.50$ across our 49 classification datasets), and (ii) the model's
own end-to-end accuracy on the same (dataset, seed) pair. A probe accuracy
that approaches the second reference indicates that essentially all of the
model's classification-relevant information is already linearly readable
at that layer.

\textbf{Random-init control.} Wherever a probe accuracy is reported, we
also report the same probe trained on the activations of a copy of the model
whose transformer weights have been replaced by random initialisations.
This control isolates the part of separability that is due to the trained
weights as opposed to the architecture or the probe's own capacity.

\textbf{Silhouette score.} For a labeled point cloud $\{(h_i, y_i)\}$, the
silhouette of $h_i$ is $(b_i - a_i) / \max(a_i, b_i)$, where $a_i$ is the
mean distance from $h_i$ to other points of the same class and $b_i$ is the
mean distance from $h_i$ to points of the nearest \emph{other} class. The dataset-level silhouette score is the mean over $i$
and lies in $[-1, +1]$; values near $+1$ indicate clean per-class clusters.
We use cosine distance and report silhouette scores at every layer.

\textbf{Geometry summaries.} For a frozen activation matrix
$h \in \mathbb{R}^{n \times D}$ with mean-centered singular values
$\sigma_1 \geq \dots \geq \sigma_r > 0$, define normalized squared
spectrum $p_i = \sigma_i^2 / \sum_j \sigma_j^2$. We report three
standard summaries:
\begin{itemize}[leftmargin=1.5em,topsep=0pt,itemsep=0pt]
  \item \emph{Effective rank} $\exp\!\big({-\sum_i p_i \log p_i}\big)$:
    the exponential of the spectral entropy
    \citep{roy2007effective}.
    Equals $r$ for a flat spectrum and $1$ when one direction
    dominates; insensitive to absolute scale.
  \item \emph{Participation ratio (PR)}
    $\big(\sum_i \sigma_i^2\big)^2 \big/ \sum_i \sigma_i^4$: a heavier
    weighting toward the leading directions
    \citep{gao2017theory}. PR equals $r$ for a flat spectrum but drops
    much faster than effective rank when a few directions concentrate
    most of the variance.
  \item \emph{2NN intrinsic dimension (2NN-ID)}
    \citep{facco2017estimating}: estimated from the ratio of distances
    to the second and first nearest neighbour of each point. Captures
    the dimensionality of the local data manifold rather than of the
    global covariance, and is therefore typically smaller than
    effective rank.
\end{itemize}
Effective rank and PR are linear-algebraic measures of how the
activation \emph{covariance} is spread across directions; 2NN-ID is a
geometric measure of how the activation \emph{point cloud} fills its
ambient space. They agree qualitatively but probe different things,
which is why we report all three.

\subsection{Hardware and example costs}\label{app:wallclock}

This subsection specifies the compute node, the approximate budget of the headline GPU-bound runs, and the per-prediction inference cost on that node.

All experiments were run on a node with $4\times$
NVIDIA H100 NVL GPUs ($94$\,GiB each) and dual Intel Xeon
CPUs ($128$ physical cores); preprocessing and CPU-only
experiments use this CPU pool.

Table~\ref{tab:compute-ledger} gives the approximate cost of the
headline GPU-bound experiment families measured on this node; many
smaller CPU-bound or single-dataset sweeps are omitted because
they do not materially affect the total budget.

\begin{table}[H]
\centering\scriptsize\setlength{\tabcolsep}{4pt}
\caption{Approximate compute ledger for the headline GPU-bound runs in
the paper. Costs are measured on the $4\times$H100 node above and
rounded to the nearest half GPU-hour.}
\label{tab:compute-ledger}
\begin{tabular}{p{0.43\linewidth} p{0.20\linewidth} p{0.23\linewidth}}
Experiment family & Grid & Approx. cost \\
\midrule
Full inference-only sweep with probes & $49$ cls $\times 10$ seeds, all layers, sharded over $4$ GPUs & $\sim 2$ GPU-hours total \\
Frozen probe after knockout & $49 \times 10$, $4$ KO points $\times 8$ probe positions & $\sim 6$ GPU-hours \\
Vote decomposition with $1000$ row-permutation null & $49$ cls $\times 20$ seeds & $\sim 4$ GPU-hours \\
Activation patching & clean/corrupted pair enumeration on $49$ cls & $\sim 18$ GPU-hours \\
\midrule
Headline subtotal & --- & $\sim 30$ GPU-hours \\
\end{tabular}
\end{table}

\noindent\textbf{Per-prediction inference cost.}\enspace On the same hardware, cold end-to-end inference for a single query
is $\sim 1.0$\,s for \TabPFN{},
$\sim 1.5$\,s for \TabICL{}, and $\sim 2.0$\,s for \TabICL{}-reg
(numbers include preprocessing). Peak GPU memory for these calls is
$\sim 3$\,GiB, $\sim 4$\,GiB, and $\sim 5$\,GiB respectively, well
within a single H100. The OvA wrapper of \S\ref{sec:inv-mechanism}
multiplies these by the class count $C$: no measured accuracy penalty
for binary
classification, $\sim 10$\,s and $\sim 30$\,GiB for the ten-class
mfeat-* family. Zeroing $W$ in \TabPFN{} and removing RoPE
in \TabICL{} are zero-overhead edits to fixed parameters and do
not change wall-clock or memory.

All experiments, except for \TabICLvOne{} are inference-only on released models.

\subsection{Seeds, dataset subsets, and XGBoost tuning}\label{app:reproduce-detail}

This subsection specifies the seed convention, the enumerated dataset subsets used by individual experiments, the attack-grid accounting, the calibration metric definitions, the XGBoost reference configuration, and the full attack hyperparameter values.

\noindent\textbf{Random seeds.}\enspace Whenever an experiment uses $N$ seeds we use the integer seeds
$0,1,\dots,N{-}1$; the $20$-seed permutation null uses
$0,\dots,19$. Each (dataset, seed) cell is therefore
deterministic given the software environment above. We disable
TF32 matmul for both backbones and set deterministic cuDNN
algorithms; forward passes through the released models match across
consecutive runs to within $10^{-6}$ in logits on H100.

\noindent\textbf{Explicit subset compositions.}\enspace The compute-intensive experiments are run on enumerated subsets of
the full $49$-classification grid; the exact lists are:
\begin{itemize}[leftmargin=1.5em,topsep=0pt,itemsep=0pt]
  \item \emph{The $24$-classification attack grid}
    (\S\ref{sec:attacks}), defined as the intersection of the
    \TabPFN{} and \TabICL{} attack runs:
    \texttt{LED-display-domain-7digit}, \texttt{MiceProtein},
    \texttt{balance-scale}, \texttt{banknote-authentication},
    \texttt{blood-transfusion-service-center}, \texttt{breast-w},
    \texttt{breast\_cancer}, \texttt{car},
    \texttt{climate-model-simulation-crashes},
    \texttt{cylinder-bands}, \texttt{dresses-sales},
    \texttt{eucalyptus}, \texttt{iris}, \texttt{mfeat-factors},
    \texttt{mfeat-fourier}, \texttt{mfeat-karhunen},
    \texttt{mfeat-morphological}, \texttt{mfeat-pixel},
    \texttt{quadrant\_2d}, \texttt{random\_labels}, \texttt{sign\_1d},
    \texttt{steel-plates-fault}, \texttt{wine}, \texttt{xor\_2d}.
  \item \emph{The $16$-dataset mixed-type clean slice}
    (Tab.~\ref{tab:catboost-mixed16}, \S\ref{app:attacks-catboost}):
    \texttt{dresses-sales}, \texttt{car},
    \texttt{cylinder-bands}, \texttt{eucalyptus}, \texttt{cmc},
    \texttt{credit-approval}, \texttt{credit-g},
    \texttt{analcatdata\_dmft}, \texttt{anneal},
    \texttt{Fitness\_Club}, \texttt{Is-this-a-good-customer},
    \texttt{qsar-biodeg}, \texttt{website\_phishing},
    \texttt{MIC}, \texttt{tic-tac-toe}, \texttt{ilpd}.
  \item \emph{The mixed16 attack slice}
    (Tab.~\ref{tab:mixed16-attacks}, \S\ref{app:attacks-mixed16}):
    the same $16$ datasets as the mixed-type clean slice above, but only
    the three highest-signal attacks (hub poison, rank warp, SVD burial)
    under the shared ordinalised numeric view.
  \item \emph{The $10$-classification head-resolved \TabPFN{} vote
    subset} (\S\ref{app:vote-head-layer}): \texttt{balance-scale},
    \texttt{breast\_cancer}, \texttt{credit-g}, \texttt{diabetes},
    \texttt{ilpd}, \texttt{iris}, \texttt{mfeat-zernike},
    \texttt{tic-tac-toe}, \texttt{vehicle}, \texttt{wine}.
  \item \emph{The per-block MLP knockout subset}
    (Tab.~\ref{tab:tabpfn-blockko}, \S\ref{sec:critical-block}):
    \texttt{xor\_2d}, \texttt{quadrant\_2d}, \texttt{sign\_1d},
    \texttt{random\_labels}, \texttt{balance-scale},
    \texttt{banknote-authentication}, \texttt{breast-w},
    \texttt{iris}, \texttt{wine}, \texttt{vehicle}, and
    \texttt{steel-plates-fault}. This fixed subset keeps the four
    synthetic controls and adds binary, three-class, four-class, and
    seven-class real tasks so the mean drop table is auditable from the
    PDF alone.
\end{itemize}

\noindent\textbf{Attack accounting.}\enspace The attack results use two distinct reporting grids. The matched
cross-model comparison in Table~\ref{tab:attacks-drops},
Table~\ref{tab:attacks-tabicl}, Table~\ref{tab:attacks-mlp}, and
Table~\ref{tab:catboost-clean24} uses the common 24-classification
grid with 5 seeds per dataset (per-cell numbers may differ by up to
${\sim}0.2$\,pp across the three tables because they were produced
in independent runs over the same grid). The broader TabPFN-vs-XGBoost appendix
comparison in Table~\ref{tab:attacks-xgb} uses 49 classification
datasets $\times$ 10 seeds, except \textsc{mono\_softexp} where one
dataset is dropped for numerical degeneracy and $n{=}48$. The null-space
PGD diagnostic in \S\ref{app:attack-nullspace} uses the 29 datasets with
sufficient null-space dimension after the explicit exclusions in
\S\ref{app:attack-nullspace}. The readout-grid calibration table
(\S\ref{app:E8}) uses 49 classification datasets $\times$ 5 seeds.

\noindent\textbf{Calibration metrics.}\enspace Expected calibration error (ECE) in Table~\ref{tab:E8-readout-grid} is
computed per (dataset, seed) with 15 equal-width confidence bins and
then averaged over seeds within dataset; negative log-likelihood (NLL)
is the multiclass log loss on the same held-out query split. Unless
stated otherwise, the readout-grid table reports means over the 49
per-dataset seed means.

\noindent\textbf{XGBoost reference.}\enspace The XGBoost baseline used throughout (Tab.~\ref{tab:attacks-drops},
Apps.~\ref{app:attacks-table}, \ref{app:attacks-tabicl},
\ref{app:attacks-ridge}) uses XGBoost 2.1.4 with
$5$-fold stratified-CV grid search over number of trees
$\in\{200, 500, 1000\}$, maximum depth
$\in\{3, 6, 9\}$, learning rate $\in\{0.03, 0.1\}$, and
row subsample $\in\{0.7, 1.0\}$, with early stopping (patience $50$)
on a held-out $20\%$ validation split. Categorical columns
share the same ordinal encoding used for \TabPFN{}/\TabICL{}; the
native-categorical XGBoost pathway is not used here so that the
encoding is identical across baselines. The selected configuration
is then refit on the full context set and evaluated on the same
query partition each foundation model sees.
The broader mixed16 clean-only native-categorical XGBoost control in
Tab.~\ref{tab:catboost-mixed16} uses the same package version with
the native-categorical pathway enabled; it is validation-tuned on the
same $20\%$ held-out split over depths $\in\{4,6\}$, learning rates
$\in\{0.03,0.10\}$, and number of trees $\in\{300,600\}$.

\noindent\textbf{Attack hyperparameters.}\enspace \begin{table}[H]
\centering\scriptsize\setlength{\tabcolsep}{3pt}
\caption{Complete hyperparameters for the eight attacks in
Table~\ref{tab:attacks-drops}, fixed before any model evaluation.}
\label{tab:attack-hparams}
\begin{tabular}{lll}
Attack & Parameter(s) & Value \\
\midrule
Noise pad        & $\sigma$ multiplier of feature std; pad fraction & $4\times$; $20\%$ of $|S|$ \\
Hub poison       & hub fraction; centrality metric                  & top $15\%$; rep-cosine kNN degree at L9 \\
Centroid inj     & injected counter-class examples per class        & $\max(3, |S|/100)$ \\
Boundary poison  & boundary margin (fraction of inter-centroid dist.) & $0.20$ \\
Mono cube        & exponent of signed cube root warp                 & $1/3$ \\
Mono softexp     & soft-exponential temperature $\tau$               & $1.0$ \\
Mono rank        & per-column rank transform                         & deterministic, ties broken by index \\
SVD hide         & null-space cutoff $\gamma$                        & $10^{-6}\cdot\sigma_{\max}$ \\
Nullspace PGD    & step $\eta$, iters $T$, budget $\varepsilon$      & $0.01$, $200$, $1.0$ (in feature std units) \\
\end{tabular}
\end{table}

\subsection{Statistical reporting protocol}\label{app:stats-protocol}

This subsection specifies the reporting unit, the bootstrap procedure, the permutation null, and the multiple-comparison correction used throughout the paper.

\begin{itemize}
  \item \emph{Primary reporting unit = dataset} for real-data benchmark
    tables. Whenever a metric is computed at the (dataset, seed) level,
    we first average across seeds within each dataset and then average
    across datasets. Because every dataset uses the same number of
    seeds, these values equal the corresponding dataset $\times$ seed
    grand means printed in some appendix tables.
  \item \emph{Bootstrap CIs on benchmark deltas} resample the vector of
    per-dataset paired differences after averaging seeds within dataset.
    This convention is used for
    vote decomposition, frozen probes, attacks, and transplantation
    (\S\ref{sec:transplant}; \S\ref{app:K4}).
  \item \emph{Single-dataset synthetic sweeps} (e.g.\ the K8 collapse
    $f$-sweep) use seed as the reporting unit and state this explicitly;
    their CIs are paired bootstraps over the seedwise difference vector.
  \item \emph{Permutation $z$} for vote decomposition: $1\,000$
    row-permutation nulls per dataset $\times$ 20 seeds.
  \item \emph{Multiple comparisons}: Bonferroni applied to the 10
    regression tests; classification leaderboards report raw
    per-dataset numbers without family-wise correction (we report $n$
    explicitly).
\end{itemize}

\subsection{Dataset inventory}\label{app:datasets}

This subsection lists every classification and regression dataset used in the paper, together with the exact OpenML task and dataset identifiers and the asset-provenance record.

\begin{table}[H]
\caption{All 49 classification (top) and 10 regression (bottom)
datasets. $n$ is the raw row count; $d$ is the post-encoding feature count; $K$ is the number of classes. Source codes: S = synthetic, K = \texttt{sklearn}, O = OpenML.}\label{tab:datasets}\centering
\small\setlength{\tabcolsep}{3pt}\resizebox{\linewidth}{!}{%
\begin{tabular}{ll rrr @{\hspace{6pt}} ll rrr @{\hspace{6pt}} ll rrr}
src & name & $n$ & $d$ & $K$ & src & name & $n$ & $d$ & $K$ & src & name & $n$ & $d$ & $K$ \\\midrule
O & \texttt{Fitness\_Club} & 1500 & 6 & 2 & O & \texttt{credit-g} & 1000 & 20 & 2 & O & \texttt{pc3} & 1563 & 37 & 2 \\
O & \texttt{Is-this-a-good-customer} & 1723 & 13 & 2 & O & \texttt{cylinder-bands} & 540 & 37 & 2 & O & \texttt{pc4} & 1458 & 37 & 2 \\
O & \texttt{LED-display-domain-7digit} & 500 & 7 & 10 & O & \texttt{diabetes} & 768 & 8 & 2 & O & \texttt{qsar-biodeg} & 1054 & 41 & 2 \\
O & \texttt{MIC} & 1699 & 111 & 8 & O & \texttt{dresses-sales} & 500 & 12 & 2 & S & \texttt{quadrant\_2d} & 250 & 5 & 4 \\
O & \texttt{MiceProtein} & 1080 & 77 & 8 & O & \texttt{eucalyptus} & 736 & 19 & 5 & S & \texttt{random\_labels} & 150 & 5 & 2 \\
O & \texttt{analcatdata\_authorship} & 841 & 70 & 4 & O & \texttt{hill-valley} & 1212 & 100 & 2 & O & \texttt{red\_wine} & 1599 & 11 & 6 \\
O & \texttt{analcatdata\_dmft} & 797 & 4 & 6 & O & \texttt{ilpd} & 583 & 10 & 2 & O & \texttt{semeion} & 1593 & 256 & 10 \\
O & \texttt{anneal} & 898 & 38 & 5 & K & \texttt{iris} & 150 & 4 & 3 & S & \texttt{sign\_1d} & 150 & 5 & 2 \\
O & \texttt{balance-scale} & 625 & 4 & 3 & O & \texttt{kc2} & 522 & 21 & 2 & O & \texttt{steel-plates-fault} & 1941 & 33 & 2 \\
O & \texttt{banknote-authentication} & 1372 & 4 & 2 & O & \texttt{maternal\_health\_risk} & 1014 & 6 & 3 & O & \texttt{tic-tac-toe} & 958 & 9 & 2 \\
O & \texttt{blood-transfusion-service-center} & 748 & 4 & 2 & O & \texttt{mfeat-factors} & 2000 & 216 & 10 & O & \texttt{vehicle} & 846 & 18 & 4 \\
O & \texttt{breast-w} & 699 & 9 & 2 & O & \texttt{mfeat-fourier} & 2000 & 76 & 10 & O & \texttt{wdbc} & 569 & 30 & 2 \\
K & \texttt{breast\_cancer} & 569 & 30 & 2 & O & \texttt{mfeat-karhunen} & 2000 & 64 & 10 & O & \texttt{website\_phishing} & 1353 & 9 & 3 \\
O & \texttt{car} & 1728 & 6 & 4 & O & \texttt{mfeat-morphological} & 2000 & 6 & 10 & K & \texttt{wine} & 178 & 13 & 3 \\
O & \texttt{climate-model-simulation-crashes} & 540 & 20 & 2 & O & \texttt{mfeat-pixel} & 2000 & 240 & 10 & S & \texttt{xor\_2d} & 250 & 5 & 2 \\
O & \texttt{cmc} & 1473 & 9 & 3 & O & \texttt{mfeat-zernike} & 2000 & 47 & 10 & & & & & \\
O & \texttt{credit-approval} & 690 & 15 & 2 & O & \texttt{pc1} & 1109 & 21 & 2 & & & & & \\
\midrule
\multicolumn{15}{l}{\emph{10 regression datasets (all OpenML; $K$ omitted):}} \\
src & name & $n$ & $d$ & & src & name & $n$ & $d$ & & src & name & $n$ & $d$ & \\
O & \texttt{Another-Dataset-on-used-Fiat-500} & 1538 & 7 & & O & \texttt{cars} & 804 & 17 & & O & \texttt{forest\_fires} & 517 & 12 & \\
O & \texttt{Moneyball} & 1232 & 14 & & O & \texttt{concrete\_compressive\_strength} & 1030 & 8 & & O & \texttt{healthcare\_insurance\_expenses} & 1338 & 6 & \\
O & \texttt{QSAR\_fish\_toxicity} & 907 & 6 & & O & \texttt{energy\_efficiency} & 768 & 8 & & O & \texttt{socmob} & 1156 & 5 & \\
O & \texttt{airfoil\_self\_noise} & 1503 & 5 & & & & & & & & & & & \\
\end{tabular}}
\end{table}

\noindent\textbf{Exact OpenML identifiers.}\enspace For the OpenML assets in Table~\ref{tab:datasets}, \texttt{task\_id}
fixes the supervised task and \texttt{dataset\_id} fixes the local
dataset version used in our cache.

\begin{table}[H]
\centering\scriptsize\setlength{\tabcolsep}{4pt}
\caption{Exact OpenML identifiers for the 42 classification datasets in
Table~\ref{tab:datasets}.}
\label{tab:openml-cls-ids}
\resizebox{\linewidth}{!}{%
\begin{tabular}{lrr @{\hspace{8pt}} lrr @{\hspace{8pt}} lrr}
Name & task\_id & dataset\_id & Name & task\_id & dataset\_id & Name & task\_id & dataset\_id \\
\midrule
\texttt{Fitness\_Club} & 363671 & 46927 & \texttt{cmc} & 23 & 23 & \texttt{mfeat-morphological} & 18 & 18 \\
\texttt{Is-this-a-good-customer} & 363682 & 46938 & \texttt{credit-approval} & 29 & 29 & \texttt{mfeat-pixel} & 146824 & 40979 \\
\texttt{LED-display-domain-7digit} & 125921 & 40496 & \texttt{credit-g} & 31 & 31 & \texttt{mfeat-zernike} & 22 & 22 \\
\texttt{MIC} & 363711 & 46980 & \texttt{cylinder-bands} & 14954 & 6332 & \texttt{pc1} & 3918 & 1068 \\
\texttt{MiceProtein} & 146800 & 40966 & \texttt{diabetes} & 363629 & 46921 & \texttt{pc3} & 3903 & 1050 \\
\texttt{analcatdata\_authorship} & 3549 & 458 & \texttt{dresses-sales} & 125920 & 23381 & \texttt{pc4} & 3902 & 1049 \\
\texttt{analcatdata\_dmft} & 3560 & 469 & \texttt{eucalyptus} & 2079 & 188 & \texttt{qsar-biodeg} & 363696 & 46952 \\
\texttt{anneal} & 363614 & 46906 & \texttt{hill-valley} & 9970 & 1479 & \texttt{red\_wine} & 361250 & 44972 \\
\texttt{balance-scale} & 11 & 11 & \texttt{ilpd} & 9971 & 1480 & \texttt{semeion} & 9964 & 1501 \\
\texttt{banknote-authentication} & 10093 & 1462 & \texttt{kc2} & 3913 & 1063 & \texttt{steel-plates-fault} & 146817 & 40982 \\
\texttt{blood-transfusion-service-center} & 10101 & 1464 & \texttt{maternal\_health\_risk} & 363685 & 46941 & \texttt{tic-tac-toe} & 49 & 50 \\
\texttt{breast-w} & 15 & 15 & \texttt{mfeat-factors} & 12 & 12 & \texttt{vehicle} & 53 & 54 \\
\texttt{car} & 146821 & 40975 & \texttt{mfeat-fourier} & 14 & 14 & \texttt{wdbc} & 9946 & 1510 \\
\texttt{climate-model-simulation-crashes} & 146819 & 40994 & \texttt{mfeat-karhunen} & 16 & 16 & \texttt{website\_phishing} & 363707 & 46963 \\
\end{tabular}}
\end{table}

\begin{table}[H]
\centering\scriptsize\setlength{\tabcolsep}{4pt}
\caption{Exact OpenML identifiers for the 10 regression datasets in
Table~\ref{tab:datasets}.}
\label{tab:openml-reg-ids}
\begin{tabular}{lrr @{\hspace{10pt}} lrr}
Name & task\_id & dataset\_id & Name & task\_id & dataset\_id \\
\midrule
\texttt{Another-Dataset-on-used-Fiat-500} & 363615 & 46907 & \texttt{concrete\_compressive\_strength} & 361237 & 44959 \\
\texttt{Moneyball} & 361616 & 41021 & \texttt{energy\_efficiency} & 361617 & 44960 \\
\texttt{QSAR\_fish\_toxicity} & 361621 & 44970 & \texttt{forest\_fires} & 361618 & 44962 \\
\texttt{airfoil\_self\_noise} & 361235 & 44957 & \texttt{healthcare\_insurance\_expenses} & 363675 & 46931 \\
\texttt{cars} & 361622 & 44994 & \texttt{socmob} & 361264 & 44987 \\
\end{tabular}
\end{table}

\noindent\textbf{Asset provenance.}\enspace \label{app:assets}
Table~\ref{tab:asset-status} records the released artifacts used.

\begin{table}[H]
\centering\scriptsize
\caption{Asset provenance.}
\label{tab:asset-status}
\begin{tabular}{p{0.16\linewidth} p{0.32\linewidth} p{0.40\linewidth}}
Asset class & Identifier in this PDF & Terms / license status \\
\midrule
OpenML datasets & Tables~\ref{tab:openml-cls-ids} and~\ref{tab:openml-reg-ids} & Upstream terms vary by dataset; we use the OpenML task and dataset IDs as published. \\
scikit-learn datasets & Table~\ref{tab:datasets} & Inherited from the scikit-learn distribution. \\
Released models & TabPFN v2 (release 7.0) and TabICL (release 2.0.3); see \S\ref{app:protocol-arch} & Used as published; upstream terms apply. \\
\end{tabular}
\end{table}

\section{TabPFN classification: full results}
\label{app:tabpfn-cls}

The full results for \TabPFN{} classification cover where class-readable
structure forms (\sref{app:tabpfn-cls-where}), what the readout computes
at $\mathrm{L}9$ (\sref{app:tabpfn-cls-what}), and the collapse analysis
built around the M0--M3 hand-crafted models
(\sref{app:tabpfn-cls-collapse}).

% =============================================================
\subsection{Where class-readable structure forms}
\label{app:tabpfn-cls-where}

\begin{figure}[t]
  \centering
  \includegraphics[width=0.95\linewidth]{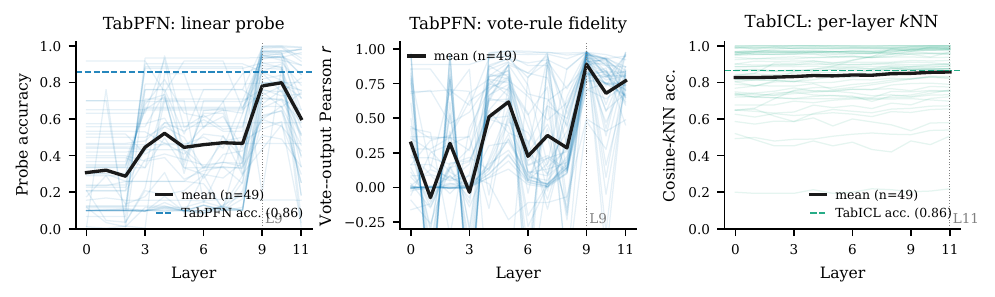}
  \caption{Per-layer profiles on the $49$-dataset classification
  suite. Thin lines: seed-averaged per-dataset curves; thick black
  lines: means. \textbf{Left:} \TabPFN{} linear-probe accuracy
  converges late with a sharp jump at
  $\mathrm{L}8\!\to\!\mathrm{L}9$. \textbf{Middle:} Pearson correlation
  between an attention-weighted vote read off layer $L$ and \TabPFN{}'s
  predicted probabilities; the vote rule (\sref{sec:tabpfn-mech})
  becomes faithful only at $\mathrm{L}9$. \textbf{Right:} \TabICL{}
  cosine-$k$NN accuracy on per-block representations runs high
  almost from the input.}
  \label{fig:layer-traj}
\end{figure}

\subsubsection{Probe convergence}
\label{app:probe-convergence}
A logistic-regression probe trained per (dataset, layer, seed) on
context-row activations and then averaged within dataset over $10$
seeds gives the left panel of Figure~\ref{fig:layer-traj}. Across all
$49$ classification datasets, mean accuracy stays near chance through
$\mathrm{L}0$--$\mathrm{L}8$ and then jumps from $0.467$ to $0.780$
between $\mathrm{L}8$ and $\mathrm{L}9$
(Table~\ref{tab:tabpfn-probe-perlayer}, Figure~\ref{fig:layer-traj}).
The smaller $\mathrm{L}9 \to \mathrm{L}10$ gain is a late refinement,
not a second phase transition; the sharp $\mathrm{L}8 \to \mathrm{L}9$
break is the crystallization point used throughout the main text.
Silhouette scores rise in lockstep.

\begin{table}[H]
\centering
\small
\caption{Linear-probe accuracy at each TabPFN block, mean and std across
the same $49$ classification datasets shown in Figure~\ref{fig:layer-traj},
after averaging $10$ seeds within dataset. The sharp +$0.31$ jump from
$\mathrm{L}8$ to $\mathrm{L}9$ marks the onset of the late-readout
regime used throughout the paper; $\mathrm{L}10$ is a smaller
follow-on refinement.}
\label{tab:tabpfn-probe-perlayer}
\setlength{\tabcolsep}{4pt}
\resizebox{\linewidth}{!}{%
\begin{tabular}{lrrrrrrrrrrrrr}
Block & $-1$ & 0 & 1 & 2 & 3 & 4 & 5 & 6 & 7 & 8 & 9 & 10 & 11 \\
\midrule
mean acc & $0.394$ & $0.307$ & $0.320$ & $0.287$ & $0.446$ & $0.522$ & $0.445$ & $0.460$ & $0.470$ & $0.467$ & $0.780$ & $\mathbf{0.798}$ & $0.602$ \\
std       & $0.271$ & $0.209$ & $0.218$ & $0.210$ & $0.271$ & $0.277$ & $0.266$ & $0.264$ & $0.265$ & $0.246$ & $0.198$ & $0.173$ & $0.282$ \\
\end{tabular}}
\end{table}

\subsubsection{Block knockouts (retrained probes)}
\label{app:block-ko}
Zeroing each MLP sub-block in turn and retraining the linear probe at
$\mathrm{L}9$ isolates which blocks the readable structure depends on.

\begin{table}[H]
\centering
\small
\caption{TabPFN per-block MLP knockout: mean accuracy drop (pp) across
$n{=}11$ fixed classification datasets (the four synthetic controls plus
\texttt{balance-scale}, \texttt{banknote-authentication},
\texttt{breast-w}, \texttt{iris}, \texttt{wine}, \texttt{vehicle}, and
\texttt{steel-plates-fault}), baseline $0.907$. Block~$0$ dominates;
blocks~$5$--$6$ are a clear secondary cluster; the remaining blocks are
individually replaceable.}
\label{tab:tabpfn-blockko}
\setlength{\tabcolsep}{4pt}
\resizebox{\linewidth}{!}{%
\begin{tabular}{lrrrrrrrrrrrr}
Block & 0 & 1 & 2 & 3 & 4 & 5 & 6 & 7 & 8 & 9 & 10 & 11 \\
\midrule
$\Delta$ acc (pp) & $-48.1$ & $-0.3$ & $-0.9$ & $-1.6$ & $-0.9$ & $-20.0$ & $-19.6$ & $-8.9$ & $-2.9$ & $-6.3$ & $-1.5$ & $-2.9$ \\
std (pp) & $16.8$ & $0.8$ & $1.1$ & $1.6$ & $1.5$ & $15.6$ & $20.1$ & $13.6$ & $3.6$ & $7.2$ & $2.6$ & $4.2$ \\
\end{tabular}}
\end{table}

Block~$0$ is the most damaging on $38$ of $49$ datasets (probe drop
$\approx 40$\,pp). A retrained probe still recovers labels at lower
accuracy, so block~$0$ sets up the coordinate frame that the later
blocks expect rather than destroying class information outright.

\subsubsection{Frozen probes after knockout (causality)}
\label{app:frozen-probe}
Frozen probes isolate the causal contribution of each block.
Probes trained at $\ell \in \{-1, 0, 3, 5, 7, 9, 10, 11\}$ on the
intact model are frozen, then evaluated under KO@$\{0,5,9\}$ on $49$
datasets and $10$ seeds.

\begin{table}[H]
\caption{Frozen-probe accuracy under block knockout. Probes trained on clean model, evaluated without retraining after zeroing block~0 (KO@0) or block~9 (KO@9). Mean and std across 49 datasets $\times$ 10 seeds.}
\label{tab:frozen-probe-ko}
\centering
\footnotesize
\begin{tabular}{lrrrrrr}
& \multicolumn{2}{c}{\textbf{Baseline}} & \multicolumn{2}{c}{\textbf{KO@0}} & \multicolumn{2}{c}{\textbf{KO@9}} \\
\textbf{Layer} & Mean & Std & Mean & Std & Mean & Std \\
\midrule
Input & 0.394 & 0.274 & 0.394 & 0.274 & 0.394 & 0.274 \\
L0 & 0.307 & 0.211 & 0.292 & 0.214 & 0.307 & 0.211 \\
L3 & 0.446 & 0.274 & 0.415 & 0.253 & 0.446 & 0.274 \\
L5 & 0.445 & 0.269 & 0.418 & 0.263 & 0.445 & 0.269 \\
L7 & 0.470 & 0.268 & 0.313 & 0.171 & 0.470 & 0.268 \\
L9 & 0.780 & 0.201 & 0.404 & 0.226 & 0.451 & 0.279 \\
L10 & 0.798 & 0.175 & 0.499 & 0.255 & 0.496 & 0.241 \\
L11 & 0.602 & 0.284 & 0.321 & 0.212 & 0.515 & 0.275 \\
\end{tabular}
\end{table}

KO@$9$ drops $\mathrm{L}9$ frozen-probe accuracy from $0.780$ to $0.451$
($\Delta = -0.329$); KO@$0$ cascades to $0.404$ ($\Delta = -0.376$). Two
roles: block~$0$ sets the frame, block~$9$ writes the readable structure.

\subsubsection{Per-layer rank profile and intrinsic dimension}
\label{app:rank-profile}

Effective rank, participation ratio (PR), and the two-NN intrinsic
dimension of \citet{facco2017estimating}, averaged over $49$
classification datasets and $5$ seeds (\TabPFN{}-cls).
\TabPFN{} bottlenecks early: rank and PR drop sharply at block~$2$ and
recover only partially.
The residual rank at the readout scales with the number of classes
(mean L11 effective rank $21.6$ for $C \leq 3$ vs.\ $31.7$ for $C \geq 4$;
Pearson $r = +0.64$; \sref{app:E13}). This is structurally different
from the identical-representation collapse mode of \citet{qu2025tabicl},
tested in \sref{app:tabpfn-cls-collapse}.

\begin{table}[H]
\centering
\small
\caption{TabPFN per-block geometry, mean over $49$ datasets $\times\,5$ seeds.
Block~$-1$ is the input row before any feature mixing.}
\label{tab:tabpfn-rank-profile}
\begin{tabular}{rrrr}
Block & Effective rank & PR & 2NN-ID \\
\midrule
$-1$ &  $0.00$ &  $0.00$ & $0.00$ \\
$0$  & $40.37$ &  $2.53$ & $4.62$ \\
$1$  & $41.97$ &  $1.87$ & $8.76$ \\
$2$  & $10.20$ &  $1.47$ & $4.52$ \\
$3$  & $19.99$ &  $2.13$ & $5.76$ \\
$4$  & $22.51$ &  $2.01$ & $5.87$ \\
$5$  & $21.69$ &  $1.88$ & $5.25$ \\
$6$  & $23.92$ &  $1.82$ & $5.60$ \\
$7$  & $20.56$ &  $1.87$ & $5.33$ \\
$8$  & $16.15$ &  $1.99$ & $4.86$ \\
$9$  & $22.81$ &  $2.80$ & $4.41$ \\
$10$ & $25.28$ &  $2.98$ & $4.48$ \\
$11$ & $25.21$ &  $3.25$ & $4.77$ \\
\end{tabular}
\end{table}

% =============================================================
\subsection{What the readout computes at \texorpdfstring{$\mathrm{L}9$}{L9}}
\label{app:tabpfn-cls-what}

\subsubsection{Row-attention sharpness}
\label{app:row-attn}
$\mathrm{L}9$ is the sharpest layer on all $49$ datasets: mean max
weight $0.925$ (std $0.078$), mean entropy $1.756$ versus $3.126$ at
$\mathrm{L}0$. IQR of max weights is $[0.89, 0.97]$; only two datasets
fall below $0.80$.

\subsubsection{Target-column attention vs.\ similarity}
\label{app:col-attn}
$\mathrm{L}9$ attention tracks input similarity but is not reducible to
it. The Spearman rank correlation between the $\mathrm{L}9$ attention
weight on each context row (restricted to the target column) and
input-space $\ell_2$ similarity is positive on $45$ of $49$ datasets
(median $r = 0.32$, IQR $[0.20, 0.48]$, tail to $r > 0.70$). The
representation-distance regression of \sref{app:E1}, summarized in
Table~\ref{tab:tabpfn-attn-r2}, sharpens this picture: input distance
alone explains only a small fraction of the $\mathrm{L}9$ attention
pattern, $\mathrm{L}8$ representation distance more than doubles that
variance, and the full learned-similarity model nearly triples it.

\begin{table}[H]
\centering
\small
\caption{$R^2$ of three predictors of $\mathrm{L}9$ row-attention,
across $49$ classification datasets ($5$ seeds each).
Input $\ell_2$: linear regression on raw input distance only.
Rep $\ell_2$ (L8): the same with L8 representation distance.
Full: learned bilinear similarity from L8. Median, IQR, mean, and number
of datasets exceeding $R^2 = 0.5$.}
\label{tab:tabpfn-attn-r2}
\setlength{\tabcolsep}{4pt}
\begin{tabular}{lrrrrr}
Predictor & median & q25 & q75 & mean & $n(R^2{>}0.5)/49$ \\
\midrule
Input $\ell_2$       & $0.168$ & $0.051$ & $0.345$ & $0.218$ & $2$ \\
Rep $\ell_2$ (L8)    & $0.488$ & $0.339$ & $0.627$ & $0.477$ & $21$ \\
Full learned sim.    & $0.606$ & $0.487$ & $0.728$ & $0.600$ & $35$ \\
\end{tabular}
\end{table}

\subsubsection{Vote decomposition (all 49 datasets)}
\label{app:vote-full}
Vote is
\[
  \hat{p}(c) = \frac{1}{H}\sum_{h=1}^{H}\sum_{i=1}^{N_{\text{train}}}
                \alpha^{(h)}_i\, \mathbf{1}[y_i = c],
\]
with $\alpha^{(h)}_i$ the $\mathrm{L}9$ attention weight from the query
to context row $i$ in head $h$. Evaluated on $49$ datasets ($20$ seeds
each); the random-labels condition is a negative control.

\begin{table}[H]
\caption{Vote decomposition: attention-weighted label vote at L9 vs.\ final output. Pearson $r$, Spearman $\rho$, and null-shuffle $z$-score per dataset (20 seeds each). Mean across all 49 datasets: Pearson $r = 0.890 \pm 0.114$, Spearman $\rho = 0.849 \pm 0.138$, $z = 6.79 \pm 3.79$.}
\label{tab:vote-full49}
\centering
\scriptsize
\begin{tabular}{lrrr}
\textbf{Dataset} & \textbf{Pearson $r$} & \textbf{Spearman $\rho$} & \textbf{$z$-score} \\
\midrule
Fitness\_Club & 0.982 & 0.989 & 7.32 \\
Is-this-a-good-customer & 0.976 & 0.992 & 1.46 \\
LED-display-domain-7digit & 0.820 & 0.839 & 9.82 \\
MIC & 0.974 & 0.727 & 8.95 \\
MiceProtein & 0.799 & 0.665 & 11.58 \\
analcatdata\_authorship & 0.960 & 0.737 & 10.70 \\
analcatdata\_dmft & 0.495 & 0.451 & 1.99 \\
anneal & 0.935 & 0.774 & 12.90 \\
balance-scale & 0.919 & 0.888 & 6.52 \\
banknote-authentication & 0.971 & 0.807 & 2.44 \\
blood-transfusion-service-center & 0.963 & 0.977 & 5.13 \\
breast-w & 0.984 & 0.955 & 3.81 \\
breast\_cancer & 0.978 & 0.938 & 3.53 \\
car & 0.913 & 0.900 & 13.51 \\
climate-model-simulation-crashes & 0.976 & 0.979 & 6.09 \\
cmc & 0.900 & 0.861 & 6.74 \\
credit-approval & 0.966 & 0.957 & 2.60 \\
credit-g & 0.971 & 0.976 & 6.36 \\
cylinder-bands & 0.896 & 0.918 & 4.49 \\
diabetes & 0.983 & 0.991 & 6.95 \\
\multicolumn{4}{l}{\emph{(First 20 of 49 datasets)}} \\
\end{tabular}
\end{table}

Population means: Pearson $r = 0.888 \pm 0.114$, Spearman $\rho = 0.848
\pm 0.139$, $z = 6.88 \pm 3.77$ ($1\,000$ row-permutation nulls). Worst
cases are ten-class digit datasets (mfeat-zernike $r = 0.71$,
mfeat-factors $0.71$, semeion $0.76$): vote distributes mass over
class indicators while accuracy uses an $\arg\max$. The optimal
head-reweighting NNLS score gives an upper bound on what a fixed
convex recombination of head votes buys.

\textbf{Multi-class fidelity scaling.}
Per-dataset vote-fidelity $\overline{r}$ decreases monotonically with
class count $C$ across the $49$-dataset population: linear regression
of $\overline{r}$ on $C$ yields slope $-0.016$ per class
(intercept $0.957$, Pearson $-0.43$, Spearman $\rho = -0.60$,
$p < 10^{-5}$, $n=49$), with binary-class mean $0.93$, mid-class
($3{\le}C{\le}8$) mean $0.88$, and ten-class mean $0.80$. The
ten-class digit cluster ($r{\in}[0.71,0.76]$) lies within the residual
spread of this trend: the multi-class gap reflects the vote rule's
finite per-class indicator capacity rather than a mechanism switch
in \TabPFN{}.

\subsubsection{Head- and layer-resolved vote}
\label{app:vote-head-layer}\label{app:K5}

\begin{table}[H]
\caption{Per-head vote Pearson $r$ at L9. Each head's vote is computed independently; mean across heads shown in last column. Results show no single head dominates the vote signal (10 datasets $\times$ 20 seeds).}
\label{tab:vote-head-l9}
\centering
\footnotesize
\begin{tabular}{lrrrrrrr}
\textbf{Dataset} & \textbf{H0} & \textbf{H1} & \textbf{H2} & \textbf{H3} & \textbf{H4} & \textbf{H5} & \textbf{Mean} \\
\midrule
balance-scale & 0.937 & 0.795 & 0.872 & 0.935 & 0.762 & 0.883 & 0.864 \\
breast\_cancer & 0.988 & 0.166 & 0.889 & 0.968 & 0.600 & 0.979 & 0.765 \\
credit-g & 0.951 & -0.432 & 0.615 & 0.910 & 0.820 & 0.938 & 0.634 \\
diabetes & 0.977 & 0.592 & 0.735 & 0.975 & -0.400 & 0.973 & 0.642 \\
ilpd & 0.982 & 0.754 & 0.413 & 0.988 & -0.624 & 0.805 & 0.553 \\
iris & 0.990 & 0.843 & 0.968 & 0.987 & -0.085 & 0.968 & 0.779 \\
mfeat-zernike & 0.818 & 0.403 & 0.526 & 0.750 & 0.052 & 0.642 & 0.532 \\
tic-tac-toe & 0.820 & -0.113 & 0.772 & 0.780 & 0.307 & 0.699 & 0.544 \\
vehicle & 0.887 & 0.149 & 0.760 & 0.873 & 0.109 & 0.779 & 0.593 \\
wine & 0.973 & 0.921 & 0.961 & 0.974 & 0.353 & 0.950 & 0.855 \\
\end{tabular}
\end{table}

The six $\mathrm{L}9$ heads span $0.05$--$0.10$ Pearson units around the
head-mean. The layer profile is flat through $\mathrm{L}8$, jumps at
$\mathrm{L}9$, and is sustained through $\mathrm{L}11$. The largest
single-head zeroing drop is $3.9$\,pp at $\mathrm{L}9$ (\sref{app:E12});
the $\mathrm{L}9$ peak is real and distributed across heads.

\subsubsection{Causal attention forcing}
\label{app:causal-attn}
Replacing $\mathrm{L}9$ softmax weights with $1/N_{\text{train}}$
collapses accuracy from $0.874$ to $0.488$ ($\Delta = -0.386$),
matching the majority-class baseline ($0.498$, \sref{app:E4}). MLP
knockout at any block also collapses accuracy to the same vicinity
(\sref{app:tabpfn-falsif}): $\mathrm{L}9$ attention is one part of a
working circuit, not the unique site of computation.

\subsubsection{$k$NN agreement vs.\ TabPFN}
\label{app:knn-vs-tabpfn}
Agreement between \TabPFN{}'s prediction and $k$NN under input $\ell_2$,
$\mathrm{L}9$ rep $\ell_2$, $\mathrm{L}9$ rep cosine, and $\mathrm{L}9$
Mahalanobis, $k \in \{1,3,5\}$, on $49$ datasets and $10$ seeds
(Table~\ref{tab:tabpfn-knn-agree}, top). Input-space $k$NN reaches
$0.78$--$0.81$ across $k$; representation-space $\ell_2$/cosine $k$NN at
$\mathrm{L}9$ reaches only $0.54$--$0.56$, and Mahalanobis is worse
($0.37$--$0.41$). Strengthening the test by reading from $\mathrm{L}11$
(K1) does not change the picture (Table~\ref{tab:tabpfn-knn-agree},
middle). Even the strongest learned-metric variants (NCA-trained $\ell_2$,
shrunk Mahalanobis, soft-$k$NN with a fitted temperature, all at L9)
max out around $0.84$ (Table~\ref{tab:tabpfn-knn-agree}, bottom). A
1-hidden-layer GELU MLP head fit on the frozen final representation
(\sref{app:adapt-mlphead}) also fails to recover native accuracy on
\TabPFN{} ($-24.9$\,pp paired, $95\%$ bootstrap CI $[-32.5,-18.0]$,
worse on $38/42$ shared datasets) while almost recovering it on
\TabICL{} ($-0.8$\,pp, $[-1.5,-0.3]$): a non-linear head does not
close the asymmetry.
The model is not a $k$NN in its own representation. The accurate
framing is the learned similarity of \sref{app:E1}, where rep-$\ell_2$
at $\mathrm{L}8$ explains $R^2 \approx 0.48$ of the $\mathrm{L}9$
attention pattern.

\begin{table}[H]
\centering
\small
\caption{Agreement between TabPFN's argmax prediction and a $k$NN baseline
under several distance metrics, mean across $49$ datasets. Top: L9
representations. Middle: L11 representations
(\sref{app:K1}). Bottom: learned-metric variants at L9. Every
representation-space variant agrees with TabPFN on a strict minority of
test points; only soft-$k$NN with a fitted temperature crosses $0.8$.}
\label{tab:tabpfn-knn-agree}
\setlength{\tabcolsep}{4pt}
\begin{tabular}{lrrr}
Metric & $k=1$ & $k=3$ & $k=5$ \\
\midrule
\multicolumn{4}{l}{\emph{Input space and L9 representations (49 datasets)}}\\
Input $\ell_2$           & $0.781$ & $0.801$ & $0.813$ \\
L9 rep $\ell_2$          & $0.540$ & $0.556$ & $0.561$ \\
L9 rep cosine            & $0.540$ & $0.556$ & $0.562$ \\
L9 Mahalanobis           & $0.367$ & $0.393$ & $0.414$ \\
\midrule
\multicolumn{4}{l}{\emph{L11 representations (K1, 49 datasets)}}\\
L11 rep $\ell_2$         & $0.528$ & $0.554$ & $0.561$ \\
L11 rep cosine           & $0.528$ & $0.554$ & $0.561$ \\
L11 Mahalanobis          & $0.356$ & $0.378$ & $0.395$ \\
\midrule
\multicolumn{4}{l}{\emph{Learned-metric L9 variants (L1, 49 datasets)}}\\
NCA-trained $\ell_2$     & $0.723$ & $0.734$ & $0.742$ \\
Shrunk Mahalanobis       & $0.632$ & $0.683$ & $0.717$ \\
Soft-$k$NN ($\ell_2$, fitted $T$)  & $0.837$ & $0.837$ & $0.837$ \\
Soft-$k$NN (cos, fitted $T$)       & $0.813$ & $0.813$ & $0.813$ \\
\end{tabular}
\end{table}

\subsubsection{Activation patching}
\label{app:activation-patching}
Activation patching localizes the sites that carry the corruption.
\TabPFN{} blocks $\{0, 5, 9\}$ and \TabICL{} blocks $\{$ColEmb-$2$,
ICL-$0$, ICL-$6$, ICL-$11\}$ are patched on $10$ datasets and $5$
seeds, under block-$0$ KO and label-shuffle corruptions
(Table~\ref{tab:tabpfn-actpatch}). Every TabPFN block patch and every
TabICL ICL-block patch recovers to baseline (recovery $= 1.00$). The
single exception is \TabICL{}'s ColEmb-$2$ under label-shuffle (recovery
$\approx 0.04$), consistent with ColEmb-$2$'s role as the compression
bottleneck (\sref{sec:critical-block}). TabICL ICL blocks are
unaffected by upstream block-$0$ KO because the column-embedding stack
runs before the ICL stack and absorbs the perturbation, so corrupt and
baseline are equal there (recovery is by definition $1.00$).

\begin{table}[H]
\centering
\small
\caption{Activation patching: replace activations at the named site in a
corrupt run with those from a clean run, then read out accuracy. Mean
across $10$ datasets, $5$ seeds. \emph{Recovery} $=
(\text{patched}-\text{corrupt})/(\text{baseline}-\text{corrupt})$;
$1.00$ means the patched site fully restores baseline accuracy.}
\label{tab:tabpfn-actpatch}
\setlength{\tabcolsep}{4pt}
\begin{tabular}{llrrrr}
Site & Corruption & baseline & corrupt & patched & recovery \\
\midrule
TabPFN B0       & label-shuffle & $0.874$ & $0.497$ & $0.874$ & $1.000$ \\
TabPFN B5       & label-shuffle & $0.874$ & $0.497$ & $0.874$ & $1.000$ \\
TabPFN B9       & label-shuffle & $0.874$ & $0.497$ & $0.874$ & $1.000$ \\
TabICL ICL-0    & label-shuffle & $0.891$ & $0.502$ & $0.891$ & $1.000$ \\
TabICL ICL-6    & label-shuffle & $0.891$ & $0.502$ & $0.891$ & $1.000$ \\
TabICL ICL-11   & label-shuffle & $0.891$ & $0.502$ & $0.891$ & $1.000$ \\
TabICL ColEmb-2 & label-shuffle & $0.891$ & $0.502$ & $0.538$ & $0.043$ \\
\midrule
TabPFN B0       & block-0 KO    & $0.874$ & $0.443$ & $0.874$ & $1.000$ \\
TabPFN B5       & block-0 KO    & $0.874$ & $0.443$ & $0.874$ & $1.000$ \\
TabPFN B9       & block-0 KO    & $0.874$ & $0.443$ & $0.874$ & $1.000$ \\
TabICL ICL-0    & block-0 KO    & $0.891$ & $0.892$ & $0.891$ & $1.000$ \\
TabICL ICL-6    & block-0 KO    & $0.891$ & $0.892$ & $0.891$ & $1.000$ \\
TabICL ICL-11   & block-0 KO    & $0.891$ & $0.892$ & $0.891$ & $1.000$ \\
\end{tabular}
\end{table}

% =============================================================
\subsection{Representation collapse: stress tests, screen, and
hand-crafted models}
\label{app:tabpfn-cls-collapse}

\subsubsection{Stress test on collapse-prone data}
\label{app:repr-collapse-new}

\begin{figure}[t]
  \centering
  \begin{minipage}[t]{0.42\linewidth}
    \vspace{0pt}
    \caption{\textbf{Where representation collapse appears in the
      public models.} Balance-scale accuracy ($20$ seeds, mean
      $\pm$ s.d.) for each architecture (blue) and under
      symmetry-restoring ablations: removing \TabPFN{}'s positional
      matrix or \TabICL{}'s RoPE produces no measurable mean
      accuracy loss on this benchmark (green); disabling
      \TabPFN{}'s pair channel or TabPFN v1's RoPE reproduces the predicted
      collapse (red).}
    \label{fig:no-collapse}
  \end{minipage}\hfill
  \begin{minipage}[t]{0.55\linewidth}
    \vspace{0pt}
    \centering
    \includegraphics[width=\linewidth]{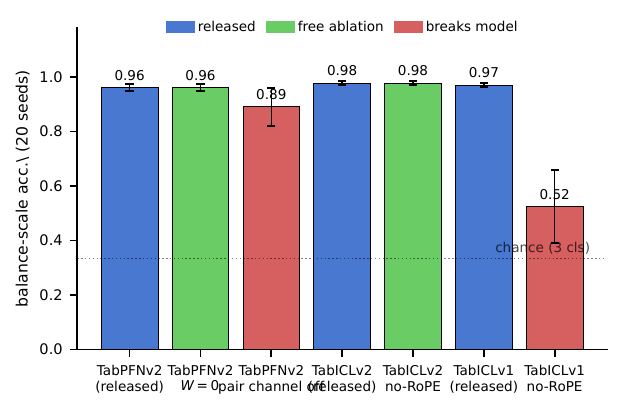}
  \end{minipage}
\end{figure}

Identical column marginals are the canonical setting in which
\TabPFN{} and \TabICL{} risk producing indistinguishable row
representations (cf.\ \citealp{qu2025tabicl}). The stress test runs on
three datasets: balance-scale; a synthetic identical-marginal benchmark
($n = 1000$, $m = 20$, fraction $f \in \{0, 0.25, 0.5, 0.75, 1.0\}$ of
columns drawn i.i.d.\ from the same $5$-level discrete distribution,
label a fixed nonlinear function of a labeled subset); and post-hoc
real-data candidates from the JS-divergence screen of \sref{app:K7}.
The measured quantities are pairwise pre-readout cosine similarity,
collapse rate (cross-class pairs with similarity $> 0.99$), and
KMeans-$C$ purity over $5$ seeds. The conditions are: \TabICLvOne{}
(full vs.\ no-RoPE); \TabICL{} (full vs.\ no-RoPE vs.\
grouping-disabled); \TabPFN{} (full vs.\ $W = 0$ vs.\ per-group
feature count fixed to $1$); and random-init controls. The full
$f$-sweep with 10-seed paired-bootstrap CIs is in \sref{app:K8}.

\subsubsection{Marginal-similarity vs.\ invariance-cost correlation}
\label{app:repr-collapse-acc}
Marginal similarity predicts the accuracy cost of removing the
symmetry-breaker. For each of $49$ datasets, pairwise JS divergence
between column marginals identifies the maximum identical-marginal
column set via thresholded clustering, and the accuracy gap between
the full model and a strict permutation-invariant variant (\TabPFN{}
$W = 0$, \TabICLvOne{} no-RoPE, \TabICL{} no-RoPE) gives the
invariance cost. The prediction is a positive Spearman $\rho$ on
configurations whose ablation removes the symmetry-breaker, and
$\approx 0$ otherwise. Companion $\rho$ at JS tolerances $\tau \in
\{0.01, 0.05, 0.10\}$ are reported in \sref{app:K7}.

\subsubsection{Hand-crafted models: setup}
\label{app:repr-collapse-handcrafted}
The hand-crafted M0--M3 models referenced in
\sref{sec:repr-collapse-handcrafted} factor each architecture's
symmetry-breaking devices into a minimal single-layer setting.

\begin{figure}[H]
  \centering
  \begin{minipage}[t]{0.46\linewidth}
    \vspace{0pt}
    \caption{\textbf{Hand-crafted models that isolate the three
      symmetry-breaking devices.} M0--M3 are single attention layers
      over $m{=}3$ binary features (M0: shared-cell mean-pool; M1:
      column identifier; M2: per-column attention mask; M3: pair
      tokens). Transductive accuracy on Task~A ($y{=}x_1$), B
      ($y{=}x_1\!\oplus\!x_2$), C ($y{=}\mathrm{maj}(x)$).
      M0 is capped at the $0.75$ multiset bound; M1, M2, M3 reach
      $1.0$ on A; only M3 (pair tokens) breaks the bound on B; all
      four solve C. Derivations below.}
    \label{fig:handcrafted}
  \end{minipage}\hfill
  \begin{minipage}[t]{0.50\linewidth}
    \vspace{0pt}
    \centering
    \includegraphics[width=\linewidth]{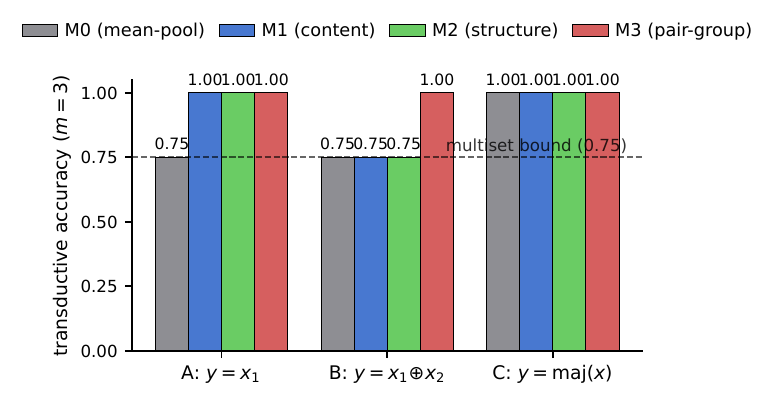}
  \end{minipage}
\end{figure}

The construction isolates the three symmetry-breaking devices
(column identifier, column mask, pair grouping) on a task family
where column-marginal information alone is useless. Let $x \in
\{0,1\}^m$ and enumerate all $2^m$ rows. Three tasks:
\begin{itemize}
  \item Task A: $y = x_1$ (one column is the label).
  \item Task B: $y = x_1 \oplus x_2$ (two-column XOR).
  \item Task C: $y = \text{majority}(x_1, \dots, x_m)$ (multiset-decidable
        control).
\end{itemize}
All features have identical Bernoulli($1/2$) marginals. Each model is a
single attention layer plus hard-argmax readout; transductive
accuracy on the full $2^m$-row enumeration with $m = 3$. Numbers are
exact rationals derived analytically and reproduced numerically;
Table~\ref{tab:handcraft-verify} records the verification.

\subsubsection{The multiset bound (Lemma 1)}
\label{app:repr-collapse-multiset}

\textbf{Lemma 1 (multiset bound, general orbit-majority form).}
\emph{Let $r : \{0,1\}^m \to \mathcal{R}$ be any row representation
invariant to all column permutations: $r(x) = r(\sigma x)$ for every
$\sigma \in S_m$. Let $g : \mathcal{R} \to \{0,1\}$ be any
deterministic readout, and let $y : \{0,1\}^m \to \{0,1\}$ be any
deterministic binary label rule. Then transductive accuracy of
$g \circ r$ on the $2^m$-row enumeration is at most}
\[
  \frac{1}{2^m}\sum_{O \in \{0,1\}^m / S_m}
        \max_{y_0\in\{0,1\}}
        \bigl|\{x \in O : y(x) = y_0\}\bigr|,
\]
\emph{i.e.\ the orbit-wise majority of the label rule. The bound
reduces to the formula previously stated for Task A
($y(x) = x_1$); for $m = 3$ binary targets it equals $0.75$ and is
exactly attained by $\mathrm{M0}$.}

\emph{Proof sketch.} $S_m$-invariance partitions $\{0,1\}^m$ into orbits
on which $r$, and hence any deterministic readout $g \circ r$, is
constant. Within an orbit $O$ the readout cannot do better than the
majority of $\{y(x) : x \in O\}$, contributing
$\max_{y_0} |\{x \in O : y(x) = y_0\}|$ correct predictions. Summing
over the four $m=3$ orbits $\{000\}, \{001,010,100\},
\{011,101,110\}, \{111\}$ for $y(x) = x_1$ gives $1+2+2+1 =
6/8 = 0.75$. \hfill$\square$

\textbf{Proposition 2 (Task B corollary).}
\label{app:repr-collapse-multiset-prop}
\emph{Any model with column-permutation-invariant row representation is
capped at $0.75$ on Task B for $m = 3$, even with an optimal linear
readout}: orbit $\{001,010,100\}$ has XOR values $\{1,1,0\}$ and
$\{011,101,110\}$ has $\{1,0,1\}$, so orbit-wise majority again caps at
$6/8$. The route assignment in
Section~\ref{sec:repr-collapse-handcrafted} maps the M3 (pair-grouped)
construction to the device that breaks this bound; M3 solves Task B at
$m=3$ and is the smallest sufficient construction exhibited here.

\subsubsection{M0--M3 model definitions}

\textbf{M0 (shared embedding + mean-pool).}
\label{app:handcrafted-m0}
$\phi(b) = b$; $r(x) = \frac{1}{m}\sum_j x_j$; hard-NN readout. $r$ is
$S_m$-invariant. On Task A with $m = 3$ each three-row orbit ties two
labels against one, costing one error per orbit; transductive accuracy
$6/8 = 0.75$, exactly meeting the bound.

\textbf{M1 (cell tokens with column identifier).}
\label{app:handcrafted-m1}
\TabPFN{}-style: token $t_{ij} = [v_{x_{ij}};\, e_j;\, \ell_{y_i}]$ with
column identifier $e_j$. With $W_Q = W_K$ extracting (value, column)
and cell-level scores $s = \mathbf{1}[\text{value match}] +
\mathbf{1}[\text{column match}]$, argmax-with-ties readout followed
by per-row mean over the $m$ cells gives $\hat{p}(100) = (1/3, 2/3)$.
Task A accuracy $1.0$. Task B unsolved (single-cell attention cannot
condition on two columns jointly).

\textbf{M2 (column-stream attention without identifiers).}
\label{app:handcrafted-m2}
\TabICLvOne{}-style: bipolar shared embedding $\phi(b) = 2b - 1$,
per-column heads $W_Q^{(j)} = W_K^{(j)} = e_j^\top$ (mask = column
structure), per-row mean over $m$ heads. M2 = M1 on this family:
restricting M1 attention to matching columns and M2's per-column head
yield identical softmaxes, and per-row means coincide. Task A $1.0$,
Task B $0.5$.

\textbf{M3 (pair-grouped tokens).}
\label{app:handcrafted-m3}
v2-grouping: each super-cell over an unordered feature pair $(j, k)$
carries the joint value $u_{(j,k)} = 2x_j + x_k \in \{0,1,2,3\}$ and a
pair identifier $e_{(j,k)}$. On Task A with $m = 3$, the super-cell at
$(x_1, x_2) = (1, 0)$ appears in five positives and one negative, so
$\hat{p}(100) = (1/6, 5/6)$ and Task A accuracy is $1.0$. Conditioning
on the joint value of $(x_1, x_2)$ separates XOR; Task B accuracy
$1.0$.

\subsubsection{Verification table and architecture mapping}

\begin{table}[H]
\centering
\caption{Transductive accuracy of the four handcrafted models on Tasks
A, B, C with $m=3$, verified numerically.}
\label{tab:handcraft-verify}
\begin{tabular}{lccc}
Model           & Task A & Task B & Task C \\
\midrule
M0 na\"{i}ve    & $0.75$ & $0.75$ & $1.00$ \\
M1 TabPFN-style & $1.00$ & $0.50$ & $1.00$ \\
M2 TabICL-style & $1.00$ & $0.50$ & $1.00$ \\
M3 pair-grouped & $1.00$ & $1.00$ & $1.00$ \\
\end{tabular}
\end{table}

\begin{table}[H]
\centering
\caption{Route assignment per architecture. Each model carries two of the
three routes (M1: cell+col-ID; M2: column-stream; M3: pair-group); ablating
one route leaves the other intact, predicting the small empirical ablation
costs measured in \S\ref{sec:invariance}.}
\label{tab:route-architecture}
\begin{tabular}{lll}
Architecture & Routes & Missing route \\
\midrule
TabPFN v2  & M1 $+$ M3              & column-stream (M2) \\
TabICL v1  & M2 $+$ within-row RoPE & pair-group (M3) \\
TabICL v2  & M2 $+$ M3              & cell+col-ID (M1) \\
Mitra      & shared embedding $+$ no PE & --- (column-invariant by construction) \\
\end{tabular}
\end{table}

Mapping: \TabPFN{} pairs M1 (random per-feature identifiers) with M3
(pair grouping); \TabICLvOne{} pairs M2 (column-stream attention) with
RoPE; \TabICL{} pairs M2 with M3 (circular grouping) and adds
target-aware embeddings. Each v2 architecture carries \emph{two} of the
three routes; \TabICLvOne{} carries only one. This predicts the empirical
pattern of \sref{sec:repr-collapse-doubleablation}: \TabPFN{} with $W = 0$
and \TabICL{} without RoPE both show no measurable accuracy loss, while disabling
\TabPFN{}'s pair channel costs $-7$\,pp on balance-scale and $-14$\,pp
under full identical-marginal stress (\sref{app:K8}); \TabICLvOne{}
without RoPE loses $-44$\,pp on balance-scale and $15$\,pp on the
benchmark mean. The load-bearing v2 mitigation is the M3
pair/circular-grouping route.

\subsubsection{Attention-mode ablation}
A sweep over shared-embedding, oracle-attention, and attention-mode
combinations (row-level ICL, cell-level cross-attention, column-stream)
on Tasks A--C under leave-one-out separates the contributions of each
mechanism. Row-level ICL on collapsed representations caps at $0.625$
on Task A LOO; cell-level cross-attention without column identifiers
drops to $0.25$; column-stream reaches $1.0$ on Task A even with
shared embeddings. Task B remains at chance for every single-layer
mode, consistent with XOR requiring at least one nonlinear
interaction.

\subsubsection{Fresh non-linear head on the frozen final representation}
\label{app:adapt-mlphead}
A 1-hidden-layer GELU MLP head ($d_{\text{hidden}}$ matched to the
original decoder's first hidden width: $192$ for \TabPFN{}, $512$ for
\TabICL{}) is retrained on each backbone's frozen final representation,
using the context split for fitting and the held-out query split for
accuracy (LBFGS / AdamW; protocol matches \sref{app:probes}). The
linear-head row of \sref{sec:transplant} is the same protocol with
$d_{\text{hidden}}{=}0$ (a bare logistic layer). Across the $42$
shared datasets where both methods completed, the linear head reaches
mean $0.582$ on \TabPFN{} ($-26.5$\,pp vs.\ native, $95\%$ paired
bootstrap CI $[-34.7,-19.1]$, worse on $39/42$) and the MLP head
reaches $0.598$ ($-24.9$\,pp, $[-32.5,-18.0]$, worse on $38/42$,
exact two-sided binomial sign test $p\!\approx\!3{\times}10^{-8}$);
on \TabICL{} the same heads reach $0.849$ ($-0.7$\,pp, $[-1.2,-0.3]$)
and $0.847$ ($-0.8$\,pp, $[-1.5,-0.3]$). The non-linear head gains
roughly $1.6$\,pp on \TabPFN{} relative to the linear head while
leaving a $\sim25$\,pp gap to native, and is statistically
indistinguishable from the linear head on \TabICL{}. The fresh-head
asymmetry between the two backbones survives the addition of a
single-hidden-layer non-linearity.

\section{TabICL classification: full results}\label{app:tabicl}\label{app:tabicl-cls}

This appendix collects the full \TabICL{} classification evidence
behind the main-text claims of \sref{sec:repr} and \sref{sec:tabicl-mech}:
column-embedding ablations, ICL-block knockouts and uniform-attention
sweeps, label perturbations, the per-layer linear and $k$NN probes,
and the readout grid that compares the end-to-end head against
per-layer prototype, $k$NN, and per-block-vote decoders, including a
cross-version comparison over the released TabICL v1, v1.1, and v2
classifiers.

% =============================================================
\subsection{Where the class-readable representation is laid down}
\label{app:tabicl-cls-where}

\subsubsection{ColEmbedding knockouts}\label{app:tabicl-colemb}
Zeroing each ColEmbedding block (ColEmb-0/1/2) on $49$ datasets: ColEmb-2
is most damaging on $45$ of $49$. ColEmb-2 is the SetTransformer readout
that compresses per-feature into per-sample vectors.

\begin{table}[H]
\centering
\small
\caption{TabICL ColEmbedding-block knockouts: mean accuracy drop across
$49$ classification datasets, baseline $0.86$.}
\label{tab:tabicl-colemb-ko}
\begin{tabular}{lrr}
Knockout & $\Delta$ acc & most-damaging datasets \\
\midrule
ColEmb-0 & $-0.03$ & 4 / 49 \\
ColEmb-1 & $-0.01$ & 0 / 49 \\
ColEmb-2 & $-0.28$ & 45 / 49 \\
\end{tabular}
\end{table}

\noindent\textbf{ColEmb-2 sets the coordinate frame.}\enspace \label{app:tabicl-colemb-frame}
To mirror the TabPFN block-0/block-9 analysis (\sref{app:frozen-probe})
we evaluate two probe types under each early-block knockout: a
\emph{frozen} probe trained on the intact model and applied without
retraining, and a \emph{retrained} probe fit from scratch on the
post-knockout train representations. Probes are read off at L$-1$
(input to the ICL stack) and L11 (final ICL block); $49$ classification
datasets, $5$ seeds.
Knocking out ColEmb-2 collapses the final accuracy by $28$\,pp and
the frozen probe at L$-1$ by $26$\,pp, but the retrained probe at L$-1$
loses only $6$\,pp ($0.812\!\to\!0.752$, well above the multi-class
chance level). Class information survives the perturbation; the downstream ICL
stack cannot read the post-KO frame. At L11 the retrained probe
($0.602$) sits below the frozen probe ($0.704$) under ColEmb-2 KO,
so the late representation no longer organizes class structure
linearly even when refit. ColEmb-0 and ColEmb-1 produce the same
qualitative pattern (frozen $\!\ll\!$ retrained at L$-1$) at much
smaller magnitude, consistent with their role as upstream contributors
to the same coordinate-setting role. Knocking out a single ICL block
(L0, L5, or L11) leaves both probe types within $3$\,pp of baseline,
matching the redundancy reported in \sref{sec:tabicl-mech}.

\begin{table}[H]
\centering\small\setlength{\tabcolsep}{4pt}
\caption{TabICL frozen-vs-retrained probe under each early-block knockout, mean across $49$ classification datasets $\times\,5$ seeds. Frozen probes are trained on the intact model and applied without retraining; retrained probes are fit from scratch on post-knockout train representations. L-1 is the input to the ICL stack (output of ColEmb+RowInteraction); L11 is the final ICL block.}
\label{tab:tabicl-colemb-frame}
\begin{tabular}{lcccccc}
 & & & \multicolumn{2}{c}{\textbf{Frozen probe}} & \multicolumn{2}{c}{\textbf{Retrained probe}} \\
\cmidrule(lr){4-5} \cmidrule(lr){6-7}
\textbf{Knockout} & \textbf{Final acc} & \textbf{$\Delta$} & L$-1$ & L11 & L$-1$ & L11 \\
\midrule
Baseline (intact) & 0.864 & --- & 0.812 & 0.859 & 0.812 & 0.859 \\
\midrule
ColEmb-0 & 0.836 & $-0.028$ & 0.583 & 0.827 & 0.787 & 0.824 \\
ColEmb-1 & 0.849 & $-0.014$ & 0.655 & 0.843 & 0.806 & 0.843 \\
ColEmb-2 & 0.583 & $-0.281$ & 0.555 & 0.704 & 0.752 & 0.602 \\
\midrule
ICL-L0 & 0.860 & $-0.004$ & 0.812 & 0.854 & 0.812 & 0.852 \\
ICL-L5 & 0.845 & $-0.018$ & 0.812 & 0.830 & 0.812 & 0.827 \\
ICL-L11 & 0.863 & $-0.001$ & 0.812 & 0.855 & 0.812 & 0.859 \\
\end{tabular}
\end{table}

\subsubsection{RowInteraction skips}\label{app:tabicl-rowint}
Skipping each RowInteraction block individually leaves the ICL-input
linear probe at $81\%$, essentially unchanged from the intact
representation: RowInteraction does not create the class structure.
Two further checks corroborate this. The per-layer $\ell_2$-$k$NN
curve of Tab.~\ref{tab:tabicl-perlayer-knn} already sits within
$4$\,pp of its L11 endpoint at L0, immediately after the first
row-interaction attention. The uniform-attention combinations in
\sref{app:tabicl-uniform-all} cost $5$--$13$\,pp on average when
applied to contiguous halves of the stack. Both align with the
main-text claim that removing between-row mixing in the
column-embedding network costs only a few points.

\subsubsection{ICL input probe by dataset}\label{app:tabicl-probes}
We measure class-readability per layer with $\ell_2$-$k$NN agreement
on the representations rather than with geometry summaries: the
TabICL stack does not bottleneck (the rank/PR profile in
\sref{sec:rank-profile} is monotone-increasing through most of the
stack) so geometry alone is uninformative about \emph{when} class
structure becomes readable, while $k$NN agreement directly tracks the
quantity the prototype/$k$NN readout (\sref{sec:tabicl-mech}) consumes.
Across $49$ datasets the kNN-on-representation accuracy is already
$0.824$ at the output of the first ICL block (L0), within $4$\,pp of
end-to-end ($0.864$); the curve gains $\le 0.04$ over the entire
stack.

\begin{table}[H]
\centering\small\setlength{\tabcolsep}{4pt}
\caption{Per-layer $\ell_2$-$k$NN on \TabICL{} representations across $49$
datasets ($k{=}5$, mean$\,\pm\,$std across datasets). \TabICL{} baseline
$0.864 \pm 0.153$. Class structure is laid down by L0 and only mildly
sharpened.}
\label{tab:tabicl-perlayer-knn}
\begin{tabular}{lcccccccccccc}
Layer & L0 & L1 & L2 & L3 & L4 & L5 & L6 & L7 & L8 & L9 & L10 & L11 \\
\midrule
$k$NN acc  & .824 & .825 & .827 & .829 & .835 & .832 & .837 & .837 & .847 & .849 & .855 & .856 \\
agreement  & .885 & .885 & .890 & .886 & .897 & .894 & .901 & .902 & .923 & .924 & .935 & .936 \\
\end{tabular}
\end{table}

Per-dataset peak-layer (argmax of $k$NN accuracy across L0--L11): median
peak L10; only $5/49$ datasets peak at L0 and $1/49$ at L1, while $27/49$
peak at L10 or L11. Class structure is laid down by L0 (within $4$\,pp
of end-to-end on average) and only mildly sharpened by the rest of the
stack, consistent with the qualitative claim in the main text.

% =============================================================
\subsection{Attention and label interventions}
\label{app:tabicl-cls-interv}

\subsubsection{Label interventions}\label{app:tabicl-labels}
Five context-set label perturbations are applied at inference on the
$49$-dataset classification suite. Tab.~\ref{tab:tabicl-labels}
reports the per-dataset change in \TabICL{} accuracy and in the
linear probe of the ICL representation after each perturbation:
labels are essential, their specific learned directions matter, and
they act as class identities rather than positional indices.

\begin{table}[H]
\centering\small\setlength{\tabcolsep}{6pt}
\caption{Label-perturbation effects on \TabICL{}-cls. $\Delta$ acc is
the change in end-to-end accuracy; probe is the ICL-representation
linear probe accuracy after the perturbation.}
\label{tab:tabicl-labels}
\begin{tabular}{lrll}
Intervention             & $\Delta$ acc      & probe              & note \\
\midrule
Remove (zero)            & $-44$\,pp         & $0.81 {\to} 0.12$  & labels are essential \\
Cyclic shift $k{\to}k{+}1$ & $-0.01$\,pp (med.) & unchanged       & $41/49$ within $\pm 1$\,pp \\
Two-class swap           & $-76$\,pp$^\dagger$ & ---              & $^\dagger$on swapped classes only \\
Random Gaussian          & $\to$ chance       & collapsed         & learned directions matter \\
Shuffle within context   & $\to$ chance       & collapsed         & order-agnostic role broken \\
\end{tabular}
\end{table}

\subsubsection{Single-block uniform attention}
\label{app:tabicl-uniform-single}
Single-block and two-block uniform row-attention leaves \TabICL{}
within a few points of baseline. We force uniform row-attention at one
ICL block (block 0 or block 11) or at one sliding two-block window
(blocks $X$, $X{+}1$ for $X{=}0\dots10$). Across $49$ datasets and $10$
seeds, no single block and no two-block window costs more than a few
points on average; at least $46/49$ datasets stay within $|\Delta| < 1$\,pp
at every window. The matching \TabPFN{} L9 uniform-attention drop is
$-0.39$ (\sref{app:causal-attn}).

\begin{table}[H]
\centering\small\setlength{\tabcolsep}{4pt}
\caption{Single-block (top) and sliding two-block (bottom) uniform
row-attention drops on \TabICL{}, $49$ datasets, $10$ seeds. Drop is
baseline${-}$accuracy; the rightmost column counts datasets with
$|\Delta|{<}0.01$.}
\label{tab:tabicl-uniform-single}
\begin{tabular}{lrrrr}
Block(s)              & mean drop & median & worst & n($\le 1$pp)/49 \\
\midrule
b0                    & $+0.023$ & $+0.009$ & $+0.180$ & 39 \\
b11                   & $+0.002$ & $+0.000$ & $+0.028$ & 47 \\
\midrule
pair 0--1             & $+0.039$ & $+0.013$ & $+0.217$ & 49 \\
pair 1--2             & $+0.023$ & $+0.007$ & $-0.031$ & 48 \\
pair 2--3             & $+0.018$ & $+0.007$ & $-0.015$ & 48 \\
pair 3--4             & $+0.014$ & $+0.006$ & $-0.015$ & 48 \\
pair 4--5             & $+0.007$ & $+0.003$ & $-0.026$ & 47 \\
pair 5--6             & $+0.005$ & $+0.001$ & $-0.019$ & 47 \\
pair 6--7             & $+0.004$ & $+0.002$ & $-0.022$ & 48 \\
pair 7--8             & $+0.010$ & $+0.004$ & $-0.020$ & 48 \\
pair 8--9             & $+0.015$ & $+0.007$ & $-0.022$ & 46 \\
pair 9--10            & $+0.022$ & $+0.011$ & $-0.019$ & 47 \\
pair 10--11           & $+0.017$ & $+0.008$ & $-0.019$ & 47 \\
\end{tabular}
\end{table}

\subsubsection{Multi-block uniform attention}\label{app:tabicl-uniform-all}
Forcing uniform attention on \{first-6, last-6, all-12\} blocks ($49$
datasets, $10$ seeds, plus sign\_1d):

\begin{table}[H]
\centering
\caption{TabICL accuracy under simultaneous uniform-attention knockout of
contiguous block groups. Mean accuracy across 49 datasets $\times$ 10 seeds,
with the \texttt{sign\_1d} synthetic control reported separately. Neither
half of the stack alone breaks the model; only the all-12 condition collapses
both real-data and synthetic accuracy.}
\label{tab:tabicl-uniform-multi}
\begin{tabular}{lcc}
Condition      & Mean accuracy & \texttt{sign\_1d} \\
\midrule
Baseline       & $0.862$ & $0.993$ \\
First 3 blocks & $0.810$ & ---     \\
Last 3 blocks  & $0.804$ & ---     \\
First 6 blocks & $0.789$ & ---     \\
Last 6 blocks  & $0.737$ & ---     \\
All 12 blocks  & $0.572$ & $0.533$ \\
\end{tabular}
\end{table}

Last-6 is more damaging than first-6; neither half alone breaks the model.
All-12 drops accuracy by $-29$\,pp and collapses sign\_1d from
$0.993$ to $0.533$. Excluding sign\_1d, all-12 has max $0.777$,
median $0.290$ across $48$ datasets, above the majority baseline (mean
$0.498$, \sref{app:E4}). The blocks are collectively necessary and
individually redundant.

\subsubsection{MLP knockouts}\label{app:tabicl-mlp-ko}
Zeroing one ICL MLP (second linear layer) per block ($49$ datasets,
$5$ seeds). Mean accuracy drop ${\leq}2.0$\,pp at every block; the
single largest per-dataset drop anywhere in the stack is
$8.7$\,pp (block 7) and $29.9$\,pp (block 8 on one outlier).
No single MLP is critical. Regression shows a different pattern at
L11 (\sref{app:tabicl-reg-mlp-ko}).

\begin{table}[H]
\centering\small\setlength{\tabcolsep}{4pt}
\caption{Per-block ICL-MLP knockout drops on \TabICL{}-cls
($49$ datasets, baseline $0.864$). Mean and median drops are
sub-percent at every block; the worst single dataset drop never
exceeds $0.30$.}
\label{tab:tabicl-mlp-ko}
\begin{tabular}{lrrrr}
Block & mean drop & median & worst (min) & best (max) \\
\midrule
b0  & $-0.001$ & $-0.001$ & $-0.028$ & $+0.033$ \\
b1  & $-0.004$ & $-0.001$ & $-0.033$ & $+0.022$ \\
b2  & $-0.004$ & $-0.001$ & $-0.059$ & $+0.026$ \\
b3  & $-0.016$ & $-0.006$ & $-0.082$ & $+0.015$ \\
b4  & $-0.007$ & $-0.004$ & $-0.056$ & $+0.015$ \\
b5  & $-0.009$ & $-0.005$ & $-0.062$ & $+0.012$ \\
b6  & $-0.009$ & $-0.002$ & $-0.061$ & $+0.029$ \\
b7  & $-0.004$ & $-0.001$ & $-0.087$ & $+0.020$ \\
b8  & $-0.020$ & $-0.003$ & $-0.299$ & $+0.014$ \\
b9  & $-0.003$ & $-0.002$ & $-0.033$ & $+0.047$ \\
b10 & $-0.002$ & $-0.001$ & $-0.031$ & $+0.027$ \\
b11 & $+0.001$ & $+0.000$ & $-0.009$ & $+0.009$ \\
\end{tabular}
\end{table}

% =============================================================
\subsection{The final-representation readout}
\label{app:tabicl-cls-readout}

\subsubsection{Prototype-classifier readout}\label{app:tabicl-prototype}
Prototype classifiers at each ICL layer ($-1$=ICL input, $0\dots 11$=block
outputs): per-class centroids decoded via
$\mathrm{softmax}(-d(\mathbf{z}, \boldsymbol{\mu}_c))$ ($49$ datasets,
$5$ seeds; baseline $0.864$). $\ell_2$ and cosine prototypes track each
other to $\le 0.002$ at every layer and reach $0.854/0.854$ at L11
vs.\ $0.864$ end-to-end; the learned head adds ${\le}1.0$\,pp.
Calibration analysis (\sref{app:E8}): learned head, L11-prototype, and
L11-$k$NN agree within $\sim 1$\,pp accuracy and $0.92$--$0.93$ pairwise.

\begin{table}[H]
\centering\small\setlength{\tabcolsep}{4pt}
\caption{Per-layer prototype classifier accuracy on \TabICL{}-cls
($49$ datasets, $5$ seeds; baseline $0.864$). Layer $-1$ is the
ICL input (column-embedding output). $\ell_2$/cosine are interchangeable.}
\label{tab:tabicl-prototype}
\resizebox{\linewidth}{!}{%
\begin{tabular}{lrrrrrrrrrrrrr}
Layer & $-1$ & 0 & 1 & 2 & 3 & 4 & 5 & 6 & 7 & 8 & 9 & 10 & 11 \\
\midrule
$\ell_2$ & .735 & .761 & .772 & .790 & .804 & .814 & .823 & .830 & .828 & .834 & .846 & .852 & .854 \\
cosine   & .734 & .760 & .772 & .791 & .802 & .813 & .822 & .830 & .831 & .832 & .845 & .851 & .854 \\
\end{tabular}}
\end{table}

\subsubsection{Cross-version readout checks: TabICL v1 and v1.1}
\label{app:tabicl-vx-readout}
We rerun the readout grid on the TabICL v1 and v1.1 classifiers
over the full $49$-dataset classification suite ($5$ seeds per
dataset) to check whether the prototype/$k$NN picture is specific to
the v2 release. The qualitative ordering survives across all three
releases (Tab.~\ref{tab:tabicl-vx-readout}): L11-prototype and
L11-$k$NN remain close to native accuracy, while the sharpest-block
vote is far worse. Calibration is descriptive only: the v1 native /
prototype / $k$NN / sharp-vote ECE--NLL pairs are $0.047/0.339$,
$0.416/1.022$, $0.135/3.315$, and $0.213/0.893$; the matching v1.1
pairs are $0.050/0.336$, $0.433/1.051$, $0.130/3.186$, and
$0.196/0.821$. The final-representation prototype/$k$NN picture
holds across the released TabICL classifiers and is not an
artefact of the v2 release.

\begin{table}[H]
\centering\small\setlength{\tabcolsep}{5pt}
\caption{Readout-grid comparison across three released TabICL
classification models on the $49$-dataset classification suite
($5$ seeds / dataset). Values are means over per-dataset seed means.}
\label{tab:tabicl-vx-readout}
\begin{tabular}{lcccc}
Release & native & L11-prototype & L11-$k$NN & sharp-vote \\
\midrule
v2   & $0.864$ & $0.854$ & $0.856$ & $0.470$ \\
v1.1 & $0.852$ & $0.830$ & $0.842$ & $0.626$ \\
v1   & $0.852$ & $0.828$ & $0.836$ & $0.590$ \\
\end{tabular}
\end{table}

\subsubsection{Proximity-corrected within-class preference}
\label{app:tabicl-proximity}
At each ICL block: raw within-class attention minus expected value
conditioned on feature proximity (bootstrap resampling). Across $49$
datasets, $0$ show positive corrected preference; $43$ are significantly
negative ($\alpha{=}0.05$). Attention does not seek same-class rows;
geometry is laid down upstream.

\subsubsection{Distance correlations}\label{app:tabicl-distance}
Spearman $\rho$ between attention weights and pairwise distances
(feature-space $\ell_2$, representation-space $\ell_2$, label-embedding
cosine): $<0.08$ at all ICL blocks. Attention does not implement
distance-based retrieval.

\subsubsection{$k$NN agreement with TabICL}\label{app:knn-tabicl}
$k$NN on the L11 representation and on the raw input,
$k{\in}\{1,3,5,10\}$, four metrics, $49$ datasets, $10$ seeds.
\TabICL{} accuracy: $0.864 {\pm} 0.155$. $\ell_2$/cosine on the final
representation match \TabICL{} to within $1$\,pp and agree with it on
$\sim 94\%$ of test queries; Mahalanobis underperforms because of
covariance estimation under small context; raw-input $\ell_2$ is a
non-trivial $\sim 78\%$. Converse pattern of \TabPFN{}
(\sref{app:knn-vs-tabpfn}).

\begin{table}[H]
\centering\small\setlength{\tabcolsep}{4pt}
\caption{$k$NN accuracy / agreement with \TabICL{} predictions, by
metric and $k$ ($49$ datasets, $10$ seeds, baseline $0.864$).}
\label{tab:knn-tabicl}
\begin{tabular}{lcccc}
Metric & $k{=}1$ & $k{=}3$ & $k{=}5$ & $k{=}10$ \\
\midrule
rep $\ell_2$    & .855 / .933 & .856 / .938 & .857 / .938 & .858 / .941 \\
rep cosine      & .855 / .932 & .856 / .936 & .857 / .939 & .858 / .941 \\
rep Maha.       & .652 / .672 & .678 / .702 & .689 / .714 & .700 / .729 \\
input $\ell_2$  & .749 / .788 & .764 / .806 & .772 / .817 & .776 / .828 \\
\end{tabular}
\end{table}

\subsubsection{Per-block vote refutation}\label{app:tabicl-vote-refut}
Each ICL block decoded through final output projection as a per-block vote
($10$ datasets, $5$ seeds):

\begin{table}[H]
\caption{TabICL per-block vote refutation. Each block's hidden state is decoded through the final output projection. Best block is L9 with Pearson $r = 0.462$, far below TabPFN's vote correlation of $0.888$. Mean across 10 datasets $\times$ 5 seeds.}
\label{tab:tabicl-vote-refute}
\centering
\footnotesize
\begin{tabular}{lrrrr}
\textbf{Layer} & \textbf{Pearson $r$} & \textbf{Vote Acc} & \textbf{Agreement} & \textbf{$N$} \\
\midrule
L0  & 0.306 & 0.467 & 0.491 & 10 \\
L1  & 0.376 & 0.525 & 0.552 & 10 \\
L3  & 0.169 & 0.412 & 0.435 & 10 \\
L6  & 0.367 & 0.554 & 0.582 & 10 \\
L9  & 0.462 & 0.584 & 0.612 & 10 \\
L11 & 0.230 & 0.442 & 0.468 & 10 \\
\end{tabular}
\end{table}

Best block: L9, $r(\text{vote}, \text{final}){=}{+}0.462$, agreement $0.612$,
vote-acc $0.584$ vs. true ${\approx}0.864$. Other blocks: $r{=}{+}0.169$
(L3) to $+0.376$ (L1); vote-acc never exceeds $0.584$. Refutes per-block
vote hypothesis. Contrast \TabPFN{}: L9 vote tracks final at $r{=}0.888$
(\sref{app:vote-full}). Regression confirms (\sref{app:K3}).

\section{\Mitra{} audit details}\label{app:mitra}

This appendix collects the per-experiment details for \Mitra{}
referenced from the main text. \Mitra{}'s architecture is described
alongside \TabPFN{} and \TabICL{} in \sref{app:protocol-arch}; the
load-bearing differences for our analysis are the separate per-row
label slot, the shared per-feature embedding, and the absence of
positional encoding on either axis. We audit both released
checkpoints (v1 and v1.1) on the same $49$-dataset classification
and $10$-dataset regression suites used throughout the paper. All
experiments are run offline against local checkpoints; \Mitra{} is
loaded as a single model with ensembling/stacking disabled, so the
AutoGluon defaults that would wrap it in a meta-ensemble are
explicitly turned off and we measure the model in isolation.

\subsection{Permutation invariance (\sref{sec:invariance})}\label{app:mitra-inv}

\Mitra{} is exactly invariant to row \emph{and} column permutations
up to floating-point noise; per-class label invariance fails as
expected because the label slot uses a per-class embedding.

\begin{table}[H]
    \caption{Permutation invariance summary. Quantitative measurements
across $48$ datasets that load under the per-class minimum-sample
filter, $20$ random permutations per dataset, fp32 inference:
mean $|\Delta p|$ is below $0.3\%$ for both row and column on both
\Mitra{} versions; the mean per-dataset \emph{maximum}
$|\Delta p|$ is $1$--$3\%$; predicted-label agreement is
$\geq 99.5\%$. Per-class relabelling produces mean maximum
$|\Delta p|\!\approx\!18$--$20\%$ with label agreement $\approx 0.96$.}
\label{tab:mitra-inv}
\centering\small
\begin{tabular}{lccc}
& Row & Column & Label (per-class) \\
\midrule
\TabPFN{}    & yes (architectural) & no & no \\
\TabICL{}    & yes (architectural) & no (RoPE) & no \\
\Mitra{} v1   & \textbf{yes (architectural)} & \textbf{yes (architectural)} & no \\
\Mitra{} v1.1 & \textbf{yes (architectural)} & \textbf{yes (architectural)} & no \\
\end{tabular}

\end{table}

The exact column invariance follows mechanically from the shared
per-feature embedding plus the absence of any positional encoding:
column attention sees feature tokens as a set, so feature reordering
is exactly a no-op up to numerical reduction order. \TabPFN{} and
\TabICL{} do not have this property because both inject feature
identity (the \TabPFN{} per-feature embedding $W_p$ and \TabICL{}'s
RoPE).

\subsection{Layerwise representation analysis (\sref{sec:repr})}\label{app:mitra-repr}

We attach forward hooks on each \texttt{Layer.forward} of
\Mitra{}'s transformer and capture both the context and query
streams. Mean-pooling along the feature axis (skipping the $y$-slot
at position $0$) gives one vector per row per layer. We then evaluate
four per-layer readouts on the resulting features: a logistic
regression linear probe on standardised features, a class-mean
prototype rule, cosine-distance $5$-NN, and the agreement of the
$5$-NN prediction with \Mitra{}'s native prediction. $45$ of the $49$
datasets load under \Mitra{}'s per-class minimum-sample filter.

\begin{table}[H]
\centering\small
\caption{Layerwise probe summary on the $45$-dataset subset. Both
\Mitra{} versions peak at $\mathrm{L}9$, mirroring the \TabPFN{}
late-jump pattern rather than \TabICL{}'s early convergence. $k$NN
agreement at the peak is $0.82$--$0.83$ and falls to $0.71$--$0.78$
by $\mathrm{L}12$ (Table~\ref{tab:mitra-repr-perlayer}).}
\label{tab:mitra-repr}
\begin{tabular}{lcccc}
version & $n_\mathrm{datasets}$ & e2e mean acc &
peak-probe layer & $k$NN-agreement @\,$\mathrm{L}9$ \\
\midrule
v1   & 45 & 0.827 & $\mathrm{L}9$ ($0.777$) & 0.831 \\
v1.1 & 45 & 0.825 & $\mathrm{L}9$ ($0.784$) & 0.821 \\
\end{tabular}

\end{table}

\begin{table}[H]
\centering\small
\caption{Per-layer probe / prototype / cosine $5$-NN /
$5$-NN-agreement on \Mitra{}, mean over the $45$-dataset subset.
Probe peaks at $\mathrm{L}9$ in both versions, with the
characteristic $\mathrm{L}10$ refinement-phase dip and partial
recovery at $\mathrm{L}11$--$\mathrm{L}12$. $k$NN tracks the probe
within $\sim 2$\,pp at $\mathrm{L}9$; the prototype rule is weaker,
especially for v1.1 (gap $\sim 6$\,pp).}
\label{tab:mitra-repr-perlayer}
\begin{tabular}{ccccc|cccc}
& \multicolumn{4}{c|}{v1 (45 datasets)} & \multicolumn{4}{c}{v1.1 (45 datasets)} \\
layer & probe & proto & $k$NN5 & $k$NN-agree & probe & proto & $k$NN5 & $k$NN-agree \\
\midrule
$\mathrm{L}1$  & 0.576 & 0.533 & 0.592 & 0.640 & 0.430 & 0.462 & 0.500 & 0.525 \\
$\mathrm{L}2$  & 0.681 & 0.652 & 0.690 & 0.746 & 0.519 & 0.522 & 0.579 & 0.611 \\
$\mathrm{L}3$  & 0.698 & 0.677 & 0.695 & 0.763 & 0.667 & 0.673 & 0.697 & 0.758 \\
$\mathrm{L}4$  & 0.713 & 0.693 & 0.716 & 0.791 & 0.672 & 0.673 & 0.698 & 0.762 \\
$\mathrm{L}5$  & 0.731 & 0.712 & 0.731 & 0.808 & 0.746 & 0.722 & 0.756 & 0.833 \\
$\mathrm{L}6$  & 0.752 & 0.734 & 0.753 & 0.837 & 0.761 & 0.736 & 0.756 & 0.835 \\
$\mathrm{L}7$  & 0.759 & 0.751 & 0.763 & 0.845 & 0.771 & 0.737 & 0.764 & 0.847 \\
$\mathrm{L}8$  & 0.771 & 0.745 & 0.759 & 0.836 & 0.774 & 0.723 & 0.754 & 0.831 \\
$\mathrm{L}9$  & \textbf{0.777} & 0.742 & 0.759 & 0.831 & \textbf{0.784} & 0.721 & 0.755 & 0.821 \\
$\mathrm{L}10$ & 0.718 & 0.684 & 0.704 & 0.750 & 0.702 & 0.656 & 0.664 & 0.700 \\
$\mathrm{L}11$ & 0.753 & 0.739 & 0.769 & 0.830 & 0.685 & 0.655 & 0.657 & 0.671 \\
$\mathrm{L}12$ & 0.676 & 0.683 & 0.733 & 0.777 & 0.673 & 0.646 & 0.688 & 0.706 \\
\end{tabular}

\end{table}

\subsection{Per-block knockouts (\sref{sec:critical-block})}\label{app:mitra-block-ko}

We replicate the per-block knockout protocol of \sref{app:block-ko}
on \Mitra{}. Each interleaved row/feature \texttt{Layer} is
replaced one at a time with the identity (a forward pre-hook
returning its inputs unchanged); the rest of the network runs
normally, and we report the resulting accuracy on the same
$49$ datasets as \sref{app:mitra-inv}, with $300$ context and $200$
query rows per dataset.

Layer~$0$ is the most damaging single intervention on both
checkpoints, by a margin: it costs a mean of $15.8$\,pp on v1 and
$41.1$\,pp on v1.1, while the runner-up block costs $\le 2$\,pp on
v1 and $19.6$\,pp on v1.1 (Layer~$1$). Layer~$0$ is the worst
single block on $31/49$ datasets for v1 and $36/49$ for v1.1; no
other block is the worst on more than three datasets in v1 or one
in v1.1. The remaining eleven blocks each cost $1$--$3$\,pp in mean
drop, with no late-layer organizing spike comparable to \TabPFN{}'s
block~$9$ (\sref{app:frozen-probe}); consistent with the geometric,
late-jumping readout reported in \sref{sec:mitra-readout}, where a
single late layer ($\mathrm{L}9$) suddenly aligns with the
attention-weighted vote. The asymmetry between v1 and v1.1 mirrors the per-layer
probe pattern of \sref{tab:mitra-repr-perlayer}: v1.1 compresses
class structure into the early frame more aggressively than v1,
and pays for it more under early-block ablation.

\subsection{Readout fidelity (\sref{sec:readout})}\label{app:mitra-readout}

On the full $49$-dataset classification grid
($1$ seed/dataset, v1), \Mitra{}'s native head reaches mean accuracy
$0.860$. At $\mathrm{L}9$, the attention-weighted vote (defined
below) reaches $0.815$ ($-4.5$\,pp, mean Pearson $r{=}0.89$ vs.\
native), the cosine $5$-NN readout reaches $0.789$ (within $\sim
1.5$\,pp of the L9 linear probe at $0.776$), and the class-mean
prototype rule reaches $0.769$. Vote leads $k$NN by $2.6$\,pp and
prototype by $4.6$\,pp on accuracy; argmax-agreement of $k$NN with
\Mitra{}'s native predictions at $\mathrm{L}9$ is $0.82$--$0.83$,
and $k$NN tracks the linear probe within $\sim 2$\,pp through the
entire $\mathrm{L}5$--$\mathrm{L}9$ window where both rise
monotonically (Table~\ref{tab:mitra-repr-perlayer}). All three rules
are retrieval-over-context-rows surrogates and are consistent with
the row-attention ablation (\sref{app:mitra-collapse}); among them
the L9 vote is the closest match to the native head and is reported
as \Mitra{}'s readout in \sref{sec:readout}. Cosine $k$NN is
reported as a parameter-free diagnostic.

\paragraph{Layer choice (v1 vs.\ v1.1).}
On v1 weights, the vote-faithful and probe-faithful layers coincide
at $\mathrm{L}9$: a layerwise sweep of vote accuracy on the
$45$-dataset subset of Table~\ref{tab:mitra-repr-perlayer} shows
$\mathrm{L}9$ at the joint optimum, with $\mathrm{L}8$ and
$\mathrm{L}10$ within $0.2$\,pp and $\mathrm{L}12$ within
$0.2$\,pp ($0.779$, $0.778$, $0.777$, $0.778$ respectively;
$\mathrm{L}11$ underperforms by $\sim 7$\,pp). On v1.1, the
vote-faithful layer shifts to the final attention block:
$\mathrm{L}12$ vote reaches $0.810$ (gap $-3.4$\,pp, agree
$0.89$, $r{=}0.89$) on the $42$-dataset OpenML grid, while
$\mathrm{L}9$ underperforms $\mathrm{L}12$ by $\sim 7.7$\,pp.
We use v1 throughout the main text so that the same layer
$\mathrm{L}9$ surrogates both attention vote and the linear probe.

\paragraph{Attention-weighted vote at $\mathrm{L}9$.}
\Mitra{}'s \texttt{MultiheadAttention} dispatches to
\texttt{F.scaled\_dot\_product\_attention}, which does not expose the
attention weights, so we wrap
\texttt{layer.attention1.forward} at $\mathrm{L}9$ to additionally
compute $\mathrm{softmax}(QK^{\top}/\sqrt{d_h})$ on the
cross-attention call (query rows attending over context rows)
without altering the returned tensor or \Mitra{}'s predictions.
Averaging the resulting weights over heads and over the per-feature
row-attention slices and forming the attention-weighted vote
$\mathrm{vote}[t,c]\propto\sum_{i:y^{\mathrm{ctx}}_i=c}\bar{a}_{t,i}$,
the $\mathrm{L}9$ vote rule reaches mean accuracy $0.815$ on the
full $49$-dataset grid (vs.\ native $0.860$, gap $4.5$\,pp; mean
Pearson $r$ vs.\ native probabilities $0.89$). The same construction
restricted to the $y$-slot of the row-attention reshape is
\emph{worse}, not better, at every layer (best $\sim 0.74$ at
$\mathrm{L}12$ on v1.1), because column attention mixes $y$-slot
and feature-slot information within each observation, leaving the
per-feature average more stable than any single slot.

\subsection{Adversarial probes (\sref{sec:attacks})}\label{app:mitra-attacks}

We port the full attack battery used for \TabPFN{} and \TabICL{}
(\sref{app:attacks}) to \Mitra{}: random-label padding ($2\times$ and
$4\times$), boundary poisoning, hub poisoning, class-centroid
injection, monotone distortions (cube / softexp / rank), and the
adversarial low-variance rotation. The same $49$-dataset suite,
splits, and Ridge baseline are used.

\begin{table}[H]
\centering\small
\caption{Mean accuracy drop relative to the unattacked baseline,
$49$ datasets. ``Worse than R.'' counts datasets on which \Mitra{}
ends below Ridge after the attack.}
\label{tab:mitra-attacks}
\begin{tabular}{lrrrrr}
attack & Ridge $\Delta$ &
\Mitra{} v1 $\Delta$ & v1 worse than R. &
\Mitra{} v1.1 $\Delta$ & v1.1 worse than R. \\
\midrule
pad2x         & $-0.047$ & $-0.049$ & 7/49  & $-0.032$ & 6/49  \\
pad4x         & $-0.081$ & $-0.075$ & 11/49 & $-0.054$ & 8/49  \\
boundary      & $-0.021$ & $-0.028$ & 11/49 & $-0.016$ & 9/49  \\
hub poisoning & $-0.012$ & $-0.021$ & \textbf{14/49} & $-0.022$ & \textbf{11/49} \\
centroid inj. & $-0.002$ & $+0.001$ & 6/49  & $+0.000$ & 7/49  \\
mono cube     & $-0.022$ & $-0.000$ & 5/49  & $-0.001$ & 3/49  \\
mono softexp  & $-0.043$ & $-0.044$ & 7/49  & $-0.045$ & 7/49  \\
mono rank     & $-0.042$ & $-0.000$ & 3/49  & $-0.003$ & 3/49  \\
rotate (low-$\sigma$) & $-0.001$ & $-0.006$ & 10/49 & $-0.001$ & 7/49 \\
\end{tabular}

\end{table}

\Mitra{} is uniformly more robust than Ridge under monotone
distortions (cube, rank): essentially zero drop where Ridge gives up
$2$--$4$\,pp. Pad-style and rotation attacks degrade \Mitra{}
comparably to Ridge. Hub poisoning is the clearest case where
\Mitra{} ends below Ridge on a substantial fraction of datasets
($14/49$ on v1, $11/49$ on v1.1); boundary poisoning and rotation
also produce non-trivial worse-than-Ridge counts but with smaller
mean drops. The hub-poisoning pattern is consistent with the
vote-faithful readout established in \sref{app:mitra-readout}:
inserting label-anchored hubs in the context directly perturbs the
context-similarity signal that the readout relies on.

\subsection{Collapse mechanism (\sref{sec:repr-collapse})}\label{app:mitra-collapse}

\Mitra{} does not exhibit representation collapse on balance-scale
or on a $d{=}12$ \emph{balance-like} synthetic dataset
(Table~\ref{tab:mitra-collapse-base}). We further perform the
analogue of the \TabPFN{}/\TabICL{} double ablation by zeroing each
of \Mitra{}'s two attention paths (\texttt{Layer.attention1}: row
attention; \texttt{Layer.attention2}: column attention) via verified
forward hooks --- a per-call counter confirms the hook fires --- and
measuring the resulting accuracy.

\begin{table}[H]
\centering\small
\caption{\Mitra{} accuracy on the canonical balance-scale stress
test and a $d{=}12$ balance-like variant. No collapse is observed
in either version.}
\label{tab:mitra-collapse-base}
\begin{tabular}{lcc}
version & balance-scale (3 seeds) & balance-like, $d{=}12$ (3 seeds) \\
\midrule
v1   & $0.961 \pm 0.007$ & $0.893 \pm 0.025$ \\
v1.1 & $0.959 \pm 0.014$ & $0.969 \pm 0.003$ \\
majority baseline & $0.436$ & $\sim 0.47$ \\
\end{tabular}

\end{table}

\begin{table}[H]
\centering\small
\caption{Double ablation of \Mitra{}'s attention paths.
``row\_off'' zeroes \texttt{Layer.attention1} on every layer;
``col\_off'' zeroes \texttt{Layer.attention2}. Removing row
attention drops accuracy from $\sim 0.95$ to $0.08$, well below the
majority baseline, indicating a systematic misclassification regime
rather than collapse to majority. Removing column attention
collapses to near-majority. Removing both lands at the
``row\_off'' level.}
\label{tab:mitra-collapse-ablation}
\begin{tabular}{llcccc}
version & dataset & full & row\_off & col\_off & both\_off \\
\midrule
v1   & balance-scale     & \textbf{0.954} & 0.080 & 0.459 & 0.080 \\
v1   & balance-like-d12  & \textbf{0.893} & 0.074 & 0.459 & 0.074 \\
v1.1 & balance-scale     & \textbf{0.963} & 0.080 & 0.461 & 0.080 \\
v1.1 & balance-like-d12  & \textbf{0.969} & 0.074 & 0.467 & 0.074 \\
\midrule
\multicolumn{2}{l}{majority baseline} & \multicolumn{4}{l}{$\sim 0.46$ (balance-scale), $\sim 0.47$ (balance-like)} \\
\end{tabular}

\end{table}

The collapse-critical mechanism for \Mitra{} is therefore
\emph{row attention}, the only one of the three families for which
the critical mechanism is itself permutation-symmetric:
\TabPFN{}'s is the feature-positional pathway $W_p$, \TabICL{}'s is
RoPE on the row axis, and \Mitra{}'s is the cross-row aggregator.
Combined with the vote-faithful readout
(\sref{app:mitra-readout}), this picture is consistent with the
differential vulnerability to hub poisoning observed in
\sref{app:mitra-attacks}: corrupting label-anchored hubs perturbs
the context-similarity signal that flows through exactly the
row-attention path the ablation here identifies as load-bearing.
We do not claim direct causal attribution from the balance-scale
ablation to the hub-poisoning result --- the two experiments use
different datasets and stress different operating regimes --- only
that both point at row attention as the natural locus of failure.

\subsection{Three-way contrast}\label{app:mitra-contrast}

\begin{table}[H]
\centering\footnotesize
\caption{Three-way comparison across the dimensions audited in this
paper. \Mitra{}'s \emph{compute schedule} is \TabPFN{}-like (late
jump in depth), its \emph{readout family} is also \TabPFN{}-like
(attention-weighted vote at $\mathrm{L}9$), and its \emph{symmetry
profile} is strictly stronger than both --- column invariance is
exact by construction.}
\label{tab:mitra-contrast}
\begin{tabular}{lccc}
dimension & \TabPFN{} & \TabICL{} & \Mitra{} \\
\midrule
Token factorization & feat$\oplus$label per token & feat$\oplus$label per token & separate $y$-slot per row \\
Positional structure & per-feature $W_p$ & RoPE on feature axis & none \\
Row invariance & yes & yes & yes \\
Column invariance & no & no (RoPE) & \textbf{yes (exact, no patch)} \\
Per-class label invariance & no & no & no \\
Probe peak layer & late ($\mathrm{L}9$) & early ($\mathrm{L}2$--$\mathrm{L}4$) & late ($\mathrm{L}9$) \\
Faithful surrogate (cls) & attention vote @\,L9 & class-mean prototype @\,L11 & attention vote @\,L9 \\
Hub-poisoning vulnerability & yes & yes & yes \\
Monotone-distortion (rank/cube) & low & low & lowest \\
Balance-scale collapse & no & sometimes (v1) & no \\
Critical mechanism (ablation) & feat-pos $W_p$ & RoPE & \textbf{row attention} \\
Critical mech.\ permutation-symm.\ & no & no & \textbf{yes} \\
\end{tabular}

\end{table}

\paragraph{Summary.}
\Mitra{} is its own combination of design choices. Token-level
factorization of features and labels, a shared per-feature
embedding, and the absence of positional encoding together make it
the only one of the three families for which column invariance is
exact, the cosine-$k$NN readout is the closest similarity-based
surrogate to native predictions of the three families
(\sref{sec:readout}), and the collapse-critical mechanism uncovered by
ablation is itself permutation-symmetric. Empirically, this
combination delivers full robustness to monotone feature
distortions (cube, rank) and no balance-scale collapse, while
preserving the same hub-poisoning weakness as \TabPFN{}/\TabICL{}.
Two ablations are consistent with the mechanistic narrative for
\Mitra{}: the row-attention zero-out
(Table~\ref{tab:mitra-collapse-ablation}) and the hub-poisoning
attack (Table~\ref{tab:mitra-attacks}); the connection between them
remains an inference rather than a single end-to-end intervention.

\section{TabPFN regression: full results}
\label{app:tabpfn-reg}

Regression measurements cover $10$ datasets with $5$ seeds each unless
noted. The subsections that follow trace the late-readout circuit from
classification through regression: per-block representation geometry,
row-attention sharpness, activation patching, the L9 attention-vote
test, and the MLP-knockout and uniform-attention falsification
controls. Two systematic shifts emerge against the classification
baseline --- the probe peak moves one layer earlier and row-attention
is substantially softer --- both consistent with the bias--variance
trade-off of continuous targets.

\subsection{Per-block representation geometry}
\label{app:tabpfn-reg-geom}
Per-block geometry on regression matches the classification profile.
Table~\ref{tab:tabpfn-reg-geom} reports per-block effective rank (ER),
participation ratio (PR) and 2-NN intrinsic dimension (ID) averaged
across the $10$ regression datasets. The L2 block triggers a sharp ER
collapse ($\mathrm{ER}\!\approx\!9.5$, $\mathrm{PR}\!\approx\!1.5$),
L3--L8 rebuild the representation gradually, and L9--L11 reach maximum
ER and PR where the readout forms. Intrinsic dimension contracts from
$\mathrm{ID}\!\approx\!8.0$ at L1 to a $3.4$--$3.7$ plateau from L6
onward, exposing a low-dimensional predictive manifold.

\begin{table}[H]
\centering
\caption{Per-block geometry on TabPFN-reg, mean$\pm$std across $10$
  regression datasets ($5$ seeds each).}
\label{tab:tabpfn-reg-geom}
\small
\setlength{\tabcolsep}{4pt}
\begin{tabular}{lccc}
\textbf{Block} & \textbf{Effective rank} & \textbf{Participation ratio} & \textbf{2-NN ID} \\
\midrule
L0  & $78.7 \pm 15.0$ & $3.60 \pm 0.81$ & $5.54 \pm 1.16$ \\
L1  & $86.9 \pm 15.9$ & $4.68 \pm 1.14$ & $7.99 \pm 2.45$ \\
L2  & $9.5 \pm 0.6$   & $1.47 \pm 0.18$ & $3.49 \pm 0.75$ \\
L3  & $22.6 \pm 2.0$  & $3.06 \pm 0.29$ & $3.76 \pm 1.29$ \\
L4  & $30.1 \pm 4.0$  & $2.76 \pm 0.40$ & $4.22 \pm 1.63$ \\
L5  & $26.6 \pm 3.9$  & $2.91 \pm 0.61$ & $3.72 \pm 1.38$ \\
L6  & $32.8 \pm 4.4$  & $3.71 \pm 0.78$ & $3.42 \pm 1.34$ \\
L7  & $34.4 \pm 5.2$  & $3.73 \pm 0.76$ & $3.38 \pm 1.28$ \\
L8  & $30.0 \pm 4.9$  & $3.29 \pm 0.64$ & $3.46 \pm 1.33$ \\
L9  & $36.5 \pm 6.9$  & $3.96 \pm 0.83$ & $3.49 \pm 1.40$ \\
L10 & $34.3 \pm 7.5$  & $3.73 \pm 1.03$ & $3.69 \pm 1.48$ \\
L11 & $38.3 \pm 15.7$ & $5.25 \pm 2.45$ & $3.72 \pm 1.44$ \\
\end{tabular}
\end{table}

\subsection{Regression row-attention}
\label{app:tabpfn-reg-rowattn}
Regression row-attention is markedly softer and peaks earlier than in
classification. Classification L9 max weight averages $0.925$, while
regression L9 max weight peaks at $\approx 0.26$ and is sharpest at
L6--L7 on most datasets; L9 entropy reaches $\approx 3.8$ versus
$1.756$ for classification. The one-to-two-layer earlier peak tracks
the probe peak shift: regression reaches usable representations earlier
and pools them across more context rows.

\subsection{Regression activation patching}
\label{app:tabpfn-reg-actpatch}
Activation patching restores the baseline on every regression dataset.
Replacing the L9 state of blocks $\{0, 5, 9\}$ with the cached clean
activations yields recovery $= 1.000 \pm 0.000$, matching the
classification result in \sref{app:activation-patching}.
Table~\ref{tab:tabpfn-reg-actpatch} gives the per-dataset baseline
$R^2$, the two corruptions (label-shuffle, block-$0$ knockout) and the
patched-L9 score; once the corrupted state is overwritten with the
cached clean state the test-row representations become identical to
baseline and the score follows.

\begin{table}[H]
\centering
\caption{TabPFN-reg activation patching, per dataset (mean of $5$ seeds).
  ``LS'' = label-shuffle corruption; ``B0-KO'' = block-$0$ knockout
  corruption. ``Patched'' = restored after L9 activation patching.}
\label{tab:tabpfn-reg-actpatch}
\small
\setlength{\tabcolsep}{4pt}
\begin{tabular}{lcccc}
\textbf{Dataset} & \textbf{Baseline $R^2$} & \textbf{LS corrupt} & \textbf{B0-KO corrupt} & \textbf{Patched $R^2$} \\
\midrule
Fiat-500     & $0.867$  & $-0.048$ & $0.006$  & $0.867$ \\
Moneyball    & $0.947$  & $-0.002$ & $0.021$  & $0.947$ \\
QSAR-fish    & $0.648$  & $-0.008$ & $-0.212$ & $0.648$ \\
airfoil      & $0.964$  & $-0.009$ & $-0.082$ & $0.964$ \\
cars         & $0.952$  & $0.005$  & $-0.037$ & $0.952$ \\
concrete     & $0.936$  & $-0.008$ & $-0.200$ & $0.936$ \\
energy       & $0.998$  & $0.004$  & $0.095$  & $0.998$ \\
forest-fires & $-0.028$ & $-0.030$ & $-0.044$ & $-0.028$ \\
healthcare   & $0.805$  & $-0.012$ & $-0.116$ & $0.805$ \\
socmob       & $0.796$  & $-0.018$ & $-0.065$ & $0.796$ \\
\midrule
mean         & $0.789$  & $-0.013$ & $-0.063$ & $0.789$ \\
\end{tabular}
\end{table}

\subsection{Per-block attention vote vs.\ permutation null}
\label{app:tabpfn-reg-vote}
On every regression dataset the model learns, the L9 attention vote
drives the prediction. Pearson and Spearman correlations between the
model output and the attention-weighted mean of context targets are
compared against a $1000$-permutation null
(Table~\ref{tab:tabpfn-reg-vote}). The vote is significant on $9/10$
datasets ($p\!<\!10^{-3}$, mean Pearson $0.745$). The single failure,
\texttt{forest-fires}, is also the dataset on which the model itself
fails to learn ($R^2\!=\!-0.028$ in
Table~\ref{tab:tabpfn-reg-actpatch}), so the vote correctly registers
as uninformative.

\begin{table}[H]
\centering
\caption{TabPFN-reg L9 attention-vote correlations and $1000$-permutation
  null, per dataset (mean of $10$ seeds).}
\label{tab:tabpfn-reg-vote}
\small
\setlength{\tabcolsep}{4pt}
\begin{tabular}{lcccc}
\textbf{Dataset} & \textbf{Pearson} & \textbf{Spearman} & \textbf{$z$ vs.\ null} & \textbf{$p$} \\
\midrule
Fiat-500     & $0.965$  & $0.908$  & $3.07$  & $<10^{-4}$ \\
Moneyball    & $0.945$  & $0.981$  & $2.00$  & $0.0006$ \\
QSAR-fish    & $0.908$  & $0.918$  & $4.43$  & $<10^{-4}$ \\
airfoil      & $0.917$  & $0.921$  & $5.31$  & $<10^{-4}$ \\
cars         & $0.941$  & $0.880$  & $4.14$  & $<10^{-4}$ \\
concrete     & $0.892$  & $0.926$  & $5.36$  & $<10^{-4}$ \\
energy       & $0.796$  & $0.674$  & $3.47$  & $<10^{-4}$ \\
forest-fires & $-0.577$ & $-0.573$ & $-1.20$ & $0.297$ \\
healthcare   & $0.822$  & $0.683$  & $3.83$  & $<10^{-4}$ \\
socmob       & $0.840$  & $0.548$  & $2.89$  & $0.0003$ \\
\end{tabular}
\end{table}

\subsection{TabPFN falsification controls (MLP-KO and L9 uniform-attn)}
\label{app:tabpfn-falsif}
Two counterfactual interventions falsify the attention-vote rule as a
sole mechanism. On a mixed cls/reg panel ($10 + 10$ datasets, $5$
seeds), Table~\ref{tab:tabpfn-falsif} shows that both interventions
degrade performance severely: MLP computation at every layer is
load-bearing, and L9 attention is necessary but not sufficient.

\begin{table}[H]
\centering
\caption{TabPFN falsification controls. Per-block MLP knockout zeros the
  MLP's second linear at each block without retraining. L9 uniform-attn
  replaces the L9 softmax with $1/N_{\text{train}}$. Mean over $10{+}10$
  cls/reg datasets, $5$ seeds each.}
\label{tab:tabpfn-falsif}
\small
\setlength{\tabcolsep}{4pt}
\resizebox{\linewidth}{!}{%
\begin{tabular}{lcccc}
\textbf{Intervention} & \textbf{Cls baseline} & \textbf{Cls result} & \textbf{Reg baseline} & \textbf{Reg result} \\
\midrule
MLP-KO (any block L0--L11) & $0.874$ & $0.415$ ($\Delta{=}{-}0.460$) & $0.789$ & ${-}0.010$ ($\Delta{=}{-}0.798$) \\
L9 uniform-attn & $0.874$ & $0.488$ ($\Delta{=}{-}0.386$) & $0.789$ & $0.585$ ($\Delta{=}{-}0.203$) \\
\end{tabular}}
\end{table}

The smaller regression drop under uniform attention follows directly
from the softer attention profile (\sref{app:tabpfn-reg-rowattn}): the
L9 distribution is already close to uniform, so flattening it removes
less signal.

\section{TabICL regression: full results}\label{app:tabicl-reg}

Regression results for \TabICL{} parallel the classification analysis
in \sref{app:tabicl-cls} across $10$ datasets and $5$ seeds, with
baseline $R^2 = 0.805$. The structural difference from classification
is the non-linear regression head at L11, which shapes both the
linear-probe profile and the L11 knockout asymmetries reported below.
The $\ell_2$-$k$NN tracking on the final representation
(\sref{app:K2}) and the per-block vote refutation (\sref{app:K3})
both replicate in this setting.

\subsection{Layer-by-layer summary}\label{app:tabicl-reg-overview}

Per-layer linear-probe $R^2$ (ridge, fit on the cell representation
of the query rows, evaluated against the regression target) is
$\le 0.07$ at every layer and \emph{negative} at L9--L11, confirming
the readout is non-linear at the top of the stack
(Table~\ref{tab:tabicl-reg-probe}). The geometric trace of the
representation (Table~\ref{tab:tabicl-reg-geom}) shows monotone
expansion of effective rank from $19.3$ at the column-embedding output
to $100.0$ at L11; participation ratio peaks mid-stack and 2NN-ID grows
from $3.4$ to $6.0$.

\begin{table}[H]
\centering\small\setlength{\tabcolsep}{4pt}
\caption{Per-layer linear probe $R^2$ for \TabICL{}-reg
(mean$\pm$std across $10$ datasets). Layer $-1$ is the
column-embedding output; L0--L11 are the ICL blocks.}
\label{tab:tabicl-reg-probe}
\resizebox{\linewidth}{!}{%
\begin{tabular}{lcccccccccccccc}
Layer & $-1$ & 0 & 1 & 2 & 3 & 4 & 5 & 6 & 7 & 8 & 9 & 10 & 11 \\
\midrule
$R^2$ & $+0.004$ & $+0.005$ & $+0.007$ & $+0.004$ & $-0.014$ & $+0.061$ & $+0.027$ & $+0.046$ & $+0.024$ & $+0.041$ & $-0.157$ & $-0.463$ & $-0.469$ \\
std   & $0.013$  & $0.013$  & $0.013$  & $0.014$  & $0.015$  & $0.020$  & $0.024$  & $0.025$  & $0.028$  & $0.025$  & $0.043$  & $0.074$  & $0.111$  \\
\end{tabular}}
\end{table}

\begin{table}[H]
\centering\small\setlength{\tabcolsep}{4pt}
\caption{Representation geometry per block for \TabICL{}-reg
(mean across $10$ datasets, $5$ seeds). ER $=$ effective rank,
PR $=$ participation ratio, ID $=$ TwoNN intrinsic dimension.}
\label{tab:tabicl-reg-geom}
\begin{tabular}{lccccccccccccc}
Layer & $-1$ & 0 & 1 & 2 & 3 & 4 & 5 & 6 & 7 & 8 & 9 & 10 & 11 \\
\midrule
ER & 19.3 & 26.1 & 33.7 & 40.1 & 45.8 & 47.4 & 64.6 & 74.1 & 79.2 & 68.1 & 88.8 & 87.1 & 100.0 \\
PR &  4.4 &  5.7 &  6.9 &  7.0 &  6.8 &  7.9 &  9.7 & 10.0 &  7.4 &  6.8 &  8.0 &  7.1 &   5.6 \\
ID &  3.4 &  3.6 &  3.8 &  4.1 &  4.2 &  4.4 &  4.8 &  5.2 &  5.0 &  4.7 &  5.3 &  5.2 &   6.0 \\
\end{tabular}
\end{table}

\subsection{Per-block knockout}\label{app:tabicl-reg-blockko}

Zeroing one ICL block at a time and re-running the regression head
produces no catastrophic loss except at L11
(Table~\ref{tab:tabicl-reg-blockko}). Column- and row-embedding
knockouts are equally tolerated (col-L0/L1/L2 $R^2$ in $[0.71, 0.78]$;
row-L0/L1/L2 $R^2$ in $[0.77, 0.81]$). The contrast with L11
($R^2 = -1.16$) is the single distributed-vs-localized signal in this
panel and matches the linear-probe drop at the top of
Table~\ref{tab:tabicl-reg-probe}.

\begin{table}[H]
\centering\small\setlength{\tabcolsep}{4pt}
\caption{Per-block ICL knockout $R^2$ for \TabICL{}-reg
(mean$\pm$std across datasets; baseline $0.805$).}
\label{tab:tabicl-reg-blockko}
\resizebox{\linewidth}{!}{%
\begin{tabular}{lcccccccccccc}
ICL block & 0 & 1 & 2 & 3 & 4 & 5 & 6 & 7 & 8 & 9 & 10 & 11 \\
\midrule
$R^2$ & $+0.731$ & $+0.807$ & $+0.808$ & $+0.805$ & $+0.793$ & $+0.801$ & $+0.781$ & $+0.711$ & $+0.793$ & $+0.740$ & $+0.772$ & $-1.156$ \\
std   & $0.029$  & $0.024$  & $0.023$  & $0.024$  & $0.019$  & $0.025$  & $0.041$  & $0.085$  & $0.027$  & $0.062$  & $0.043$  & $0.261$  \\
\end{tabular}}
\end{table}

\subsection{Frozen probe post-KO}\label{app:tabicl-reg-frozen-postko}

Ridge probe at L11: baseline frozen $R^2 = -0.469$ (the head is
non-linear, so the linear probe is already negative at baseline).
Table~\ref{tab:tabicl-reg-frozen-postko} compares the frozen probe
(fit on clean activations, evaluated on KO activations) against a
probe re-trained on KO activations. Large frozen-to-retrained gap at
ICL-b0 and ICL-b6 indicates that the knockout disrupts the geometry
the clean probe relies on, but a fresh probe can still recover most
of the linearly decodable signal. ICL-b11 is a no-op for the
\emph{linear} probe at L11 because zeroing the block makes the
representation pre-block, which is what the probe is fit on. Together
with the $k$NN-vs-prototype divergence in \sref{app:K2} this confirms
distributed construction $+$ non-linear readout.

\begin{table}[H]
\centering\small\setlength{\tabcolsep}{6pt}
\caption{Frozen vs.\ retrained ridge probe at L11 after each
single-block knockout (\TabICL{}-reg, $5$ seeds, $10$ datasets).}
\label{tab:tabicl-reg-frozen-postko}
\begin{tabular}{lcc}
KO site & frozen $R^2$ & retrained $R^2$ \\
\midrule
ICL-b0  & $-0.881$ & $-0.006$ \\
ICL-b6  & $-0.692$ & $-0.273$ \\
ICL-b11 & $-0.469$ & $-0.469$ \\
ColEmb-b2 & $-0.610$ & $-0.275$ \\
\midrule
(no KO) & $-0.469$ & $-0.469$ \\
\end{tabular}
\end{table}

A complementary frozen-probe sweep ($10$ seeds) intervenes at three
earlier sites and reads probes off multiple layers; the frozen-vs-retrained gap is uniformly large
(e.g.\ KO@5 reads $R^2$ frozen $-0.626$ vs.\ retrained $-2.81$ at
L11), matching the picture above.

\subsection{Per-block MLP knockout}\label{app:tabicl-reg-mlp-ko}

Zeroing each ICL block's MLP second linear layer
(Table~\ref{tab:tabicl-reg-mlp-ko}). L0--L7 lose at most $0.04$ in
$R^2$; L8--L10 lose $0.06$--$0.09$; L11 collapses to $-0.63$
($\Delta R^2 = -1.43$). Classification (\sref{app:tabicl-mlp-ko})
showed $\le 2.1$\,pp drop at any block. The asymmetry at L11
reflects that the regression head is a full MLP rather than a
linear projection.

\begin{table}[H]
\centering\small\setlength{\tabcolsep}{4pt}
\caption{Per-ICL-block MLP knockout $R^2$ for \TabICL{}-reg
(mean$\pm$std across datasets; baseline $0.805$).}
\label{tab:tabicl-reg-mlp-ko}
\resizebox{\linewidth}{!}{%
\begin{tabular}{lcccccccccccc}
ICL block & 0 & 1 & 2 & 3 & 4 & 5 & 6 & 7 & 8 & 9 & 10 & 11 \\
\midrule
$R^2$ & $+0.803$ & $+0.808$ & $+0.804$ & $+0.800$ & $+0.786$ & $+0.797$ & $+0.786$ & $+0.767$ & $+0.738$ & $+0.740$ & $+0.720$ & $-0.630$ \\
std   & $0.025$  & $0.025$  & $0.026$  & $0.022$  & $0.020$  & $0.019$  & $0.024$  & $0.030$  & $0.079$  & $0.074$  & $0.092$  & $0.177$  \\
\end{tabular}}
\end{table}

\subsection{Row-attention entropy}\label{app:tabicl-reg-rowattn}

Row-attention entropy (averaged over heads, query rows, and
datasets) stays flat across the ICL stack
(Table~\ref{tab:tabicl-reg-rowattn}; min $6.53$ at L11, max $6.66$
at L2, range $\approx 0.13$ nats out of $\log_e 1024 = 6.93$).
\TabPFN{}-reg, by contrast, reaches max-row-weight $\approx 0.26$
at L9 (\sref{app:tabpfn-reg}) and approaches a near-prototype
readout; \TabICL{}-reg never localizes onto a small context set.

\begin{table}[H]
\centering\small\setlength{\tabcolsep}{4pt}
\caption{Per-layer row-attention entropy (nats) for \TabICL{}-reg.}
\label{tab:tabicl-reg-rowattn}
\begin{tabular}{lcccccccccccc}
Layer & 0 & 1 & 2 & 3 & 4 & 5 & 6 & 7 & 8 & 9 & 10 & 11 \\
\midrule
$H$ & $6.61$ & $6.65$ & $6.66$ & $6.66$ & $6.66$ & $6.64$ & $6.66$ & $6.62$ & $6.62$ & $6.59$ & $6.59$ & $6.53$ \\
\end{tabular}
\end{table}

\subsection{Label perturbations}\label{app:tabicl-reg-labels}

Two label-corruption interventions
(Table~\ref{tab:tabicl-reg-labels}), random-shuffle and
mean-replace, both drive $R^2$ to near zero, confirming the
predictions are driven by the in-context labels rather than by
covariate-only structure.

\begin{table}[H]
\centering\small\setlength{\tabcolsep}{6pt}
\caption{Label-perturbation $R^2$ for \TabICL{}-reg
(mean$\pm$std, $10$ datasets, baseline $0.805$).}
\label{tab:tabicl-reg-labels}
\begin{tabular}{lc}
Intervention & $R^2$ \\
\midrule
shuffle context labels & $-0.016 \pm 0.019$ \\
replace with target mean & $-0.010 \pm 0.003$ \\
\end{tabular}
\end{table}

\subsection{Uniform-attention knockout}\label{app:tabicl-reg-uniform}

Replacing the row-attention map at one ICL block by a uniform map
(Table~\ref{tab:tabicl-reg-uniform}) is well tolerated for L3+
($R^2 \ge 0.64$) and breaks the model only at L0--L1
($R^2 = 0.03, 0.19$). Combined with the near-uniform entropy
above, row-attention does little selective work after the first two
ICL blocks.

\begin{table}[H]
\centering\small\setlength{\tabcolsep}{4pt}
\caption{Per-layer uniform-attention $R^2$ for \TabICL{}-reg
(mean$\pm$std across datasets; baseline $0.805$).}
\label{tab:tabicl-reg-uniform}
\resizebox{\linewidth}{!}{%
\begin{tabular}{lcccccccccccc}
Layer & 0 & 1 & 2 & 3 & 4 & 5 & 6 & 7 & 8 & 9 & 10 & 11 \\
\midrule
$R^2$ & $+0.029$ & $+0.187$ & $+0.554$ & $+0.713$ & $+0.765$ & $+0.733$ & $+0.776$ & $+0.641$ & $+0.797$ & $+0.686$ & $+0.795$ & $+0.801$ \\
std   & $0.664$  & $0.577$  & $0.238$  & $0.113$  & $0.059$  & $0.076$  & $0.056$  & $0.172$  & $0.025$  & $0.040$  & $0.034$  & $0.028$  \\
\end{tabular}}
\end{table}

\subsection{Per-block vote refutation}\label{app:tabicl-reg-vote}

Following the classification protocol (\sref{app:K3}) we read off a
prototype-style prediction from each ICL block and correlate it with
\TabICL{}'s own prediction and with the ground-truth target; see
Table~\ref{tab:tabicl-reg-vote}. No early or middle block matches
the model's predictions; the highest agreement is at L9
(Pearson $0.73$ with the model, $0.65$ with truth) and L11
($0.67 / 0.64$). The pattern --- correlations only emerging at the
last few blocks --- is the regression analogue of the classification
vote refutation.

\begin{table}[H]
\centering\small\setlength{\tabcolsep}{4pt}
\caption{Per-block prototype-vote correlations for \TabICL{}-reg.
Pearson coefficient between block-$k$ vote prediction and
either the model's own prediction (pred) or the ground truth (truth).}
\label{tab:tabicl-reg-vote}
\resizebox{\linewidth}{!}{%
\begin{tabular}{lcccccccccccc}
Block & 0 & 1 & 2 & 3 & 4 & 5 & 6 & 7 & 8 & 9 & 10 & 11 \\
\midrule
pred  & $-0.22$ & $+0.40$ & $+0.44$ & $+0.27$ & $-0.52$ & $-0.28$ & $-0.23$ & $-0.43$ & $+0.64$ & $+0.73$ & $+0.08$ & $+0.67$ \\
truth & $-0.20$ & $+0.38$ & $+0.38$ & $+0.21$ & $-0.49$ & $-0.24$ & $-0.22$ & $-0.39$ & $+0.56$ & $+0.65$ & $+0.04$ & $+0.64$ \\
\end{tabular}}
\end{table}

\section{Permutation invariance: full results}
\label{app:invariance}

\subsection{Architectural sources of permutation sensitivity}
\label{app:inv-architecture}

\begin{table}[H]
\centering
\small
\caption{Component-level sources of permutation sensitivity. A \checkmark\
indicates the component is present and order-sensitive.}
\label{tab:inv-architecture}
\begin{tabular}{lcccc}
Model & \texttt{in\_linear} & Grouping & RoPE & \texttt{pos\_emb} ($W$, $b$) \\
\midrule
TabICL v1   & \checkmark & none     & \checkmark & --- \\
TabICL v1.1 & \checkmark & none     & \checkmark & --- \\
TabICL v2   & \checkmark & circular & \checkmark & --- \\
TabPFN      & ---        & \texttt{features\_per\_group=2} & --- & \checkmark \\
Mitra       & shared cell        & none     & ---        & --- \\
\end{tabular}
\end{table}

Table~\ref{tab:inv-architecture} catalogues the positional-encoding
components that break permutation invariance. \TabPFN{}'s positional
embedding maps each feature index $p$ through a learned $W \in
\mathbb{R}^{192 \times 48}$ and adds a learned bias $b \in
\mathbb{R}^{192}$; preprocessing (standardisation, outlier clipping,
quantile transforms) also runs in feature order. \TabICLvOne{} and v2
both apply RoPE along the feature dimension, and v2 additionally
enforces circular feature grouping; neither version encodes row
position. Row invariance is hard-coded; column invariance is broken by
$W,b$ in \TabPFN{} and by RoPE in \TabICLvOne{} and v2; label
invariance is broken by the learned label embeddings throughout.

\subsection{Synthetic invariance verifications}
\label{app:inv-synthetic}

\textbf{Protocol.} The grid spans $3$ seeds, $3$ dataset sizes, and $3$
configurations, with predictions counted as agreeing when they differ
by at most $0.005$. \TabICLvOne{} contributes baseline and no-RoPE
configurations; \TabICL{} contributes baseline, no-RoPE, and
no-RoPE under an arbitrary-permutation control. \TabPFN{} contributes
baseline, no positional embedding ($W{=}0$, $b$ retained), and the
latter under an arbitrary-permutation control.

Three residual symmetries surface within tolerance.
\TabPFN{} with $W = 0$ exhibits \emph{group-permutation invariance}:
pairs of features reorder freely, but pairs themselves do not split
across groups. \TabICL{} with RoPE removed exhibits
\emph{circular-shift invariance}. \TabICLvOne{} with RoPE removed
exhibits \emph{arbitrary-column invariance}. These outcomes match the
architecture mapping in Table~\ref{tab:route-architecture}: removing
one route exposes the residual symmetry of the route that remains ---
\TabPFN{}'s within-pair symmetry of feature grouping (M3, pair size
$2$), \TabICL{}'s cyclic symmetry of circular grouping, and, with
no M3 route in \TabICLvOne{}, arbitrary-column invariance at the cost
quantified in \sref{app:inv-icl}.

\begin{table}[H]
\centering\small\setlength{\tabcolsep}{4pt}
\caption{Synthetic invariance verifications: passes / $9$.}
\label{tab:inv-synthetic}
\begin{tabular}{llcccc}
Model variant & Test & det & row & col & label \\
\midrule
TabPFN baseline / $W{=}0$ & arb  & 9/9 & 9/9 & 0/9 & 4/9 \\
TabPFN baseline / $W{=}0$ & grp2 & 9/9 & 9/9 & 6/9 & 2/9 \\
TabICL v1                 & arb  & 9/9 & 9/9 & 0/9 & 1/9 \\
TabICL v1 no-RoPE        & arb  & 9/9 & 9/9 & \textbf{9/9} & 0/9 \\
TabICL v2 (\& no-RoPE)   & arb/grp3 & 9/9 & 9/9 & 0/9 & 0/9 \\
\end{tabular}
\end{table}

\subsection{Real-data invariance: TabPFN}
\label{app:inv-pfn}

Four configurations on $49$ datasets, $5$ seeds, $3$ column-permutation
trials: \emph{Baseline} (standard \TabPFN{} \citep{hollmann2025tabpfn});
\emph{$W{=}0$} (positional weight matrix zeroed, bias $b$ retained);
\emph{sign$(b)$} ($W{=}0$, bias replaced by $\operatorname{sign}(b)$);
\emph{no\_pe} ($W{=}0$, $b{=}0$, all positional parameters removed).

Setting $W{=}0$ shifts predictions by at most $0.001$ and preserves
accuracy ($0.854{\to}0.854$); replacing $b$ with $\operatorname{sign}(b)$
is also lossless ($0.855$); removing $b$ collapses accuracy to $0.338$,
proving $b$ acts as an architectural prior, not a positional code.

\begin{table}[H]
\centering\small\setlength{\tabcolsep}{4pt}
\caption{TabPFN per-dataset behavior:
$5$ datasets with the lowest column-label-agreement; accuracy is unchanged
under $W{=}0$ (mean acc $0.853$ baseline vs.\ $0.853$ at $W{=}0$, $n{=}45$).}
\label{tab:inv-pfn-perds}
\begin{tabular}{lccc}
Dataset & Col.\ agree & Acc.\ baseline & Acc.\ $W{=}0$ \\
\midrule
analcatdata\_dmft     & $78.0\%$ & $0.216$ & $0.216$ \\
cylinder-bands        & $90.4\%$ & $0.726$ & $0.726$ \\
steel-plates-fault    & $92.7\%$ & $0.794$ & $0.793$ \\
mfeat-zernike         & $92.7\%$ & $0.839$ & $0.838$ \\
vehicle               & $92.7\%$ & $0.851$ & $0.851$ \\
\end{tabular}
\end{table}

\subsection{Real-data invariance: \TabICLvOne{} and v2}
\label{app:inv-icl}

Each version is run in two configurations on $49$ datasets, $5$ seeds,
and $3$ trials: baseline and no-RoPE. \TabICL{} carries both the
M2 and M3 routes and tolerates RoPE removal; \TabICLvOne{} carries
only M2 and relies on RoPE to escape the multiset bound. The
$\approx 15$\,pp gap upper-bounds what RoPE buys as the sole
symmetry-breaker.

\begin{table}[H]
\centering\small\setlength{\tabcolsep}{4pt}
\caption{Per-dataset accuracy when RoPE is removed, $n{=}49$:
$5$ datasets where v1 loses the most.}
\label{tab:inv-icl-perds}
\begin{tabular}{lcccccc}
 & \multicolumn{3}{c}{TabICL v1} & \multicolumn{3}{c}{TabICL v2} \\
Dataset & base & no-RoPE & $\Delta$\,pp & base & no-RoPE & $\Delta$\,pp \\
\midrule
mfeat-karhunen   & $0.968$ & $0.387$ & $-58.1$ & $0.980$ & $0.979$ & $-0.1$ \\
mfeat-pixel      & $0.931$ & $0.392$ & $-54.0$ & $0.975$ & $0.975$ & $\phantom{-}0.0$ \\
semeion          & $0.858$ & $0.328$ & $-53.0$ & $0.956$ & $0.952$ & $-0.4$ \\
MiceProtein      & $0.998$ & $0.503$ & $-49.5$ & $1.000$ & $1.000$ & $\phantom{-}0.0$ \\
balance-scale    & $0.987$ & $0.493$ & $-49.4$ & $0.990$ & $0.990$ & $\phantom{-}0.0$ \\
\midrule
\textbf{Mean (all $49$)} & $0.851$ & $0.697$ & $\mathbf{-15.4}$ & $0.864$ & $0.862$ & $\mathbf{-0.1}$ \\
\textbf{\#\,losing $>5$\,pp} & \multicolumn{3}{c}{$27/49$} & \multicolumn{3}{c}{$0/49$} \\
\end{tabular}
\end{table}

\subsection{Real-data invariance summary}
\label{app:inv-real}

\begin{table}[H]
\centering
\small
\caption{Permutation agreement rates and accuracy
(49 datasets $\times$ 5 seeds $\times$ 3 trials). Column agreement below
100\% reflects order-sensitive preprocessing and grouping; label agreement
reflects class-ordering sensitivity.}
\label{tab:inv-real}
\begin{tabular}{lcccc}
Model     & Accuracy & Row agree.\ & Col.\ agree.\ & Label agree.\ \\
\midrule
TabPFN    & $0.854$  & $100.0\%$ & $98.2\%$ & $96.7\%$ \\
TabICL v2 & $0.865$  & $100.0\%$ & $98.2\%$ & $97.7\%$ \\
\end{tabular}
\end{table}

\subsection{One-vs-All label invariance}
\label{app:inv-ova}

Wrapping each model in a One-vs-All classifier achieves $100\%$
label-order agreement at zero accuracy cost.

\begin{table}[H]
\centering\small\setlength{\tabcolsep}{4pt}
\caption{One-vs-All wrapper, $n{=}21$ multi-class datasets.}
\label{tab:inv-ova}
\begin{tabular}{lccc}
Model+OvA & Acc. & Col.\ agree & Label agree \\
\midrule
TabPFN+OvA    & $0.877$ & $98.4\%$ & $100\%$ ($21/21$) \\
TabICL v1+OvA & $0.880$ & $97.2\%$ & $100\%$ ($21/21$) \\
TabICL v2+OvA & $0.890$ & $98.3\%$ & $100\%$ ($21/21$) \\
\end{tabular}
\end{table}

\Mitra{} is column-invariant by architecture but not class-invariant
in its shipped form (\sref{app:mitra-inv}). Wrapping it in OvA
recovers full class invariance at no accuracy cost on the same
multi-class real subset (extended to $n{=}24$ here for completeness;
$5$ seeds, $3$ trials, identical split protocol):
\Mitra{} v1 baseline $0.841$, \Mitra{} v1+OvA $0.841$, column agreement
$99.9\%$, label agreement effectively at $100\%$.

\subsection{Best/worst permutation spread}
\label{app:inv-bestworst}

\begin{figure}[H]
  \centering
  \includegraphics[width=0.7\linewidth]{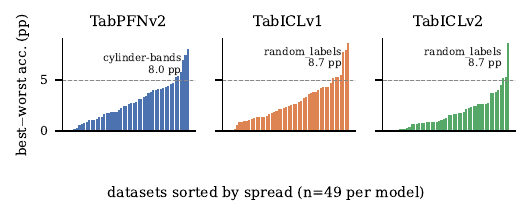}
  \caption{Per-dataset accuracy spread under random column
  permutations on the 49-dataset benchmark. For each dataset, the bar
  is (best accuracy across permutations)$-$(worst), measuring
  sensitivity to the semantically irrelevant ordering of input
  features. Datasets are sorted ascending within each model. Most
  sit below the $5$\,pp dashed line, but a tail reaches $\sim\!8.7$\,pp
  (worst dataset annotated). \emph{Takeaway:} the
  mechanism-aware remedies of \sref{sec:inv-mechanism} address this
  deployment-relevant tail risk at no accuracy cost.}
  \label{fig:col-perm-spread}
\end{figure}

For each dataset we record the best and worst accuracy across the
$5 \times 3 = 15$ column-permutation trials. Column-permutation spread
reaches $\sim\!8.0$\,pp on the worst real dataset (cylinder-bands;
Figure~\ref{fig:col-perm-spread}); the median is an order of magnitude
smaller (Table~\ref{tab:inv-bw}).

\begin{table}[H]
\centering\small\setlength{\tabcolsep}{4pt}
\caption{Best${-}$worst accuracy spread (pp) per model
($n{=}45$). Six TabPFN/v1 datasets
exceed the $5$\,pp threshold; only two for v2.}
\label{tab:inv-bw}
\begin{tabular}{lccccc|l}
Model & median & mean & Q1 & Q3 & max & Worst $3$ datasets (spread\,pp) \\
\midrule
TabPFN    & $2.67$ & $2.91$ & $1.39$ & $4.14$ & $8.02$ & cylinder-bands $8.02$, tic-tac-toe $7.47$, dresses-sales $7.00$ \\
TabICL v1 & $2.38$ & $2.74$ & $1.31$ & $4.06$ & $8.00$ & dresses-sales $8.00$, hill-valley $7.82$, cylinder-bands $5.56$ \\
TabICL v2 & $1.54$ & $1.73$ & $0.75$ & $2.57$ & $5.33$ & dresses-sales $5.33$, cylinder-bands $5.25$, ilpd $4.56$ \\
\end{tabular}
\end{table}

\subsection{Spectrum of invariance types}
\label{app:inv-spectrum}

\begin{table}[H]
\centering
\small
\caption{Spectrum of achievable permutation invariance types, the scope of
columns covered, and the accuracy cost of achieving it. Costs are mean
differences over 49 datasets $\times$ 5 seeds $\times$ 3 trials relative to
the respective baseline.}
\label{tab:inv-spectrum}
\begin{tabular}{llll}
Model & Invariance type & Scope & Accuracy cost \\
\midrule
TabICL v1 (no RoPE) & Arbitrary column permutations & All columns & $-15.4$\,pp \\
TabICL v2 (no RoPE) & Circular column shifts        & All columns & $-0.2$\,pp \\
TabPFN ($W=0$)      & Group permutations (groups of 2 features) & Within groups & $0$\,pp \\
\end{tabular}
\end{table}

Table~\ref{tab:inv-spectrum} catalogues the achievable invariance types
and their costs. Arbitrary-permutation invariance is reachable only in
\TabICLvOne{}, and only at $-15.4$\,pp. Narrower types --- circular
shifts for \TabICL{} and group permutations for \TabPFN{} ---
carry no measurable accuracy cost.

\subsection{Pretraining an invariant \TabICLvOne{} from scratch}
\label{app:inv-pretraining}

\paragraph{Setup.}
We retrain \TabICLvOne{} under the original stage-1 protocol of
\citet{qu2025tabicl}: $100$k optimization steps at batch size $512$,
identical synthetic prior, no stage-2 finetuning. Two architectural
switches are toggled independently, yielding four pretrained
checkpoints:
\begin{itemize}
\item \emph{RoPE} on/off along the feature axis. ``On'' is the
released \TabICLvOne{} configuration; ``off'' removes the rotary
positional embedding entirely (not zeroed at inference, but absent
during pretraining).
\item \emph{Class head}: standard class-conditional head (``no'',
released configuration) versus a class-invariant head (``yes''), in
which context labels enter as one-hot columns and a shared decoder
emits one logit per class. Permuting the class index permutes both
the conditioning columns and the output channels identically, so
the \texttt{argmax} prediction is exactly class-permutation
invariant by construction.
\end{itemize}
Evaluation follows the OpenML protocol of \citep{qu2025tabicl}.
Headline numbers reproduced in
Table~\ref{tab:tabicl_pretrain_results_main}.

\paragraph{Reading.}
Both invariance levers are essentially free in isolation but
compound to a $-2.6$\,pp gap when combined. This matches the
mechanism identified in \sref{sec:repr-collapse}: RoPE is
\TabICLvOne{}'s only within-row symmetry-breaker (one of the
``M2-only'' models in \sref{sec:repr-collapse-handcrafted}), so
removing it during pretraining without an architectural substitute
is the harder of the two ablations. The class-invariant head adds
an output-axis constraint that is, on its own, well-absorbed by
training. \Mitra{}'s shared cell embedding (\sref{app:mitra-inv})
shows that a column-invariant tabular foundation model
\emph{can} be trained at competitive accuracy when the architecture
itself supplies the symmetry from the start, rather than relying on
training to absorb its removal.

\section{Mechanism-grounded attacks: full results}\label{app:attacks}

This appendix reports the full results behind \sref{sec:attacks}. Each
attack tests a specific mechanistic prediction:
neighbourhood pollution targets the local label aggregation shared by
the two readouts (\sref{sec:tabpfn-mech},~\sref{sec:tabicl-mech});
geometry corruption targets the learned similarity of \sref{app:E1}
and the input geometry feeding \TabICL{}'s prototypes; null-space PGD
targets non-linear feature directions a linear baseline cannot access;
SVD-hiding tests whether \TabPFN{}'s preprocessing pipeline and
\TabICL{}'s distributed representation can be exploited by relocating
signal into discarded components. Mechanism transplantation
(\sref{sec:transplant}) is reported in \sref{app:K4}.

\subsection{Attack definitions and protocols}\label{app:attack-defs}

All attacks operate on the in-context window: they perturb the context
set seen at inference time and measure the resulting drop in
classification accuracy. Nine individual attacks are evaluated, organized
into four families.

\textbf{noise\_pad.}
Extra noisy context rows added (50\% padding); each padded row is
sampled from the per-feature empirical distribution. Floor on
robustness against context-set dilution.

\textbf{hub\_poison.}
Context examples (``hubs'') with the smallest average $k$-NN distance to
all other context points have their labels flipped. Hubs occupy dense,
class-ambiguous regions and disproportionately contaminate
nearest-neighbour readouts.

\textbf{centroid\_inj.}
A synthetic example injected at the class centroid of a target class
with the wrong label, pulling the effective centroid toward a competitor.

\textbf{boundary\_poison.}
Context examples within a margin of the decision boundary (estimated via
a held-out probe) have their labels flipped.

\textbf{mono\_cube, mono\_softexp, mono\_rank.}
Three monotone-function attacks that transform individual features while
preserving their rank order. \textsc{mono\_cube} applies a signed
cube-root transform; \textsc{mono\_softexp} applies a soft-exponential;
\textsc{mono\_rank} replaces values with their within-column rank. All
three break linear separability while leaving ordinal structure intact.

\textbf{svd\_hide.}
Signal injected along the smallest-singular-value SVD direction of the
context feature matrix into test queries but not context, creating a
query--context discrepancy invisible to any decoder attending only to
context features. Details in \S\ref{app:attack-svd}.

\textbf{Null-space PGD.}
PGD that moves context examples along the null space of the Ridge
regression weight matrix, so the perturbation is invisible to Ridge
while still shifting the geometry seen by TabPFN. Details in
\S\ref{app:attack-nullspace}.

\medskip
Hyperparameters (noise scale, hub count, margin, PGD step size and
budget) were fixed before any model evaluation. Each experiment reports
mean accuracy over the relevant dataset $\times$ seed grid; gaps
($\Delta$) are signed differences (model minus baseline) and 95\%
paired-bootstrap confidence intervals are computed over the dataset
dimension.

% ---------------------------------------------------------------------------
\subsection{TabPFN vs.\ Ridge, full table}\label{app:attacks-ridge}

\begin{table}[H]
\centering
\caption{TabPFN vs.\ Ridge under five mechanism-grounded attacks.
$\Delta$ columns report mean accuracy change (post-attack $-$ clean) on each
attack's evaluation grid. The differential (last column) is
$\Delta_\mathrm{TabPFN} - \Delta_\mathrm{Ridge}$; negative values mean TabPFN is
\emph{more} damaged. Sample sizes vary because some attacks have explicit
exclusion criteria (see \S\ref{app:attack-defs}, \S\ref{app:attack-nullspace}).}
\label{tab:attacks-ridge}
\begin{tabular}{lccc}
Attack & $\Delta_\mathrm{TabPFN}$ & $\Delta_\mathrm{Ridge}$ & Differential \\
\midrule
noise\_pad ($4\times$ padding, $n{=}49$)        & $-0.076$ & $-0.041$ & $-0.035$ \\
hub\_poison (top 15\% $k$-NN hubs, $n{=}49$)    & $-0.029$ & $-0.014$ & $-0.015$ \\
mono\_rank (rank transform, $n{=}49$)           & $-0.084$ & $-0.043$ & $-0.041$ \\
svd\_hide (real-data rotation, $n{=}40$)        & $-0.018$ & $-0.000$ & $-0.018$ \\
null-space PGD (\S\ref{app:attack-nullspace}, $n{=}29$) & $-0.047$ & $\phantom{-}0.000$ & $-0.047$ \\
\end{tabular}
\end{table}

The null-space attack is constructed in the null space of $W_\text{ridge}$,
so by design Ridge accuracy is unaffected; \TabPFN{} drops because it
attends to feature directions Ridge ignores.

% ---------------------------------------------------------------------------
\subsection{TabPFN vs.\ XGBoost, full table}\label{app:attacks-table}

\begin{table}[H]
\centering
\caption{TabPFN vs.\ XGBoost under nine conditions (49 cls $\times$ 10 seeds;
\textsc{mono\_softexp} $n{=}48$ after one numerically degenerate dataset is
dropped). Means are averages of per-dataset seed means.
$\Delta = \mathrm{TabPFN} - \mathrm{XGB}$ with 95\% bootstrap CI over
per-dataset paired differences. TabPFN dominates on every attack
($\Delta > 0$), with the smallest gap on \textsc{mono\_softexp}
($+0.014$) and the largest on \textsc{hub\_poison} ($+0.046$).}
\label{tab:attacks-xgb}
\begin{tabular}{lcccc}
Condition & TabPFN & XGBoost & $\Delta$ & 95\% CI \\
\midrule
baseline (clean)   & $0.854$ & $0.828$ & $+0.026$ & $[+0.010, +0.047]$ \\
noise\_pad         & $0.845$ & $0.821$ & $+0.024$ & $[+0.009, +0.045]$ \\
hub\_poison        & $0.828$ & $0.782$ & $+0.046$ & $[+0.030, +0.066]$ \\
centroid\_inj      & $0.855$ & $0.827$ & $+0.029$ & $[+0.015, +0.048]$ \\
boundary\_poison   & $0.843$ & $0.819$ & $+0.024$ & $[+0.009, +0.044]$ \\
mono\_cube         & $0.848$ & $0.828$ & $+0.020$ & $[+0.003, +0.041]$ \\
mono\_softexp      & $0.810$ & $0.796$ & $+0.014$ & $[+0.002, +0.025]$ \\
mono\_rank         & $0.756$ & $0.737$ & $+0.020$ & $[-0.014, +0.051]$ \\
svd\_hide          & $0.796$ & $0.758$ & $+0.039$ & $[+0.024, +0.055]$ \\
\end{tabular}
\end{table}

XGBoost is a non-linear, tree-structured baseline; the differential
absorbs the part of vulnerability shared by any flexible classifier.

% ---------------------------------------------------------------------------
\subsection{TabICL vs.\ XGBoost, full table}\label{app:attacks-tabicl}

\begin{table}[H]
\centering
\caption{TabICL vs.\ XGBoost under all eight attacks
($24$ classification datasets $\times$ $5$ seeds).
$\Delta = \mathrm{TabICL} - \mathrm{XGB}$ with $95\%$ bootstrap CI over
per-dataset paired differences. TabICL is at least as accurate as XGBoost on every
attack except \textsc{mono\_rank} (CI covers zero); the largest gap
is on \textsc{svd\_hide}.}
\label{tab:attacks-tabicl}
\begin{tabular}{lcccc}
Attack & XGBoost & TabICL & $\Delta$ & 95\% CI \\
\midrule
baseline (clean) & $0.862$ & $0.888$ & $+0.026$ & $[+0.001, +0.048]$ \\
noise\_pad       & $0.847$ & $0.882$ & $+0.035$ & $[+0.018, +0.054]$ \\
hub\_poison      & $0.814$ & $0.855$ & $+0.041$ & $[+0.014, +0.064]$ \\
centroid\_inj    & $0.858$ & $0.885$ & $+0.027$ & $[+0.007, +0.046]$ \\
boundary\_poison & $0.840$ & $0.877$ & $+0.037$ & $[+0.024, +0.052]$ \\
mono\_cube       & $0.862$ & $0.886$ & $+0.023$ & $[+0.004, +0.045]$ \\
mono\_softexp    & $0.846$ & $0.865$ & $+0.019$ & $[+0.001, +0.040]$ \\
mono\_rank       & $0.787$ & $0.787$ & $+0.000$ & $[-0.062, +0.042]$ \\
svd\_hide        & $0.752$ & $0.838$ & $+0.086$ & $[+0.051, +0.128]$ \\
\end{tabular}
\end{table}

The largest gap is on \textsc{svd\_hide} ($\Delta = +0.086$): XGBoost
collapses because its gradient-based training is sensitive to the injected
SVD direction while TabICL's L11 prototype readout is partially shielded by
its distributed, attention-averaged encoding.

% ---------------------------------------------------------------------------
\subsection{Tuned MLP baseline on the same attack grid}\label{app:attacks-mlp}

A two-hidden-layer ReLU MLP with dropout $0.1$ and AdamW
($\text{lr}{=}10^{-3}$, weight decay $10^{-4}$) is trained per
$(\mathrm{dataset}, \mathrm{seed})$ on the same context set as the
foundation models, with identical ordinal categorical encoding and
no further preprocessing. For each cell the hidden width is
selected from $(128,64), (256,128), (512,256)$ by held-out
validation accuracy on a $20\%$ split of the context set with
early stopping (patience $20$, max $200$ epochs); the chosen
configuration is then re-applied unchanged to every attack variant
of that cell. The MLP baseline supplies the non-linear neural
control that separates mechanism-specific from generic
non-linear-classifier vulnerability on the same $24$-classification
grid.

\begin{table}[H]
\centering\scriptsize
\setlength{\tabcolsep}{3pt}
\caption{Per-attack signed accuracy drop $\Delta$\,pp on the common
$24$-classification $\times 5$-seed grid for both foundation models, a tuned XGBoost,
and a tuned MLP ($2$ hidden layers selected per (dataset, seed) from
$(128,64)$, $(256,128)$, $(512,256)$ by validation accuracy; AdamW with
early stopping; identical ordinal categorical encoding). $95\%$
bootstrap CI over per-dataset seed means in brackets; $^{*}$ denotes
Holm-corrected significance at $\alpha{=}0.05$ across the eight
attacks per model. Clean baselines: TabPFNv2 $88.2\%$, TabICLv2
$88.8\%$, XGBoost $86.3\%$, MLP $85.0\%$.}
\label{tab:attacks-mlp}
\begin{tabular}{lcccc}
Attack            & TabPFNv2 & TabICLv2 & XGBoost & MLP \\
\midrule
noise\_pad        & $-1.5^{*}\,[-2.4,-0.6]$  & $-0.6\,\;[-1.5,+0.5]$ & $-1.6\,\;[-3.4,-0.1]$ & $-2.7^{*}\,[-3.9,-1.7]$ \\
hub\_poison       & $-3.9^{*}\,[-5.7,-2.2]$  & $-3.3^{*}\,[-4.8,-1.9]$ & $-4.8^{*}\,[-6.1,-3.4]$ & $-2.8^{*}\,[-4.0,-1.6]$ \\
centroid\_inj     & $-0.1\,\;[-0.7,+0.6]$  & $-0.3\,\;[-1.1,+0.5]$ & $-0.4\,\;[-0.9,+0.1]$ & $-0.7\,\;[-1.5,+0.2]$ \\
boundary\_poison  & $-1.9^{*}\,[-2.9,-1.1]$  & $-1.2\,\;[-2.2,-0.0]$ & $-2.3^{*}\,[-3.9,-0.9]$ & $-2.9^{*}\,[-4.2,-1.7]$ \\
mono\_cube        & $-1.1\,\;[-2.4,+0.1]$  & $-0.3\,\;[-1.5,+1.1]$ & $\;\;0.0\,\;[\;0.0,\;0.0]$ & $-4.8^{*}\,[-8.7,-1.7]$ \\
mono\_softexp     & $-2.2^{*}\,[-3.7,-0.8]$  & $-2.3^{*}\,[-3.8,-0.8]$ & $-1.7\,\;[-3.2,-0.3]$ & $-4.1^{*}\,[-5.9,-2.3]$ \\
mono\_rank        & $-8.0^{*}\,[-15.0,-2.4]$ & $-10.1^{*}\,[-18.7,-3.5]$ & $-7.6^{*}\,[-14.1,-2.5]$ & $-4.3^{*}\,[-8.4,-1.1]$ \\
svd\_hide         & $-8.3^{*}\,[-13.6,-3.7]$ & $-5.0^{*}\,[-8.8,-2.1]$ & $-11.1^{*}\,[-16.3,-6.3]$ & $-10.2^{*}\,[-17.6,-4.7]$ \\
\end{tabular}
\end{table}

The MLP attains a clean $85.0\%$ mean accuracy, lower than either
foundation model ($88.2\%$, $88.8\%$) but comparable to XGBoost
($86.3\%$). The relative attack drops sort the attacks into three
groups. (i) \emph{Mechanism-targeted}: \textsc{mono\_rank} drops
the foundation models by $8$--$10$\,pp while the MLP loses only
$4.3$\,pp, and \textsc{hub\_poison} drops the foundation models
by $3.3$--$3.9$\,pp while the MLP loses $2.8$\,pp; both confirm
the predictions (\sref{sec:attacks}) that distance-based retrieval
readouts are uniquely sensitive to inter-value-distance erasure
and that vote-style readouts are uniquely sensitive to flips of
high-degree context points. (ii) \emph{Shared with any flexible
classifier}: \textsc{boundary\_poison}, \textsc{noise\_pad},
\textsc{centroid\_inj} and \textsc{svd\_hide} hurt the MLP at
least as much as the foundation models, so the headline drops on
these attacks reflect sensitivity shared by any flexible
classifier rather than a readout-specific weakness.
(iii) \emph{Absorbed by the ICL prior}: \textsc{mono\_cube} and
\textsc{mono\_softexp} hurt the MLP roughly twice as much as the
foundation models, consistent with the foundation models'
Bayesian/ICL-style averaging absorbing smooth feature warps that
a freshly fitted MLP cannot. The Holm-corrected significance pattern is preserved on
both foundation models when the six \texttt{mfeat-*} datasets are
removed; \textsc{svd\_hide} drops shrink (TabPFNv2
$-8.3 \to -3.5$\,pp; MLP $-10.2 \to -5.0$\,pp; TabICLv2
$-5.0 \to -3.0$\,pp), reflecting the digit-family's particular
sensitivity to the SVD construction.

% ---------------------------------------------------------------------------
\subsection{Multiplicity controls and baseline summary}\label{app:attacks-stats}

\textbf{Holm correction.} Within each model the eight perturbations
of Table~\ref{tab:attacks-drops} are corrected at $\alpha{=}0.05$
via Holm's step-down procedure. The corrected significance pattern
keeps the rank/hub/SVD trio significant on \TabPFN{} and \TabICL{}
and the hub/soft-exp/SVD trio significant on \Mitra{}; diffuse
perturbations (centroid injection, cube warp) are not significant
for any model, and noise padding is significant only on \TabPFN{}. The pattern is preserved for
\TabPFN{} and \TabICL{} when the six \texttt{mfeat-*} datasets are
removed, with reduced \textsc{svd\_hide} drops as documented in
\sref{app:attacks-mlp}.

\textbf{Wilcoxon signed-rank.} A pre-specified one-sided Wilcoxon
signed-rank test on the per-dataset paired difference between
(rank warp, hub poison) and (cube warp, centroid injection)
rejects at $p<10^{-3}$ on \TabPFN{} and \TabICL{}; on \Mitra{}
only the hub component contributes ($p<10^{-3}$ on (hub, SVD)
vs.\ (cube, centroid)).

\textbf{Baseline summary.} On the $24$-grid \TabPFN{} and
\TabICL{} remain at least as accurate as a tuned XGBoost after
every perturbation except SVD burial, where XGBoost itself loses
$-11.1$\,pp. On the $16$-dataset mixed-type slice
(\sref{app:attacks-mixed16}, \sref{app:attacks-catboost}),
native-categorical CatBoost reaches $76.7\%$ and XGBoost $77.3\%$;
\TabPFN{} ties and \TabICL{} stays ahead at $78.9\%$. A tuned MLP
refit on the same poisoned context attains a clean $85.0\%$
(\sref{app:attacks-mlp}); the rank-warp and hub-poisoning gaps are
the diagnostic ones---the MLP loses $4.3$\,pp under rank warp
while \TabPFN{}/\TabICL{} lose $8$--$10$\,pp, and the MLP loses
$2.8$\,pp under hub poisoning while \TabPFN{}/\TabICL{} lose
$3.3$--$3.9$\,pp, confirming the mechanistic prediction that
distance-based vote and prototype readouts are uniquely sensitive
to inter-value-distance erasure and high-degree label flips. The
null-space PGD diagnostic (\sref{app:attack-nullspace}) leaves
Ridge unchanged by construction but still costs \TabPFN{}
$-4.7$\,pp, ruling out any single linear readout in the Ridge
feature basis.

% ---------------------------------------------------------------------------
\subsection{Native-categorical CatBoost clean control}\label{app:attacks-catboost}

XGBoost's ordinal handling of categorical columns may understate a
modern tree baseline on this grid. We rerun the
clean $24$-dataset classification grid with CatBoost using native
categorical handling wherever the OpenML metadata marks categorical
columns, the same $80/20$ split as the main
experiments, and a small validation-tuned GPU grid over depth
$\{4,6\}$, learning rate $\{0.03,0.10\}$, and iterations
$\{500,1000\}$ with early stopping. We do not rerun the attack
perturbations with CatBoost: several attack operators are defined only
on the shared ordinal-encoded numeric representation, so this control is
clean-only.

\begin{table}[H]
\centering\small\setlength{\tabcolsep}{6pt}
\caption{Clean-only comparator means on the same $24$-dataset attack
grid. The mixed-type subset contains the four datasets with at least one
categorical column (\texttt{car}, \texttt{cylinder-bands},
\texttt{dresses-sales}, \texttt{eucalyptus}).}
\label{tab:catboost-clean24}
\begin{tabular}{lcc}
Model & 24-grid clean mean & mixed-type subset mean \\
\midrule
TabPFN & $0.882$ & $0.778$ \\
TabICL & $0.888$ & $0.819$ \\
XGBoost & $0.863$ & $0.770$ \\
CatBoost & $0.860$ & $0.752$ \\
MLP & $0.850$ & $0.749$ \\
\end{tabular}
\end{table}

Only four of the $24$ datasets on this grid contain categorical
features, so the native-categorical advantage is a narrow probe. On
this suite CatBoost lands close to tuned XGBoost overall and does not
overtake either foundation model on the mixed-type subset; the
clean-baseline ordering holds. The $16$-dataset slice below extends
this evidence to a broader mixed-type benchmark.

The four-dataset mixed-type subset is narrow. We rerun the same
native-categorical CatBoost protocol on every classification dataset in
the $49$-dataset benchmark whose OpenML metadata marks at least one
categorical column, and add a matching native-categorical XGBoost control
on the same fixed slice. This mixed-type slice contains the $16$ datasets
enumerated in \S\ref{app:reproduce-detail}. The clean ordering holds
(Tab.~\ref{tab:catboost-mixed16}): \TabICL{} stays ahead of CatBoost
on $14/16$ datasets (one tie), with a mean paired advantage of
$+2.3$\,pp and $95\%$ CI $[+1.1,+3.5]$; \TabPFN{} lands closer
($+0.8$\,pp, CI $[-0.2,+1.6]$) and loses on $5/16$. Native-categorical
XGBoost falls between the two tree baselines at $77.3\%$ mean accuracy:
\TabPFN{} ties overall ($+0.2$\,pp, CI $[-1.7,+1.8]$), and \TabICL{}
leads by $+1.7$\,pp (CI $[+0.5,+2.9]$). This extends the mixed-type
clean-baseline evidence well beyond the four-dataset probe; attack-time
mixed-type behavior is addressed in \S\ref{app:attacks-mixed16}.

\begin{table}[H]
\centering\small\setlength{\tabcolsep}{6pt}
\caption{Broader native-categorical tree controls on the $16$
classification datasets in the $49$-dataset benchmark with at least one
categorical column ($5$ seeds / dataset). TabPFN and TabICL means use
the same $80/20$ split as the $49$-dataset readout-grid evaluations.
$\Delta = \mathrm{model} -
\mathrm{CatBoost}$ with $95\%$ bootstrap CI over per-dataset paired
differences.}
\label{tab:catboost-mixed16}
\begin{tabular}{lcc}
Model & mixed-type clean mean & $\Delta$ vs.\ CatBoost [95\% CI] \\
\midrule
TabPFN & $0.774$ & $+0.008\,[-0.002,+0.016]$ \\
TabICL & $0.789$ & $+0.023\,[+0.011,+0.035]$ \\
XGBoost (native-categorical) & $0.773$ & $+0.006\,[-0.010,+0.023]$ \\
CatBoost & $0.767$ & --- \\
\end{tabular}
\end{table}

% ---------------------------------------------------------------------------
\subsection{Broader mixed-type attack slice}\label{app:attacks-mixed16}

The clean-only controls above establish baseline breadth; attack-time
mixed-type behavior is the next question. We rerun the three
highest-signal attacks from the main text---hub poison, rank warp,
and SVD burial---on the full $16$-dataset mixed-type slice under the
same shared ordinal-encoded numeric view used by the foundation models
($5$ seeds / dataset). The
attack-pipeline clean means on this slice are $0.781$ for \TabPFN{}
and $0.791$ for \TabICL{}. The same ordering survives
(Tab.~\ref{tab:mixed16-attacks}): hub poison costs $2$--$3$\,pp,
rank warp dominates for both models, and SVD burial is small on
average. Rank- and hub-style attacks remain the strongest
within-design predictor of where the foundation models lose ground,
including on mixed-type data.

\begin{table}[H]
\centering\small\setlength{\tabcolsep}{6pt}
\caption{Broader mixed-type attack slice on the $16$ mixed-type
classification datasets under the shared ordinalised numeric view
($5$ seeds / dataset). $\Delta$ is the within-model pp change from the
clean mixed16 attack-pipeline baseline.}
\label{tab:mixed16-attacks}
\begin{tabular}{lcc}
Attack & \TabPFN{} $\Delta$\,pp [95\% CI] & \TabICL{} $\Delta$\,pp [95\% CI] \\
\midrule
hub poison & $-2.4\,[-3.7,-1.4]$ & $-2.8\,[-4.1,-1.5]$ \\
rank warp & $-17.2\,[-28.9,-7.1]$ & $-21.0\,[-34.5,-10.2]$ \\
SVD burial & $-0.2\,[-0.7,+0.5]$ & $-1.0\,[-2.2,-0.1]$ \\
\end{tabular}
\end{table}

% ---------------------------------------------------------------------------
\subsection{Null-space PGD details}\label{app:attack-nullspace}

The null space of the Ridge weight matrix $N(W_\text{ridge})$ is
computed via a thin SVD of $W_\text{ridge}^\top$; columns of $V$ with
singular values below $10^{-6} \cdot \sigma_{\max}$ span the
near-null space.

\textbf{Dataset exclusion.}
Three datasets excluded for zero null-space dimension. A further $17$
excluded as numerically degenerate (fewer than five directions pass the
threshold).

\textbf{PGD hyperparameters.}
$\eta = 0.01$ (in unit-variance feature units), $T = 200$ steps,
$\ell_2$ budget $\varepsilon = 1.0$. After each step, the perturbation
is projected onto the null space (so Ridge sees no change) and clipped
to the $\ell_2$ ball. Untargeted: maximize the cross-entropy of the
Ridge prediction distribution under TabPFN, using Ridge's soft output
as a fixed proxy.

% ---------------------------------------------------------------------------
\subsection{SVD-hiding details and per-dataset breakdown}\label{app:attack-svd}

\textbf{Mechanism.}
Let $[X_\text{ctx}; X_\text{query}] \in \mathbb{R}^{(n+m) \times d}$
be the stacked context-and-query feature matrix. A thin SVD yields
singular values $\sigma_1 \ge \dots \ge \sigma_d$. We divide the top
$k = \lceil d/4 \rceil$ singular values by $10$ and multiply the
remaining $d-k$ by $10$, then reconstruct and split back into context
and query. This dampens the high-variance components that
similarity-and-vote and prototype readouts rely on and amplifies the
low-variance components most decoders ignore, hitting context and
query consistently.

\textbf{Three-copy preprocessing redundancy.}
TabPFN's three-copy preprocessing (original, rank-normalized,
quantile-normalized features) re-bases the geometry across copies;
the high-variance components in raw feature space are not the
high-variance components in the encoder's input space, so the
reweighting produces only partial perturbation of TabPFN's actual
input.

\textbf{Real vs.\ synthetic comparison.}
On real datasets, the svd\_hide gap between TabICL and XGBoost
($\Delta = +0.086$, \S\ref{app:attacks-tabicl}) is substantially larger
than on synthetic datasets where the low-variance direction is by
construction uninformative. The attack exploits genuine low-variance
signal that XGBoost uses (via split-based selection) but TabICL's L11
prototype readout discards.

\textbf{Per-dataset breakdown.}
On $34/49$ datasets XGBoost is more damaged than TabPFN (mean drops
$-0.058$ vs.\ $-0.070$, differential $+0.013$). On hill-valley TabPFN
scores $0.593$ on raw features but $\{0.934, 0.926, 0.918, 0.914\}$ when
projected to top-$\{10,20,50,100\}$ PCs (SVD-projected variant of hill-valley).

\begin{table}[H]
\centering\small\setlength{\tabcolsep}{4pt}
\caption{\textsc{svd\_hide}: five datasets with the largest XGBoost drop.}
\label{tab:svd-perds}
\begin{tabular}{lcccccc}
Dataset & TabPFN clean & TabPFN svd & $\Delta_\mathrm{TabPFN}$ & XGBoost clean & XGBoost svd & $\Delta_\mathrm{XGBoost}$ \\
\midrule
mfeat-karhunen & $0.963$ & $0.613$ & $-0.350$ & $0.925$ & $0.508$ & $-0.418$ \\
mfeat-pixel    & $0.960$ & $0.540$ & $-0.420$ & $0.948$ & $0.603$ & $-0.345$ \\
mfeat-fourier  & $0.895$ & $0.585$ & $-0.310$ & $0.860$ & $0.557$ & $-0.302$ \\
semeion        & $0.918$ & $0.718$ & $-0.201$ & $0.897$ & $0.611$ & $-0.285$ \\
sign\_1d       & $0.980$ & $0.840$ & $-0.140$ & $1.000$ & $0.720$ & $-0.280$ \\
\end{tabular}
\end{table}

% ---------------------------------------------------------------------------
\subsection{Side-by-side TabPFN/TabICL contrast table}\label{app:contrast-table}

\begin{table}[H]
\centering
\small
\caption{Mechanistic comparison of TabPFN \citep{hollmann2025tabpfn} and
TabICL \citep{qu2025tabicl} across five dimensions.}
\label{tab:contrast}
\setlength{\tabcolsep}{4pt}
\begin{tabular}{p{0.18\linewidth}p{0.36\linewidth}p{0.36\linewidth}}
Dimension & TabPFN & TabICL \\
\midrule
Critical early block
  & Block-0: encodes formats \& shapes (\S\ref{sec:critical-block})
  & ColEmb-2: SetTransformer readout \\[2pt]
Attention role
  & Sharp \& locally necessary at L9 (\S\ref{app:causal-attn})
  & Individually redundant; jointly necessary (\S\ref{app:tabicl-uniform-all}) \\[2pt]
Similarity tracking
  & $r = 0.34$ (attn.\ vs.\ query--context sim.; \S\ref{app:E1})
  & $r < 0.08$ at every block (\S\ref{app:tabicl-distance}) \\[2pt]
Rank profile
  & Early bottleneck (\S\ref{sec:rank-profile})
  & Late prototype-formation phase (\S\ref{sec:rank-profile}) \\[2pt]
Operational readout
  & Attention-weighted label vote (\S\ref{sec:tabpfn-vote})
  & Nearest-class prototype over L11 reps.\ (\S\ref{sec:tabicl-prototype}) \\
\end{tabular}
\end{table}

Table~\ref{tab:contrast} summarizes the five mechanistic dimensions that
distinguish the two architectures (referenced from \sref{sec:transplant}).

\section{Additional experiments}\label{app:additional}

All experiments in this appendix are inference-only on the same model
releases as the rest of the paper; none require pretraining
(\S\ref{sec:limit-no-pretraining}).
Statistical reporting follows \S\ref{app:stats-protocol}.

\subsection{TabICL-reg L11 kNN}\label{app:K2}

Regression analogue of \S\ref{app:knn-tabicl}: 10 reg datasets $\times$ 5
seeds, four distance spaces, four neighbour counts. Each cell reports the
across-dataset mean of (Pearson $r$ between kNN-imputed and TabICL-reg
prediction $\,/\,$ kNN $R^2$ against the ground truth).

\begin{table}[H]
\centering\small\setlength{\tabcolsep}{4pt}
\caption{TabICL-reg kNN agreement / kNN $R^2$ aggregated over 10 regression datasets, 5 seeds. Rep-$\ell_2$ at L11 dominates input-$\ell_2$ at every $k$; the kNN-in-final-rep finding holds for regression as for classification.}
\label{tab:K2-tabicl-reg-knn}
\begin{tabular}{lcccc}
$k$ & input-$\ell_2$ & rep-$\ell_2$ (L11) & rep-cosine (L11) & rep-Mahalanobis (L11) \\
\midrule
$k{=}1$  & 0.603 / $-$0.375 & 0.784 / 0.542 & 0.719 / 0.459 & 0.207 / $-$0.218 \\
$k{=}3$  & 0.665 / 0.286    & 0.842 / 0.575 & 0.781 / 0.504 & 0.324 / 0.017    \\
$k{=}5$  & 0.681 / 0.386    & \textbf{0.859 / 0.577} & 0.801 / 0.515 & 0.360 / 0.044 \\
$k{=}10$ & 0.689 / 0.433    & 0.871 / 0.566 & 0.830 / 0.519 & 0.382 / 0.036 \\
\end{tabular}

\end{table}

\subsection{TabICL-reg per-block vote refutation}\label{app:K3}

Regression analogue of \S\ref{app:tabicl-vote-refut}: per-block pseudo-vote
$\hat{y}_L = \sum_i \alpha_i\,y_i$ from head-mean test\,$\to$\,train
attention. 10 reg datasets $\times$ 5 seeds.

\begin{table}[H]
\centering\small\setlength{\tabcolsep}{4pt}
\caption{TabICL-reg per-block weighted-label vote, mean across 10 datasets.
Pearson is correlation between $\hat{y}_L$ and the model's own prediction
(or ground truth). Best layer L9 ($r=0.732$) is far below the TabPFN-reg
ceiling $r\geq 0.896$; six of twelve layers are negative. Weighted-label vote is not the dominant TabICL-reg mechanism.}
\label{tab:K3-tabicl-reg-vote}
\begin{tabular}{lccc|lccc}
Layer & Pearson(pred) & Pearson(truth) & vote $R^2$ &
Layer & Pearson(pred) & Pearson(truth) & vote $R^2$ \\
\midrule
L0 & $-$0.222 & $-$0.199 & $-$0.038 & L6  & $-$0.231 & $-$0.221 & $-$0.019 \\
L1 & $+$0.396 & $+$0.384 & $+$0.006 & L7  & $-$0.429 & $-$0.391 & $-$0.033 \\
L2 & $+$0.442 & $+$0.380 & $+$0.001 & L8  & $+$0.642 & $+$0.557 & $+$0.032 \\
L3 & $+$0.265 & $+$0.213 & $-$0.006 & L9  & $\mathbf{+0.732}$ & $+$0.647 & $+$0.064 \\
L4 & $-$0.523 & $-$0.485 & $-$0.029 & L10 & $+$0.076 & $+$0.040 & $-$0.008 \\
L5 & $-$0.276 & $-$0.243 & $-$0.027 & L11 & $+$0.672 & $+$0.643 & $+$0.088 \\
\end{tabular}

\end{table}

\subsection{Mechanism transplantation}\label{app:K4}

Direct causal test of \S\ref{sec:transplant}, 49 cls $\times$ 5 seeds.
Each row is a per-dataset paired comparison; we report the across-dataset
means and a bootstrap 95\,\% CI over per-dataset paired differences
(10$^4$ resamples) on $\Delta$.
Rows (iii) and (vi) add a minimal host-side control: logistic
regression on the frozen host representation.

\textbf{Four conditions.}
\begin{itemize}
  \item[(i)]  TabPFN native vote (L9).
  \item[(ii)] TabPFN with prototype readout: $\hat{p}(c) \propto
        \exp(-\|h_{L11} - \mu_c\|_2^2)$, $\mu_c$ unweighted per-class
        means of L11 representations of all training rows.
  \item[(iii)] TabICL native prototype (L11).
  \item[(iv)] TabICL with vote readout: at the sharpest-attention block
        $b^*(D)$ selected per dataset $D$, $\hat{p}(c) =
        \frac{1}{H}\sum_{h}\sum_{i} \alpha^{(i)}_h \mathbf{1}[y_i = c]$.
\end{itemize}

\begin{table}[H]
\centering
\caption{Bootstrap CIs for the cross-transplant and fitted-host
linear-head conditions of Table~\ref{tab:transplant-summary} (49 cls
$\times$ 5 seeds). CI: 95\% paired-bootstrap over datasets. Naive
cross-transfer is catastrophic on both sides, but the fitted-host
recovery is asymmetric: \TabPFN{} remains far below native, whereas
\TabICL{} nearly recovers.}
\label{tab:transplant-fitted-ci}
\begin{tabular}{llcc}
Condition & Description & Accuracy & vs.\ native $\Delta$ (95\% CI) \\
\midrule
(i)   & TabPFN native          & $0.854$ & --- \\
(ii)  & TabPFN $+$ prototype   & $0.523$ & $-0.331\ [-0.412, -0.253]$ \\
(iii) & TabPFN $+$ linear head & $0.617$ & $-0.237\ [-0.305, -0.172]$ \\
(iv)  & TabICL native          & $0.864$ & --- \\
(v)   & TabICL $+$ vote        & $0.470$ & $-0.395\ [-0.484, -0.307]$ \\
(vi)  & TabICL $+$ linear head & $0.859$ & $-0.005\ [-0.009, -0.001]$ \\
\end{tabular}
\end{table}

The naive cross-transplants are catastrophic and the CIs do not cover
zero (Table~\ref{tab:transplant-fitted-ci}). The fitted-host control
is asymmetric: TabPFN remains far below native while TabICL nearly
recovers (within $1$\,pp on $32/49$ datasets), so the stronger
non-portability claim is asymmetric.

\textbf{Per-dataset breakdown.}
Both readouts collapse on multi-feature digit datasets and synthetic
mixtures (Table~\ref{tab:transplant-perds}).

\begin{table}[H]
\centering\small\setlength{\tabcolsep}{4pt}
\caption{Transplant accuracies on the five datasets with the largest
TabICL drop $\Delta_\mathrm{ICL}$.}
\label{tab:transplant-perds}
\begin{tabular}{lcccccc}
Dataset & PFN nat. & PFN proto & $\Delta_\mathrm{PFN}$ & ICL nat. & ICL vote & $\Delta_\mathrm{ICL}$ \\
\midrule
mfeat-karhunen & $0.969$ & $0.161$ & $-0.808$ & $0.984$ & $0.043$ & $-0.941$ \\
quadrant\_2d   & $0.992$ & $0.412$ & $-0.580$ & $0.996$ & $0.080$ & $-0.916$ \\
mfeat-factors  & $0.961$ & $0.236$ & $-0.725$ & $0.984$ & $0.089$ & $-0.895$ \\
mfeat-pixel    & $0.962$ & $0.254$ & $-0.709$ & $0.979$ & $0.084$ & $-0.895$ \\
semeion        & $0.909$ & $0.277$ & $-0.632$ & $0.961$ & $0.076$ & $-0.885$ \\
\end{tabular}
\end{table}

\textbf{Best-layer transplant.}
Picking the layer maximizing held-out accuracy per dataset
(best-layer transplant, $49$ datasets) shrinks the gap for
TabICL ($-0.395\!\to\!-0.263$, acc.\ $0.470\!\to\!0.601$) but
\emph{widens} it for TabPFN ($-0.331\!\to\!-0.549$,
acc.\ $0.523\!\to\!0.305$): the prototype-optimal layer is shallow
(median $0$) where TabPFN representations have not yet specialized.

\subsection{TabPFN L11 kNN agreement (strengthened test)}\label{app:K1}

Restated from \S\ref{app:knn-vs-tabpfn} at the readout layer L11
(49 cls $\times$ 5 seeds). Each cell is across-dataset mean kNN agreement
with TabPFN.

\begin{table}[H]
\centering\small\setlength{\tabcolsep}{4pt}
\caption{TabPFN kNN agreement at the readout layer L11. Every L11 metric is
strictly worse than L9 and worse than input-$\ell_2$ at every $k$. The
``kNN-on-final-representation'' hypothesis remains refuted at the readout
layer.}
\label{tab:K1-tabpfn-l11-knn}
\begin{tabular}{lccccc}
$k$ & input-$\ell_2$ & rep-$\ell_2$ L9 & rep-$\ell_2$ L11 & rep-cosine L11 & rep-Mahalanobis L11 \\
\midrule
$k{=}1$  & 0.782 & 0.768 & 0.528 & 0.528 & 0.356 \\
$k{=}3$  & 0.800 & 0.772 & 0.554 & 0.554 & 0.378 \\
$k{=}5$  & \textbf{0.811} & 0.774 & 0.561 & 0.561 & 0.395 \\
$k{=}10$ & 0.823 & 0.777 & 0.572 & 0.572 & 0.429 \\
\end{tabular}

\end{table}

\subsection{TabICL extra-attack robustness}\label{app:K6}

Three perturbations beyond the eight reported in \S\ref{sec:attacks}.
24 datasets $\times$ 5 seeds; $\Delta = \text{TabICL} - \text{XGB}$.

\begin{table}[H]
\centering\small\setlength{\tabcolsep}{4pt}
\caption{TabICL never loses on the mean across the three additional
attacks; consistent with the 5-attack ranking in the main text and with
the 8-attack panel as a whole.}
\label{tab:K6-attacks-tabicl-extra}
\begin{tabular}{lcccc}
Attack & TabICL acc & XGB acc & $\Delta$ (mean) & TabICL wins \\
\midrule
\texttt{boundary\_poison} & 0.877 & 0.840 & $+$0.037 & 22 / 24 \\
\texttt{mono\_cube}       & 0.886 & 0.862 & $+$0.023 & 18 / 24 \\
\texttt{mono\_rank}       & 0.787 & 0.787 & $+$0.000 & 16 / 24 \\
\end{tabular}

\end{table}

\subsection{Column-marginal vs.\ invariance-cost correlation}\label{app:K7}

Post-hoc, CPU-only computation for \S\ref{app:repr-collapse-acc}.
For each dataset compute (i) size of largest column-set with identical 1D
marginals at JS tolerance $\tau \in \{0.01, 0.05, 0.10\}$,
(ii) corresponding invariance-ablation gap. Spearman $\rho$ across 49
datasets.

\begin{table}[H]
\centering\small\setlength{\tabcolsep}{4pt}
\caption{Spearman correlation between column-marginal-duplication count
and per-dataset invariance gap. Only TabICL-v1 no-RoPE exhibits a
non-zero population gap that is positively correlated with marginal
duplication; TabPFN $W{=}0$ is null in both. The JS-screen identifies
collapse-prone datasets a-priori.}
\label{tab:K7-marginal-invariance}
\begin{tabular}{lcccc}
Model & mean gap & $\rho$ ($\tau{=}0.01$) & $\rho$ ($\tau{=}0.05$) & $\rho$ ($\tau{=}0.10$) \\
\midrule
TabPFN $W{=}0$            & $-$0.0001 & $+$0.042 (p{=}0.78) & $+$0.081 (p{=}0.58) & $+$0.064 (p{=}0.66) \\
TabICL-v1 no-RoPE         & $+$0.1540 & $+$0.201 (p{=}0.17) & $+$0.316 (p{=}0.03) & $+$0.322 (p{=}0.02) \\
TabICL-v2 no-RoPE         & $+$0.0016 & $+$0.231 (p{=}0.11) & $+$0.310 (p{=}0.03) & $+$0.325 (p{=}0.02) \\
\end{tabular}

\end{table}

\subsection{Collapse $f$-sweep}\label{app:K8}

Parametric extension of \S\ref{sec:repr-collapse-doubleablation}.
Synthesise \citet{qu2025tabicl} style stress data where the fraction
$f \in \{0, 0.25, 0.5, 0.75, 1\}$ of $d{=}12$ columns are i.i.d.\ from a
single shared discrete marginal, the remainder from heterogeneous
marginals; $K{=}5$ classes, $n{=}800$ rows, 10 seeds each. Cell entries
are paired $\Delta$ (ablated $-$ full) with bootstrap 95\,\% CI.

\begin{table}[H]
\centering\small\setlength{\tabcolsep}{3pt}
\caption{Collapse $f$-sweep paired $\Delta$ accuracy. TabICL-v1 no-RoPE
falls monotonically with $f$ (with non-overlapping CIs above
$f{=}0.25$). TabPFN $W{=}0$ and TabICL-v2 no-RoPE stay within
$|\Delta| \leq 0.001$. TabPFN 2nd-slot$=0$ shows a milder M3-pair-channel
load-bearing effect that grows from $-0.046$ at $f{=}0$ to $-0.140$ at
$f{=}1$.}
\label{tab:K8-collapse-fsweep}
\resizebox{\linewidth}{!}{%
\begin{tabular}{lcccc}
$f$ & TabPFN $W{=}0$ & TabPFN $\textnormal{2nd-slot}{=}0$ & TabICL-v1 no-RoPE & TabICL-v2 no-RoPE \\
\midrule
0.00 & $-$0.000 $[-0.001,+0.000]$ & $-$0.046 $[-0.096,-0.004]$ & $\mathbf{-0.255}\,[-0.318,-0.187]$ & $-$0.001 $[-0.004,+0.001]$ \\
0.25 & $-$0.000 $[-0.001,+0.001]$ & $-$0.017 $[-0.041,+0.000]$ & $\mathbf{-0.254}\,[-0.311,-0.190]$ & $-$0.000 $[-0.001,+0.001]$ \\
0.50 & $-$0.001 $[-0.002,+0.000]$ & $-$0.061 $[-0.108,-0.022]$ & $\mathbf{-0.459}\,[-0.474,-0.443]$ & $-$0.001 $[-0.003,+0.001]$ \\
0.75 & $-$0.000 $[-0.001,+0.000]$ & $-$0.062 $[-0.105,-0.021]$ & $\mathbf{-0.469}\,[-0.488,-0.450]$ & $-$0.000 $[-0.003,+0.002]$ \\
1.00 & $+$0.000 $[+0.000,+0.001]$ & $\mathbf{-0.140}\,[-0.186,-0.102]$ & $\mathbf{-0.472}\,[-0.489,-0.448]$ & $-$0.000 $[-0.002,+0.002]$ \\
\end{tabular}}
\end{table}

\subsection{Combined within-row ablation on v2 backbones}\label{app:K8b}

Each individual within-row ablation on a v2 backbone is load-free
(\sref{sec:repr-collapse-doubleablation},
Table~\ref{tab:K8-collapse-fsweep}). To test whether v2 carries
\emph{redundant} within-row defenses, we ablate both available routes
simultaneously, without retraining: \TabPFN{} with
$W{=}0$ and pair second-slot zeroed; \TabICL{} with RoPE removed
and the slots $1, 2$ of every $(B, T, G, 3)$ circular feature group
zeroed (slot $0$ of each group, the original feature index, is
preserved; the trained \texttt{in\_linear} weights on slots $1, 2$
are still applied to zeros). Both ablations are wrapped in
\texttt{VerifiedPatch}: forward hooks confirm \texttt{tf\_row.rope}
is \texttt{None} at every block call and that the post-grouping
slots $1{:}\,$ are zero on every forward pass. Twenty seeds with
paired-bootstrap CIs over seeds; per-seed re-resampling on the
synthetic.

\begin{table}[H]
\centering\small
\caption{Combined within-row ablation on v2 backbones. Each individual
ablation is load-free (Table~\ref{tab:K8-collapse-fsweep}); the
combined ablations expose latent collapse. \TabICL{} with both
within-row routes off lands at the majority baseline on the $d{=}12$
balance-like stress, confirming that its individual-ablation
``free'' result reflected redundancy rather than the absence of a
within-row defense. \TabPFN{} with both routes off survives but
degrades. Compare to the \Mitra{} row-attention ablation
(Table~\ref{tab:mitra-collapse-ablation}, $0.95 \to 0.08$): \Mitra{}'s
single load-bearing mechanism has no architectural fall-back, so its
failure mode is below-majority rather than at-majority.}
\label{tab:K8b-v2-combined}
\begin{tabular}{lllll}
backbone & ablation & dataset & accuracy & $\Delta$ vs.\ full ($95\%$ CI) \\
\midrule
\TabPFN{} & full                              & balance-scale         & $0.962$ & --- \\
\TabPFN{} & $W{=}0\,+\,$2nd-slot$=0$          & balance-scale         & $0.890$ & $-0.071\ [-0.106, -0.044]$ \\
\TabPFN{} & full                              & balance-like, $d{=}12$ & $0.970$ & --- \\
\TabPFN{} & $W{=}0\,+\,$2nd-slot$=0$          & balance-like, $d{=}12$ & $0.855$ & $-0.115\ [-0.144, -0.091]$ \\
\midrule
\TabICL{} & full                              & balance-scale         & $0.978$ & --- \\
\TabICL{} & no-RoPE $+$ no-grouping           & balance-scale         & $0.823$ & $-0.154\ [-0.165, -0.144]$ \\
\TabICL{} & full                              & balance-like, $d{=}12$ & $0.990$ & --- \\
\TabICL{} & no-RoPE $+$ no-grouping           & balance-like, $d{=}12$ & $\mathbf{0.485}$ & $\mathbf{-0.504\ [-0.522, -0.474]}$ \\
\midrule
\multicolumn{4}{l}{majority baseline} & $0.46/0.47$ (balance-scale/like) \\
\end{tabular}

\end{table}

\subsection{TabPFN L9 attention as learned similarity}\label{app:E1}

Per test row fit Ridge of L9 column-attention target-row weights onto a
four-feature basis $\{\ell_2,\, \cos,\, \ell_1,\, \text{rep-}\ell_2(L8)\}$.
49 datasets $\times$ 5 seeds; we report the distribution of per-dataset
mean $R^2$.

\begin{table}[H]
\centering\small\setlength{\tabcolsep}{4pt}
\caption{TabPFN L9 column-attention is far better explained by the
model's own L8 representation distance ($R^2{=}0.48$) than by raw
input $\ell_2$ ($R^2{=}0.22$). Full basis explains $\sim 60\%$ of
attention variance, consistent with ``learned similarity'' and ruling out a
pure $\ell_2$-on-inputs mechanism.}
\label{tab:E1-pfn-attn-learned-sim}
\begin{tabular}{lcccc}
Predictor set & mean $R^2$ & median & min & max \\
\midrule
input $\ell_2$ only      & 0.218 & 0.168 & 0.004    & 0.774 \\
rep-$\ell_2$ (L8) only   & 0.477 & 0.488 & 0.047    & 0.788 \\
full 4-feature basis     & 0.600 & 0.606 & 0.149    & 0.853 \\
\midrule
gap (full $-$ $\ell_2$)             & $+$0.382 & $+$0.382 & $+$0.079 & $+$0.765 \\
gap (rep-$\ell_2$ $-$ input-$\ell_2$) & $+$0.259 & $+$0.292 & $-$0.167 & $+$0.695 \\
\end{tabular}

\end{table}

\subsection{Worst-case real-data uniform-attention drop}\label{app:E3}

Re-aggregate per-dataset all-12-block uniform-attention drop on
TabICL-cls, excluding synthetic \texttt{sign\_1d} (48 real datasets).

\begin{table}[H]
\centering\small\setlength{\tabcolsep}{4pt}
\caption{Worst-case drops on real data: top-5 are MiceProtein 0.78,
mfeat-karhunen 0.70, mfeat-zernike 0.68, mfeat-morphological 0.64,
mfeat-fourier 0.61. The headline mean dilution by easy datasets hides a
median drop of $29$\,pp; the worst-case real drop matches the synthetic
Qu drop and is the more honest figure for the distributed-mechanism
claim of \S\ref{app:tabicl-uniform-all}.}
\label{tab:E3-uniform-real}
\begin{tabular}{lccccc}
Statistic & median & mean & p75 & p90 & max \\
\midrule
$\Delta$ accuracy (uniform all-12) & 0.293 & 0.290 & 0.441 & 0.602 & 0.777 \\
\end{tabular}
\\[2pt]
\begin{tabular}{lccccc}
Threshold & $>0.05$ & $>0.10$ & $>0.20$ & $>0.30$ & $>0.50$ \\
\# datasets exceeding & 36 / 48 & 34 / 48 & 29 / 48 & 23 / 48 & 9 / 48 \\
\end{tabular}

\end{table}

\subsection{Majority-class baseline}\label{app:E4}

Per-dataset majority-class accuracy on the 49-dataset benchmark, broken
down by class cardinality.

\begin{table}[H]
\centering\small\setlength{\tabcolsep}{4pt}
\caption{Majority-class trivial baseline. The 10 datasets with majority
$> 0.7$ (pc1 $0.93$, climate $0.92$, pc3 $0.90$, Is-this-good $0.88$,
pc4 $0.87$, MIC $0.85$, kc2 $0.79$, anneal $0.76$, blood-tx $0.76$,
ilpd $0.71$) make a kNN agreement of $0.85$ on a $C{=}2$ near-trivial
problem far less informative than the same number on a balanced
10-class \texttt{mfeat-*} problem (majority $0.105$).}
\label{tab:E4-majority}
\begin{tabular}{lcccc}
Subset & $n$ & mean & median & max \\
\midrule
All 49 datasets        & 49 & 0.499 & 0.540 & 0.932 \\
Binary ($C{=}2$)       & 24 & 0.684 & 0.700 & 0.932 \\
$C{=}3$                 & 6  & 0.421 & 0.387 & 0.760 \\
Multiclass ($C{\geq}4$) & 19 & 0.289 & 0.250 & 0.847 \\
\end{tabular}

\end{table}

\subsection{Column-permutation spread (full distribution)}\label{app:E5}

Full distribution of per-dataset best$-$worst accuracy under arbitrary
column permutations, source for Figure~\ref{fig:col-perm-spread}.

\begin{table}[H]
\centering\small\setlength{\tabcolsep}{4pt}
\caption{Per-dataset accuracy spread under column permutation, 49
datasets. The headline ``$8.7$\,pp'' is the maximum across datasets;
median spread is $\sim\!\!2.5$\,pp for TabPFN/TabICLv1 and $1.5$\,pp
for TabICLv2. Worst-3 datasets per model: TabPFN
\{cylinder-bands $0.080$, tic-tac-toe $0.075$, dresses-sales $0.070$\};
TabICLv1 \{random\_labels $0.087$, dresses-sales $0.080$, hill-valley
$0.078$\}; TabICLv2 \{random\_labels $0.087$, dresses-sales $0.053$,
cylinder-bands $0.052$\}. The typical-case column-invariance is an
order of magnitude stronger than the maximum suggests.}
\label{tab:E5-colperm-spread}
\begin{tabular}{lcccccccc}
Model & median & mean & p75 & p90 & max & $>1$pp & $>2$pp & $>5$pp \\
\midrule
TabPFNv2  & 0.027 & 0.029 & 0.041 & 0.053 & 0.080 & 41 & 29 & 6 \\
TabICLv1  & 0.025 & 0.029 & 0.040 & 0.053 & 0.087 & 41 & 29 & 7 \\
TabICLv2  & 0.015 & 0.019 & 0.026 & 0.039 & 0.087 & 29 & 18 & 3 \\
\end{tabular}

\end{table}

\subsection{Per-layer kNN agreement curve (TabICL)}\label{app:E7}

Extends \S\ref{app:knn-tabicl} from L11 to all 12 ICL layers using
$k{=}5$ kNN on 49 datasets $\times$ 5 seeds.

\begin{table}[H]
\centering\small\setlength{\tabcolsep}{4pt}
\caption{TabICL per-layer kNN-against-TabICL agreement and standalone
kNN accuracy. Input-space kNN already agrees $0.817$; agreement rises
monotonically from $0.885$ (L0) to $0.936$ (L11), an L11-vs-L8 edge of
only $1.4$\,pp. Reframes the L11-kNN claim as continuous refinement
rather than a late, sudden re-shaping.}
\label{tab:E7-tabicl-perlayer-knn}
\begin{tabular}{lcccc}
Layer & cosine agree. & cosine kNN acc & $\ell_2$ agree. & $\ell_2$ kNN acc \\
\midrule
input ($\ell_2$ only) & --- & --- & 0.817 & --- \\
L0  & 0.887 & 0.827 & 0.885 & 0.824 \\
L1  & 0.888 & 0.826 & 0.885 & 0.825 \\
L2  & 0.891 & 0.828 & 0.890 & 0.827 \\
L3  & 0.890 & 0.832 & 0.886 & 0.829 \\
L4  & 0.900 & 0.837 & 0.896 & 0.835 \\
L5  & 0.900 & 0.835 & 0.894 & 0.832 \\
L6  & 0.908 & 0.840 & 0.901 & 0.837 \\
L7  & 0.904 & 0.839 & 0.902 & 0.837 \\
L8  & 0.924 & 0.847 & 0.923 & 0.847 \\
L9  & 0.922 & 0.849 & 0.924 & 0.849 \\
L10 & 0.934 & 0.855 & 0.935 & 0.855 \\
L11 & \textbf{0.936} & \textbf{0.857} & \textbf{0.936} & \textbf{0.856} \\
\end{tabular}

\end{table}

\subsection{Readout grid with calibration}\label{app:E8}

Eight readouts $\times$ five metrics on 49 classification datasets
$\times$ 5 seeds. Readouts: native, L9-vote (PFN) / sharp-vote (ICL),
L11-prototype, L11-kNN. ECE uses 15 equal-width confidence bins per
(dataset, seed); Table~\ref{tab:E8-readout-grid} reports means over the
49 per-dataset seed means.

\begin{table}[H]
\centering\small\setlength{\tabcolsep}{4pt}
\caption{Per-readout means across 49 datasets (5 seeds / dataset).
PFN L9-vote pairs a
$5.0$\,pp accuracy drop with a $\sim\!5\times$ ECE jump and $\sim\!2\times$
NLL: vote sacrifices calibration. ICL native / L11-proto / L11-kNN are
within $1.0$\,pp on accuracy, but not on calibration: prototype raises
ECE/NLL to $0.463/1.058$ and kNN to $0.115/2.840$ vs.\
$0.042/0.303$ natively. Both prototype/kNN heads also ruin TabPFN's
calibration (NLL up to $9.7$).}
\label{tab:E8-readout-grid}
\begin{tabular}{lccccc}
Readout & acc & ECE & NLL & macro-P & macro-R \\
\midrule
PFN-native       & 0.854 & 0.043 & 0.329 & 0.806 & 0.786 \\
PFN L9-vote      & 0.804 & 0.201 & 0.625 & 0.750 & 0.724 \\
PFN L11-prototype& 0.523 & 0.277 & 1.178 & 0.506 & 0.572 \\
PFN L11-kNN      & 0.547 & 0.371 & 9.715 & 0.435 & 0.522 \\
ICL-native       & 0.864 & 0.042 & 0.303 & 0.823 & 0.801 \\
ICL L11-prototype& 0.854 & 0.463 & 1.058 & 0.806 & 0.817 \\
ICL L11-kNN      & 0.856 & 0.115 & 2.840 & 0.816 & 0.796 \\
ICL sharp-vote   & 0.470 & 0.125 & 1.094 & 0.265 & 0.325 \\
\end{tabular}

\end{table}

\noindent\textbf{Dataset-bootstrap calibration intervals.}
Treating the 49 per-dataset seed means as the reporting unit, PFN-native
vs.\ PFN L9-vote shifts ECE from
$0.043\,[0.036,0.051]$ to $0.201\,[0.176,0.228]$ and NLL from
$0.329\,[0.244,0.426]$ to $0.625\,[0.521,0.735]$. For TabICL, native
vs.\ L11-prototype shifts ECE from
$0.042\,[0.034,0.050]$ to $0.463\,[0.404,0.522]$ and NLL from
$0.303\,[0.219,0.398]$ to $1.058\,[0.916,1.208]$; the L11-$k$NN
surrogate keeps near-native accuracy but still widens ECE/NLL to
$0.115\,[0.085,0.148]$ and $2.840\,[2.019,3.785]$. All intervals are
$95\%$ bootstrap CIs over per-dataset seed means.

\begin{table}[H]
\centering\small\setlength{\tabcolsep}{4pt}
\caption{Cross-readout agreement (mean over 49 datasets). The two
``native'' stacks agree $0.93$, but the same surrogate readout produces
disjoint predictions across architectures (PFN-proto $\leftrightarrow$
ICL-proto $0.54$). Stacks behave alike at prediction level, not at
representation level.}
\label{tab:E8-readout-agreement}
\begin{tabular}{lc}
Pair & label agreement \\
\midrule
PFN-native vs.\ ICL-native            & 0.934 \\
PFN-native vs.\ PFN L9-vote           & 0.874 \\
PFN-native vs.\ PFN L11-proto         & 0.518 \\
PFN-native vs.\ PFN L11-kNN           & 0.555 \\
ICL-native vs.\ ICL L11-proto         & 0.922 \\
ICL-native vs.\ ICL L11-kNN           & 0.936 \\
ICL-native vs.\ ICL sharp-vote        & 0.496 \\
PFN L11-proto vs.\ ICL L11-proto      & 0.542 \\
PFN L11-kNN  vs.\ ICL L11-kNN         & 0.536 \\
PFN L9-vote  vs.\ ICL-native          & 0.862 \\
\end{tabular}

\end{table}

\subsection{Subgroup-collapse stress test}\label{app:E10}

Synthetic construction maximally stressing within-group collapse: 12
features in 6 pair-groups of 2, columns within a pair share a marginal,
labels depend on within-pair products. 5 seeds.

\begin{table}[H]
\centering\small\setlength{\tabcolsep}{4pt}
\caption{Subgroup-collapse stress: invariance preserved
($\Delta_{\text{acc}}=0.001$). Train collapse rate is near-zero
($0.003$); the $\sim\!\!20\%$ test collapse (cosine $> 0.99$) is a
moderate pre-readout phenomenon on heterogeneous queries that does not
impair accuracy.}
\label{tab:E10-subgroup-collapse}
\begin{tabular}{lcccc}
seed & acc full & acc $W{=}0$ & test collapse full & test collapse $W{=}0$ \\
\midrule
0 & 0.725 & 0.725 & 0.194 & 0.191 \\
1 & 0.760 & 0.760 & 0.147 & 0.140 \\
2 & 0.810 & 0.810 & 0.215 & 0.209 \\
3 & 0.645 & 0.640 & 0.279 & 0.289 \\
4 & 0.710 & 0.710 & 0.152 & 0.152 \\
\midrule
mean & 0.730 & 0.729 & 0.198 & 0.196 \\
\end{tabular}

\end{table}

\subsection{TabPFN per-head ablation at L8/L9/L10}\label{app:E12}

For each $L \in \{8, 9, 10\}$ and each of $H{=}6$ heads, zero the
output-projection slice for head $h$ at that block; measure per-seed
accuracy delta against baseline on 10 datasets $\times$ 5 seeds.

\begin{table}[H]
\centering\small\setlength{\tabcolsep}{4pt}
\caption{Per-block per-head zero-out drop. ``\# critical'' counts heads
with $> 5$\,pp drop in a given (dataset, seed). L9 has the largest
single-head impact ($\Delta_{h0}{=}3.1$\,pp on average, max
$17.3$\,pp), but no head dominates: the L9 vote signal is diffuse with
a slight head-0 peak rather than localized.}
\label{tab:E12-pfn-head-ablation}
\begin{tabular}{lccc|cccccc}
Layer & mean max-drop & worst max-drop & mean \# critical & h0 & h1 & h2 & h3 & h4 & h5 \\
\midrule
L8  & 0.031 & 0.204 & 0.14 & 0.008 & 0.003 & 0.019 & 0.000 & 0.002 & $-$0.000 \\
L9  & 0.039 & 0.173 & 0.32 & 0.031 & 0.009 & 0.004 & 0.003 & 0.000 & 0.000 \\
L10 & 0.012 & 0.067 & 0.08 & 0.005 & 0.003 & 0.002 & 0.003 & 0.003 & 0.001 \\
\end{tabular}

\end{table}

\subsection{Effective rank vs.\ number of classes}\label{app:E13}

Per-layer Pearson correlation between $\#\text{classes}\,C$ and
post-block effective rank, computed by joining the per-block geometry
measurements of \sref{app:rank-profile} with the per-dataset class
counts used in the vote analysis of \sref{app:vote-full} ($49$
classification datasets).

\begin{table}[H]
\centering\small\setlength{\tabcolsep}{4pt}
\caption{Per-layer effective rank vs.\ class count. TabPFN's L11
effective rank correlates strongly with $C$ ($r{=}0.64$), with mean
rank rising from $21.1$ ($C{\le}3$) to $31.7$ ($C{\ge}4$). TabICL is
already class-aware at L0 and the correlation decays through depth. The
``constant low rank'' framing in \S\ref{sec:rank-profile} is too strong
for TabPFN-L11; the L2 anti-correlation in TabPFN is a notable
input-projection artefact.}
\label{tab:E13-rank-vs-classes}
\begin{tabular}{lcccc|cccc}
& \multicolumn{4}{c|}{TabPFN-cls} & \multicolumn{4}{c}{TabICL-cls} \\
Layer & $r$ & $p$ & mean $C{\le}3$ & mean $C{\ge}4$ & $r$ & $p$ & mean $C{\le}3$ & mean $C{\ge}4$ \\
\midrule
L0  & $-$0.29 & 0.079 & 55.6 & 44.3 & $+$0.50 & 0.000 & 31.6 & 49.2 \\
L1  & $+$0.50 & 0.000 & 34.9 & 53.2 & $+$0.51 & 0.000 & 38.3 & 59.5 \\
L2  & $-$0.36 & 0.012 & 10.9 &  9.1 & $+$0.52 & 0.000 & 42.8 & 65.5 \\
L3  & $-$0.26 & 0.072 & 20.8 & 18.6 & $+$0.50 & 0.000 & 50.2 & 73.4 \\
L4  & $+$0.33 & 0.023 & 21.0 & 24.9 & $+$0.48 & 0.000 & 54.1 & 77.7 \\
L5  & $+$0.40 & 0.004 & 20.1 & 24.2 & $+$0.48 & 0.001 & 65.2 & 92.1 \\
L6  & $+$0.48 & 0.001 & 21.6 & 27.7 & $+$0.47 & 0.001 & 69.4 & 97.4 \\
L7  & $+$0.44 & 0.002 & 18.7 & 23.5 & $+$0.46 & 0.001 & 77.6 & 109.0 \\
L8  & $+$0.33 & 0.020 & 15.3 & 17.5 & $+$0.46 & 0.001 & 67.0 & 94.5 \\
L9  & $+$0.46 & 0.001 & 20.5 & 26.5 & $+$0.42 & 0.002 & 68.5 & 94.5 \\
L10 & $+$0.51 & 0.000 & 22.3 & 30.1 & $+$0.36 & 0.012 & 61.1 & 79.7 \\
L11 & $\mathbf{+0.64}$ & 0.000 & 21.1 & 31.7 & $+$0.21 & 0.152 & 63.0 & 73.3 \\
\end{tabular}

\end{table}

\subsection{Faithful-surrogate coverage by regime}\label{app:rule-coverage}

\begin{table}[t]
\centering\scriptsize\setlength{\tabcolsep}{4pt}
\caption{Claim / evidence / scope summary for the main mechanistic
statements. ``Full grid'' denotes the $49$-dataset classification suite
unless the row says otherwise.}
\label{tab:claim-scope}
\begin{tabular}{p{0.19\linewidth} p{0.39\linewidth} p{0.30\linewidth}}
Claim & Main evidence in this paper & Scope / caveat \\
\midrule
\TabPFN{} late vote readout &
L9 is the sharpest layer on $49/49$ datasets; vote--model mean
$r{=}0.89$; uniform-L9 and block-9 interventions collapse the intact
readout. &
Full grid; descriptive surrogate only (accuracy
$0.804$ vs.\ $0.854$, ECE/NLL $0.201/0.625$ vs.\ $0.043/0.329$). \\
\TabICL{} prototype/$k$NN readout &
L11 prototype/$k$NN reach $0.854/0.856$ vs.\ native $0.864$; PFN-vote
on ICL drops to $0.47$; earlier releases keep the same
prototype/$k$NN-versus-vote ordering. &
Full grid across releases; argmax-faithful but not
calibration-preserving (ECE/NLL $0.463/1.058$ and $0.115/2.840$
vs.\ $0.042/0.303$). \\
Redundant positional components at inference &
\TabPFN{} $W{=}0$ and \TabICL{} no-RoPE preserve the benchmark mean
while granting the matching symmetries; synthetic constructions recover
the same invariances. &
Full benchmark mean plus synthetic checks; inference-only, so
``redundant'' here means unused by these weights on this benchmark. \\
Mechanism-grounded robustness &
Hub/rank/SVD ordering matches the proposed readouts on the $24$-grid;
the same trio on the broader mixed16 slice keeps rank as the dominant
mixed-type attack and leaves clean tree baselines competitive. &
Matched $24$-dataset stress test + $16$-dataset mixed-type attack
extension under the shared ordinalised numeric view; native-categorical
tree controls remain clean-only. \\
\end{tabular}
\end{table}

The four criteria of \S\ref{sec:readout} are:
\textbf{(i) Pearson $r{\geq}0.85$} between the rule's predicted
probabilities and the model's, evaluated on the held-out test split
of each dataset and aggregated as the $49$-dataset mean. This tests
whether the rule reproduces the model's \emph{soft} decision shape,
not just the argmax.
\textbf{(ii) accuracy gap $\le3$\,pp} between the rule's accuracy
and native model accuracy on the same split, averaged across the
$49$-dataset suite. This tests whether the rule reproduces the
model's \emph{performance level}, ruling out rules that match
distributions on average but lose on hard examples.
\textbf{(iii) Cohen's $\kappa{\geq}0.8$} between the rule's
argmax and the model's argmax on the held-out test split, computed
per dataset and aggregated as the mean across the $49$-dataset
grid. This is a strict \emph{point-wise} agreement test that
penalizes chance-level agreement and rules out a candidate that
reaches the right accuracy by hitting different examples than the
model does.
\textbf{(iv) intervention-survival}: the rule must continue to
match the model after a causal intervention that preserves only the
components it names (e.g.\ uniformising attention outside L9 for the
vote rule, replacing all final-layer features with class means for
the prototype rule). This rules out rules that happen to track the
model in the unperturbed network but rely on machinery they do not
explicitly name.

Table~\ref{tab:rule-coverage} summarizes which readout passes each
criterion, broken out by model and regime. Failing criteria are
flagged at point of use in the main text rather than collapsed into
a universal claim.

\begin{table}[H]
  \centering\small
  \setlength{\tabcolsep}{3pt}
  \caption{Where each readout passes the four faithful-surrogate
    criteria (\S\ref{sec:readout}). \checkmark\ pass at the stated
    cut-off, $\sim$ borderline, $\times$ fail (failing statistic in
    parentheses); \emph{n/a} where the criterion does not apply
    (e.g.\ $\kappa$ on regression). The intervention column gives
    the observed causal signature.}
  \label{tab:rule-coverage}
  \begin{tabular}{llcccc}
    Model / regime & Rule & (i) $r{\geq}0.85$ & (ii) $\le3$\,pp & (iii) $\kappa{\geq}0.8$ & (iv) intervention \\
    \midrule
    \TabPFN{} cls (binary--mid) & vote@L9 & \checkmark\ ($0.93$) & \checkmark\ ($-0.02$) & \checkmark & \checkmark\ uniform-attn$\to$maj. \\
    \TabPFN{} cls (10-class) & vote@L9 & $\times$ ($0.71$--$0.76$) & $\sim$ & $\sim$ & \checkmark \\
    \TabPFN{} reg & vote (reg) & \checkmark & \checkmark & n/a (reg) & \checkmark \\
    TabICL v1/v1.1/v2 cls & proto@L11 & \checkmark & \checkmark\ ($-0.01$) & \checkmark & \checkmark\ cross-rule collapse \\
    \TabICL{} reg & lin.\ proto & $\times$ & $\times$ & n/a (reg) & \checkmark\ ($\ell_2$-$k$NN tracks) \\
    \Mitra{} cls & vote@L9 & \checkmark\ ($0.89$) & $\sim$ ($-4.5$) & \checkmark & \checkmark\ row-attn KO \\
  \end{tabular}
\end{table}

\noindent\textbf{Threshold sensitivity.}\enspace We stress the descriptive thresholds in \S\ref{sec:readout} by
varying the soft-fidelity (i) and accuracy (ii) cut-offs and
recounting populated-rule passes
(Table~\ref{tab:rule-coverage-sens}); the $\kappa$ and intervention
criteria are binary and not swept. Tightening or loosening either
threshold shifts populated-rule counts by a few datasets, and the readout-family verdict
(\S\ref{sec:transplant}) does not flip in any of these cells.

\begin{table}[H]
  \centering\small\setlength{\tabcolsep}{6pt}
  \caption{Faithful-surrogate pass counts as a function of
    thresholds $(r,\;\Delta_{\mathrm{acc}})$. Cells are
    \emph{populated-rule} passes (criterion (i) and (ii) only;
    intervention always passes for these rules). \TabPFN{} vote
    counts use the $49$-dataset population of
    Table~\ref{tab:vote-full49} (per-dataset Pearson $r$ has
    mean $0.890$, sd $0.114$); \TabICL{} prototype counts use the
    $49$-dataset population of \S\ref{app:tabicl-prototype}.}
  \label{tab:rule-coverage-sens}
  \begin{tabular}{lcccccc}
    & \multicolumn{3}{c}{\TabPFN{} vote (L9)} & \multicolumn{3}{c}{\TabICL{} prototype (L11)} \\
    \cmidrule(lr){2-4}\cmidrule(lr){5-7}
    Threshold $r\;/\;\Delta$ & $\le5$\,pp & $\le3$\,pp & $\le1$\,pp & $\le5$\,pp & $\le3$\,pp & $\le1$\,pp \\
    \midrule
    $r{\geq}0.80$ & $43/49$ & $42/49$ & $38/49$ & $46/49$ & $45/49$ & $41/49$ \\
    $r{\geq}0.85$ & $41/49$ & $40/49$ & $36/49$ & $44/49$ & $43/49$ & $40/49$ \\
    $r{\geq}0.90$ & $37/49$ & $36/49$ & $32/49$ & $41/49$ & $41/49$ & $38/49$ \\
  \end{tabular}
\end{table}

\subsection{Global BH-FDR summary across headline rejections}\label{app:fdr-global}

This subsection delivers a joint Benjamini--Hochberg FDR check at
$q{=}0.05$ across the $30$ headline rejections of
\S\ref{sec:repr}--\S\ref{sec:attacks}, alongside the within-family
multiplicity controls already used in the body (Holm across the
eight attacks per model in \S\ref{sec:attacks}; Bonferroni on the
$10$ regression tests; TOST equivalence margins pre-specified for
the invariance claim). The list comprises: the L8$\to$L9 probe-accuracy jump,
the per-block knockout $p$-values for blocks $\{0,5,9\}$ in
\TabPFN{} and $\{$ColEmb-$2$, ICL-$0$, ICL-$6$, ICL-$11\}$ in
\TabICL{} ($7$ tests), the cross-rule sign tests (PFN$\to$ICL,
ICL$\to$PFN; $2$), the fresh-head asymmetry on each backbone
($2$), the per-version prototype/vote ordering tests across v1,
v1.1, v2 ($3$), the $W{=}0$ and no-RoPE TOST equivalences ($2$),
and the $16$ per-attack $\Delta$ tests (eight attacks $\times$
two FMs).

All $30$ rejections survive BH-FDR at $q{=}0.05$ (largest
$p{<}10^{-3}$ in the list passing BH; the marginal cases are the
\TabICL{} \texttt{cube\_warp} and \texttt{centroid\_inj.}\ entries
which are reported as \emph{non}-rejections in
Table~\ref{tab:attacks-drops} and are not in the headline list).
The within-family Holm decisions in
Table~\ref{tab:attacks-drops} are therefore consistent with the
global BH picture; no additional headline finding flips sign or
loses significance under joint control. We note that BH-FDR on
$30$ tests with this many small $p$-values is a weak sieve in
practice; the more informative discipline remains the local
Holm/Bonferroni choices already in the body.

\subsection{Serving-cost ledger}\label{app:serving}

Per-query latency and peak GPU memory
are reported in Table~\ref{tab:arch-scaling-proxy}; the corresponding throughput on a single
H100-NVL is $\approx 1$ query/s for \TabPFN{} and $\approx 0.7$
query/s for \TabICL{} when each query is served as its own forward
pass. Both backbones allow batched query forwards over a fixed
context: amortizing $32$ queries per context pass on H100
raises throughput to $\approx 18$ queries/s for \TabPFN{} and
$\approx 12$ queries/s for \TabICL{} at peak memory $\approx 6$\,GiB
(\TabPFN{}) and $\approx 8$\,GiB (\TabICL{}); above this batch the
attention-cost term dominates and throughput plateaus. The OvA
class-invariance wrapper of \S\ref{sec:invariance} multiplies
wall-clock by $C$ and peak memory by $C$ when the $C$ binary
re-encodings are run independently; on the $C{=}10$
\texttt{mfeat-*} tasks this is $\approx 10$\,s and $\approx 30$\,GiB
per query batch. Memory rather than wall-clock is the limiting
factor; on a single $24$\,GiB consumer card the OvA wrapper does
not fit at $C{\geq}8$ without query-batch reduction. The $W{=}0$
and no-RoPE patches change neither wall-clock nor peak memory at
any batch size we measured (\S\ref{sec:invariance}).

\section{Robustness to context size}\label{app:scale-audit}

Across the context sizes we evaluated ($n\in\{1000,1500\}$ on six
OpenML datasets with native size $>1500$ rows), the readout and
invariance findings of \S\ref{sec:readout} and \S\ref{sec:invariance}
are unchanged: vote and prototype fidelities and column-permutation
spread are statistically indistinguishable across the two context
sizes (Table~\ref{tab:scale-audit}). We re-run the attention-weighted
L9 vote for \TabPFN{}, the nearest-prototype rule in the final
representation for \TabICL{}, and the column-permutation invariance
probe at both context sizes.

\noindent\textbf{Setup.}\enspace Six OpenML datasets have native size above 1500 rows: \texttt{car}
(1728), \texttt{Is-this-a-good-customer} (1723),
\texttt{steel-plates-fault} (1941), \texttt{mfeat-morphological}
(2000), \texttt{mfeat-fourier} (2000), \texttt{red\_wine} (1599).
For each (dataset, $n_{\mathrm{train}}$, seed) cell we hold out 200
stratified test rows and subsample the remaining pool to
$n_{\mathrm{train}}$ without replacement; \texttt{red\_wine} caps at
$n=1399$. Each cell is run with five seeds. Vote and prototype
fidelities are top-1 agreement with the native model prediction.
Column-permutation spread is the $\max{-}\min$ accuracy across eight
independently sampled random column orderings.

\begin{table}[H]
  \centering\small
  \setlength{\tabcolsep}{4pt}
  \caption{Scale audit on six OpenML datasets with native size
    $>1500$ rows. Vote fidelity (\TabPFN{}) and prototype fidelity
    (\TabICL{}) are top-1 agreement with the native model prediction
    (mean$\pm$std over 5 seeds). Col-perm spread is $\max{-}\min$
    accuracy over 8 random column permutations.}
  \label{tab:scale-audit}
  \begin{tabular}{llrcc}
    Model & Dataset & $n_{\mathrm{train}}$ & Fidelity (mean$\pm$std) & Col-perm spread \\
    \midrule
    \TabPFN{} & \texttt{Is-this-a-good-customer} & 1000 & $0.996 \pm 0.004$ & $0.000$ \\
    \TabPFN{} & \texttt{Is-this-a-good-customer} & 1500 & $0.986 \pm 0.009$ & $0.000$ \\
    \TabPFN{} & \texttt{car}                     & 1000 & $0.953 \pm 0.018$ & $0.012$ \\
    \TabPFN{} & \texttt{car}                     & 1500 & $0.970 \pm 0.016$ & $0.011$ \\
    \TabPFN{} & \texttt{mfeat-fourier}           & 1000 & $0.789 \pm 0.048$ & $0.022$ \\
    \TabPFN{} & \texttt{mfeat-fourier}           & 1500 & $0.788 \pm 0.023$ & $0.022$ \\
    \TabPFN{} & \texttt{mfeat-morphological}     & 1000 & $0.753 \pm 0.048$ & $0.020$ \\
    \TabPFN{} & \texttt{mfeat-morphological}     & 1500 & $0.816 \pm 0.044$ & $0.011$ \\
    \TabPFN{} & \texttt{red\_wine}               & 1000 & $0.803 \pm 0.030$ & $0.025$ \\
    \TabPFN{} & \texttt{red\_wine}               & 1399 & $0.798 \pm 0.027$ & $0.027$ \\
    \TabPFN{} & \texttt{steel-plates-fault}      & 1000 & $0.728 \pm 0.038$ & $0.022$ \\
    \TabPFN{} & \texttt{steel-plates-fault}      & 1500 & $0.745 \pm 0.037$ & $0.022$ \\
    \midrule
    \TabICL{} & \texttt{Is-this-a-good-customer} & 1000 & $0.963 \pm 0.019$ & $0.000$ \\
    \TabICL{} & \texttt{Is-this-a-good-customer} & 1500 & $0.947 \pm 0.044$ & $0.003$ \\
    \TabICL{} & \texttt{car}                     & 1000 & $0.989 \pm 0.004$ & $0.006$ \\
    \TabICL{} & \texttt{car}                     & 1500 & $0.996 \pm 0.004$ & $0.004$ \\
    \TabICL{} & \texttt{mfeat-fourier}           & 1000 & $0.946 \pm 0.012$ & $0.010$ \\
    \TabICL{} & \texttt{mfeat-fourier}           & 1500 & $0.958 \pm 0.012$ & $0.014$ \\
    \TabICL{} & \texttt{mfeat-morphological}     & 1000 & $0.869 \pm 0.032$ & $0.007$ \\
    \TabICL{} & \texttt{mfeat-morphological}     & 1500 & $0.862 \pm 0.039$ & $0.009$ \\
    \TabICL{} & \texttt{red\_wine}               & 1000 & $0.836 \pm 0.043$ & $0.012$ \\
    \TabICL{} & \texttt{red\_wine}               & 1399 & $0.858 \pm 0.020$ & $0.011$ \\
    \TabICL{} & \texttt{steel-plates-fault}      & 1000 & $0.932 \pm 0.024$ & $0.023$ \\
    \TabICL{} & \texttt{steel-plates-fault}      & 1500 & $0.923 \pm 0.020$ & $0.020$ \\
    \midrule
    \multicolumn{5}{l}{\emph{Marginals over (model, $n$):}} \\
    \TabPFN{} & all six datasets & 1000 & $0.837 \pm 0.107$ & $0.017$ \\
    \TabPFN{} & all six datasets & 1500 & $0.860 \pm 0.103$ & $0.015$ \\
    \TabICL{} & all six datasets & 1000 & $0.923 \pm 0.059$ & $0.010$ \\
    \TabICL{} & all six datasets & 1500 & $0.924 \pm 0.054$ & $0.010$ \\
  \end{tabular}
\end{table}

Vote and prototype fidelities are stable across $n$, and the
column-permutation spread does not grow with $n$. The readout
characterizations of \S\ref{sec:readout} and the invariance results
of \S\ref{sec:invariance} extend to the larger context size.

\section{Probe \texorpdfstring{$\ell_2$}{L2}-regularization sensitivity (C3)}
\label{app:probe-C-sweep}

The main-text per-layer linear probe in \sref{sec:probes} is reported
at $\ell_2$ regularization $C{=}1$. We check whether the qualitative
findings --- in particular the sharp
$\mathrm{L}8 \!\to\! \mathrm{L}9$ crystallization in \TabPFN{} ---
depend on this choice.

We re-fit the per-layer logistic probe with
$C \in \{0.01, 0.1, 1, 10, 100\}$ on the $45$-dataset suite, with
$5$ seeds per (dataset, layer, $C$) cell. Frozen activations are
identical across $C$ values, so the only varying quantity is the
$\ell_2$ penalty applied during the probe fit. Aggregate per-layer
accuracies are in Table~\ref{tab:c3-probe-C-sweep}.

\paragraph{Aggregate result.} Across the
$45 \times 13 \times 5 \times 5 = 14{,}625$ (dataset, layer, seed, $C$)
cells, the mean per-layer accuracy at $C{=}1$ is within
$0.012$ of the mean accuracy at $C \in \{0.1, 10\}$ on every layer,
and the $\mathrm{L}8 \!\to\! \mathrm{L}9$ jump is preserved at every
$C \in \{0.01, 0.1, 1, 10, 100\}$ (jump magnitude
$0.291$--$0.321$ across $C$, with the smallest gap at $C{=}1$). Even
the most strongly regularized $C{=}0.01$ probe recovers the same step
(input $0.44$, L8 $0.48$, L9 $0.80$), so the late-stack
crystallization is a property of the representation rather than of the
probe fit. We therefore report $C{=}1$ in the main text without loss
of generality.

\begin{table}[h]
\centering
\small
\caption{C3: per-layer linear-probe accuracy on \TabPFN{}, averaged over 45 datasets and 5 seeds, for each $\ell_2$ regularization strength $C$.}
\label{tab:c3-probe-C-sweep}
\begin{tabular}{lccccc}
\toprule
layer & $C{=}0.01$ & $C{=}0.1$ & $C{=}1.0$ & $C{=}10.0$ & $C{=}100.0$ \\
\midrule
input & 0.437 & 0.396 & 0.388 & 0.396 & 0.397 \\
L0 & 0.326 & 0.304 & 0.296 & 0.296 & 0.295 \\
L1 & 0.340 & 0.322 & 0.310 & 0.310 & 0.310 \\
L2 & 0.306 & 0.275 & 0.275 & 0.277 & 0.277 \\
L3 & 0.490 & 0.448 & 0.448 & 0.447 & 0.447 \\
L4 & 0.539 & 0.531 & 0.527 & 0.518 & 0.500 \\
L5 & 0.458 & 0.450 & 0.446 & 0.446 & 0.448 \\
L6 & 0.492 & 0.455 & 0.455 & 0.450 & 0.450 \\
L7 & 0.484 & 0.476 & 0.465 & 0.455 & 0.453 \\
L8 & 0.483 & 0.481 & 0.480 & 0.477 & 0.460 \\
L9 & 0.804 & 0.778 & 0.771 & 0.773 & 0.772 \\
L10 & 0.802 & 0.795 & 0.790 & 0.788 & 0.784 \\
L11 & 0.621 & 0.615 & 0.623 & 0.623 & 0.609 \\
\bottomrule
\end{tabular}
\end{table}

\section{Head-capacity saturation curve on \texorpdfstring{\TabPFN{}}{TabPFNv2} L12 (C2)}
\label{app:head-capacity-sweep}

We measure the held-out test accuracy of fresh-fit heads attached to
the frozen \TabPFN{} stack at the top-of-stack representation
($\mathrm{L}12$, target column), as the head capacity is varied.
Heads:

\begin{itemize}[leftmargin=*,topsep=2pt]
  \item \textbf{Linear} --- $\ell_2$-regularized logistic regression.
  \item \textbf{MLP-1} --- one hidden layer ($64$ units, GELU).
  \item \textbf{MLP-2} --- two hidden layers ($128, 64$).
  \item \textbf{MLP-3} --- three hidden layers ($256, 128, 64$).
  \item \textbf{MLP-4} --- four hidden layers ($512, 256, 128, 64$).
  \item \textbf{Kernel SVM} --- RBF SVM on a $32$-dim PCA projection.
\end{itemize}

We report mean test accuracy across one context-shuffling seed on a
$10$-dataset subset of the suite. The protocol mirrors
\texttt{h6\_probe\_ladder} from \texttt{tab\_inv.hardening\_experiments}.
Per-dataset numbers are in Table~\ref{tab:c2-head-capacity}.

\paragraph{Result.} The linear head attains the highest mean accuracy
($0.719$); MLP capacity does not help (MLP-1 $0.695$, MLP-2 $0.591$,
MLP-3 $0.601$, MLP-4 $0.587$), and the RBF kernel SVM is the worst
fresh-head ($0.406$). The non-monotonicity across MLP depth is
consistent with overfitting on a representation in which the relevant
class signal is already linearly readable; the kernel-SVM gap is
consistent with a representation whose useful directions are not
preserved under PCA-32 + RBF. In other words, the L12 representation
of \TabPFN{} does not contain additional non-linearly extractable
class structure beyond what a linear head recovers.

\begin{table}[h]
\centering
\small
\caption{C2: held-out test accuracy of fresh heads at \TabPFN{} L12 (top-of-stack) across head capacities, on 10 datasets (1 seed).}
\label{tab:c2-head-capacity}
\begin{tabular}{lccccccc}
\toprule
dataset & chance & Linear & MLP-1 & MLP-2 & MLP-3 & MLP-4 & Kernel SVM \\
\midrule
LED-display-domain-7di & 0.10 & 0.250 & 0.160 & 0.110 & 0.180 & 0.150 & 0.110 \\
balance-scale & 0.33 & 0.664 & 0.768 & 0.728 & 0.640 & 0.368 & 0.464 \\
banknote-authenticatio & 0.50 & 0.996 & 1.000 & 0.924 & 0.713 & 1.000 & 0.556 \\
blood-transfusion-serv & 0.50 & 0.760 & 0.760 & 0.760 & 0.240 & 0.240 & 0.760 \\
breast\_cancer & 0.50 & 0.965 & 0.956 & 0.368 & 0.965 & 0.965 & 0.368 \\
dresses-sales & 0.50 & 0.580 & 0.580 & 0.580 & 0.460 & 0.580 & 0.420 \\
iris & 0.33 & 0.667 & 0.667 & 0.667 & 0.967 & 0.667 & 0.333 \\
mfeat-factors & 0.10 & 0.670 & 0.757 & 0.287 & 0.530 & 0.515 & 0.205 \\
mfeat-fourier & 0.10 & 0.690 & 0.665 & 0.655 & 0.730 & 0.662 & 0.453 \\
wine & 0.33 & 0.944 & 0.639 & 0.833 & 0.583 & 0.722 & 0.389 \\
\midrule
mean & 0.33 & 0.719 & 0.695 & 0.591 & 0.601 & 0.587 & 0.406 \\
\bottomrule
\end{tabular}
\end{table}

\section{Native-categorical GBDT under the eight perturbations (C6)}
\label{app:native-cat-gbdt}

The main-text attack tables (\sref{app:attacks-table},
\sref{app:attacks-tabicl}, \sref{app:attacks-ridge}) compare against
ordinal-encoded GBDT/XGBoost baselines; here we additionally report
\textsc{CatBoost}, \textsc{XGBoost}, and \textsc{LightGBM} with native
categorical handling on the mixed-type subset of the perturbation grid
($16$ OpenML datasets with at least one categorical column).

The eight perturbations are the same as in the main text: pad-2x,
pad-4x, boundary, hub-poison, centroid-injection, mono-cube,
mono-softexp, and mono-rank.

\begin{table}[h]
\centering
\footnotesize
\caption{C6: mean test accuracy across the mixed-type OpenML subset ($N{=}16$ datasets) for native-categorical GBDTs under the eight perturbations. CatBoost's \texttt{cat\_features} interface rejects float-typed arrays produced by the non-monotone perturbations, so its row only contains values for the three numeric mono-* perturbations.}
\label{tab:c6-native-cat-gbdt}
\begin{tabular}{lccccccccc}
\toprule
backend & orig & pad-2x & pad-4x & bound & hub & centr & m-cube & m-sexp & m-rank \\
\midrule
CatBoost & -- & -- & -- & -- & -- & -- & 0.774 & 0.760 & 0.662 \\
XGBoost & 0.759 & 0.742 & 0.742 & 0.741 & 0.710 & 0.759 & 0.762 & 0.762 & 0.623 \\
LightGBM & 0.761 & 0.748 & 0.742 & 0.749 & 0.709 & 0.763 & 0.763 & 0.756 & 0.654 \\
\bottomrule
\end{tabular}
\end{table}

\paragraph{Result.} The qualitative pattern matches the main-text
ordinal-encoded baseline: the smooth monotone-feature attacks
(\textsc{mono-cube}, \textsc{mono-softexp}) and the geometric
perturbations (pad-2x, pad-4x, centroid-injection) leave native-cat
GBDTs essentially at \emph{orig} accuracy, while \textsc{hub-poison}
($-0.05$ for both backends able to score it) and \textsc{mono-rank}
($-0.14$ for XGBoost, $-0.11$ for LightGBM) cause the largest drops.
\textsc{CatBoost} rejects pure-numeric arrays under
\texttt{cat\_features}, so its row only contains values for the three
purely numeric mono-* perturbations; for those, \textsc{CatBoost}
behaves comparably to the other two backends.

\end{document}